\documentclass[runningheads]{llncs}
\usepackage{graphicx}

\usepackage{tikz}
\usepackage{comment}
\usepackage{amsmath,amssymb} %
\usepackage{color}
\usepackage{multirow}
\usepackage{booktabs}
\usepackage[accsupp]{axessibility}  %

\usepackage{xspace}
\usepackage[numbers,sort,compress]{natbib}
\usepackage{xcolor}         %
\definecolor{ForestGreen}{RGB}{34,139,34}

\usepackage[pagebackref=false,breaklinks=true,letterpaper=true,colorlinks,bookmarks=false,citecolor=ForestGreen]{hyperref} %

\usepackage{url}            %
\usepackage{amsfonts}       %
\usepackage{nicefrac}       %
\usepackage{microtype}      %
\usepackage[labelfont=bf,font=small,skip=5pt]{caption}
\usepackage{bbm}
\usepackage{mathtools}
\usepackage{adjustbox}
\usepackage{wrapfig}
\usepackage[symbol]{footmisc}

\usepackage[normalem]{ulem}
\useunder{\uline}{\ul}{}
\usepackage[linesnumbered,ruled,vlined]{algorithm2e}

\SetKwInput{KwInput}{Input}                %
\SetKwInput{KwOutput}{Output}              %

\makeatletter
\def\NAT@def@citea{\def\@citea{\NAT@separator}}
\makeatother

\makeatletter
\DeclareRobustCommand\onedot{\futurelet\@let@token\@onedot}
\def\@onedot{\ifx\@let@token.\else.\null\fi\xspace}

\def\eg{\emph{e.g}\onedot} 
\def\ie{\emph{i.e}\onedot} 
 
\def\etc{\emph{etc}\onedot} 
\def\wrt{w.r.t\onedot} 
 
\def\etal{\emph{et al}\onedot}
\makeatother

\usepackage[capitalize]{cleveref}
\crefname{section}{Sec.}{Secs.}
\Crefname{section}{Section}{Sections}
\Crefname{table}{Table}{Tables}
\crefname{table}{Tab.}{Tabs.}

\begin{document}
\pagestyle{headings}
\mainmatter
\def\ECCVSubNumber{1018}  %

\title{Discover and Mitigate Unknown Biases with Debiasing Alternate Networks} %

\titlerunning{Discover and Mitigate Unknown Biases with Debiasing Alternate Networks}
\author{Zhiheng Li$^1$, Anthony Hoogs$^2$, and Chenliang Xu$^1$}
\authorrunning{Li et al.}
\institute{
  $^1$University of Rochester \enspace $^2$Kitware, Inc. \\
  \email{\{zhiheng.li,chenliang.xu\}@rochester.edu} \enspace \email{anthony.hoogs@kitware.com}
}
\maketitle

\begin{abstract}
  Deep image classifiers have been found to learn biases from datasets. To mitigate the biases, most previous methods require labels of protected attributes (e.g., age, skin tone) as full-supervision, which has two limitations: 1) it is infeasible when the labels are unavailable; 2) they are incapable of mitigating unknown biases---biases that humans do not preconceive. To resolve those problems, we propose Debiasing Alternate Networks (DebiAN), which comprises two networks---a Discoverer and a Classifier. By training in an alternate manner, the discoverer tries to find multiple unknown biases of the classifier without any annotations of biases, and the classifier aims at unlearning the biases identified by the discoverer. While previous works evaluate debiasing results in terms of a single bias, we create Multi-Color MNIST dataset to better benchmark mitigation of multiple biases in a multi-bias setting, which not only reveals the problems in previous methods but also demonstrates the advantage of DebiAN in identifying and mitigating multiple biases simultaneously. We further conduct extensive experiments on real-world datasets, showing that the discoverer in DebiAN can identify unknown biases that may be hard to be found by humans. Regarding debiasing, DebiAN achieves strong bias mitigation performance.
\end{abstract}

\section{Introduction}
\label{sec.intro}

Many studies have verified that AI algorithms learn undesirable biases from the dataset. Some biases provide shortcuts~\cite{geirhos2020Nat.Mach.Intell.} for the network to learn superficial features instead of the intended decision rule causing robustness issues, \eg, static cues for action recognition~\cite{choi2019Adv.NeuralInf.Process.Syst.,li2018Eur.Conf.Comput.Vis.ECCV,bahng2020Int.Conf.Mach.Learn.}. Other biases make AI algorithms discriminate against different protected demographic groups such as genders\footnote[1]{In this work, ``gender'' denotes visually perceived gender, not real gender identity.}~\cite{hendricks2018Eur.Conf.Comput.Vis.ECCV,wang2019IEEEInt.Conf.Comput.Vis.ICCVa,zhao2017Empir.MethodsNat.Lang.Process.,albiero2020IEEEWinterConf.Appl.Comput.Vis.WorkshopWACVW,wang2021Empir.MethodsNat.Lang.Process.,joo2020Int.WorkshopFairnessAccount.Transpar.EthicsMultimed.,jia2020Annu.Meet.Assoc.Comput.Linguist.} and skin tones~\cite{buolamwini2018ACMConf.FairnessAccount.Transpar.,hazirbas2021ArXiv210402821Cs}, leading to serious fairness problems.
Therefore, it is imperative to mitigate the biases in AI algorithms. However, most previous bias mitigation methods~\cite{wang2020IEEEConf.Comput.Vis.PatternRecognit.CVPRe,alvi2018Eur.Conf.Comput.Vis.WorkshopECCVW,tzeng2015IEEEInt.Conf.Comput.Vis.ICCV,zhang2018AAAIACMConf.AIEthicsSoc.,zhao2017Empir.MethodsNat.Lang.Process.} are supervised methods---requiring annotations of the biases, which has several limitations: First, bias mitigation cannot be performed when labels are not available due to privacy concerns. Second, they cannot mitigate \textit{unknown} biases---biases that humans did not preconceive, making the biases impossible to be labeled and mitigated.

Since supervised debiasing methods present many disadvantages, in this work, we focus on a more challenging task---unsupervised debiasing, which mitigates the \textit{unknown} biases in a learned classifier without any annotations. Without loss of generality, we focus on mitigating biases in image classifiers. Solving this problem contains two steps~\cite{nam2020Adv.NeuralInf.Process.Syst.,sohoni2020Adv.NeuralInf.Process.Syst.,lahoti2020Adv.NeuralInf.Process.Syst.,creager2021Int.Conf.Mach.Learn.,ahmed2021Int.Conf.Learn.Represent.}: bias identification and bias mitigation.

\begin{wrapfigure}[15]{r}{0.4\textwidth}
  \centering
    \includegraphics[width=\linewidth]{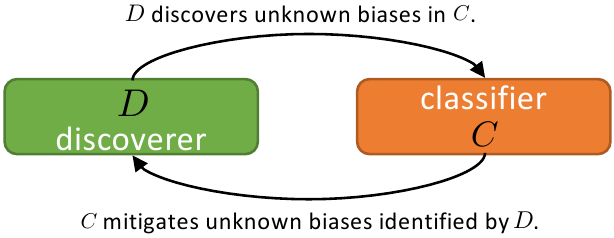}
    \caption{\textsc{Debiasing Alternate Networks} (DebiAN). We alternately train two networks---a \textit{discover} and a \textit{classifier}. \textit{Discoverer} actively identifies \textit{classifier}'s unknown biases. At the same time, the \textit{classifier} mitigates the biases identified by the \textit{discoverer}
}
  \label{fig.teaser}
\end{wrapfigure}

Due to the absence of bias annotations, the first step is to assign the training samples into different bias groups as the pseudo bias labels, which is challenging since the biases are even unknown. The crux of the problem is to define the unknown bias. Some previous works make strong assumptions about the unknown biases based on empirical observations, such as biases are easier to be learned~\cite{nam2020Adv.NeuralInf.Process.Syst.}, samples from the same bias group are clustered in feature space~\cite{sohoni2020Adv.NeuralInf.Process.Syst.}, which can be tenuous for different datasets or networks. Other works quantify the unknown biases by inversely using the debiasing objective functions~\cite{creager2021Int.Conf.Mach.Learn.,ahmed2021Int.Conf.Learn.Represent.}, which can face numerical or convergence problems (more details in \cref{sec.related_work}). Unlike previous works, we follow an axiomatic principle to define the unknown biases---classifier's predictions that violate a fairness criterion~\cite{hardt2016Adv.NeuralInf.Process.Syst.,pleiss2017Adv.NeuralInf.Process.Syst.,kusner2017Adv.NeuralInf.Process.Syst.,dwork2012Proc.3rdInnov.Theor.Comput.Sci.Conf.,grgic-hlaca2016NIPSSymp.Mach.Learn.Law,verma20182018IEEEACMInt.WorkshopSoftw.FairnessFairWare,corbett-davies2017Proc.23rdACMSIGKDDInt.Conf.Knowl.Discov.DataMin.}. Based on this definition, we propose a novel \textit{Equal Opportunity Violation} (EOV) loss to train a \textit{discoverer} network to identify the classifier’s biases. In specific, it shepherds the \textit{discoverer} network to predict bias group assignments such that the \textit{classifier} violates the Equal Opportunity~\cite{hardt2016Adv.NeuralInf.Process.Syst.,pleiss2017Adv.NeuralInf.Process.Syst.} fairness criterion
(Figs.~\ref{fig.teaser}, \ref{fig.method_overview}).

Regarding debiasing as the second step, most previous approaches~\cite{sohoni2020Adv.NeuralInf.Process.Syst.,creager2021Int.Conf.Mach.Learn.,ahmed2021Int.Conf.Learn.Represent.} preprocess the identified biases into pseudo bias labels and resort to other supervised bias mitigation methods~\cite{arjovsky2020ArXiv190702893CsStat,sagawa*2020Int.Conf.Learn.Represent.} for debiasing. In contrast, we propose a novel \textit{Reweighted Cross-Entropy} (RCE) loss that leverages soft bias group assignments predicted by the \textit{discoverer} network to mitigate the biases in the \textit{classifier} (\cref{fig.teaser}). In this way, the \textit{classifier} is guided to meet the Equal Opportunity.

In addition, many previous works~\cite{sohoni2020Adv.NeuralInf.Process.Syst.,creager2021Int.Conf.Mach.Learn.,ahmed2021Int.Conf.Learn.Represent.} treat bias identification and bias mitigation as two isolated steps. In~\cite{creager2021Int.Conf.Mach.Learn.,ahmed2021Int.Conf.Learn.Represent.}, the biases are identified from an undertrained classifier, which is suboptimal since the classifier may learn different biases at different training stages. Consequently, these two-stage methods fail to mitigate other biases learned by the classifier at later training stages. In contrast, we employ an alternate training scheme to carry out bias identification and bias mitigation simultaneously. We jointly update the \textit{discoverer} and \textit{classifier} in an interleaving fashion (\cref{fig.teaser,fig.method_overview}). In this way, the \textit{discoverer} can repetitively inspect multiple biases that the \textit{classifier} learns at the entire training stage.

We integrate our novel losses and training scheme into a unified framework---\textsc{Debiasing Alternate Networks} (DebiAN), which contains two networks---a \textit{discoverer} $D$ and a \textit{classifier} $C$ (see Fig.~\ref{fig.teaser}). We jointly train the two networks in an alternate manner. Supervised by our novel EOV loss, $D$ tries to discover $C$'s multiple unknown biases that violate the Equal Opportunity fairness criterion. Trained with our RCE loss, $C$ aims at mitigating multiple biases identified by the \textit{discoverer} $D$ to satisfy Equal Opportunity. After the alternate training, the unknown biases in \textit{classifier} $C$ are mitigated, leading to a fairer and more robust classification model. Besides, when employed with other network explanation methods~\cite{zhou2016IEEEConf.Comput.Vis.PatternRecognit.CVPR,selvaraju2017IEEEInt.Conf.Comput.Vis.ICCV,selvaraju2019Int.J.Comput.Vis.}, the \textit{discoverer} is helpful to interpret the discovered unknown biases, facilitating dataset curators to locate dataset biases~\cite{torralba2011IEEEConf.Comput.Vis.PatternRecognit.CVPR}.

While previous works~\cite{sagawa*2020Int.Conf.Learn.Represent.,nam2020Adv.NeuralInf.Process.Syst.,kim2019IEEEConf.Comput.Vis.PatternRecognit.CVPR,li2019IEEEConf.Comput.Vis.PatternRecognit.CVPRa,bahng2020Int.Conf.Mach.Learn.} only evaluate debiasing results in terms of a single bias, we create Multi-Color MNIST dataset with two biases in the dataset, which benchmarks debiasing algorithms in the multi-bias setting. Our new dataset surfaces the problems in previous methods (\eg, LfF~\cite{nam2020Adv.NeuralInf.Process.Syst.}) and demonstrates the advantage of DebiAN in discovering and mitigating multiple biases. We further conduct extensive experiments to verify the efficacy of DebiAN in real-world image datasets. In the face image domain, DebiAN achieves better gender bias mitigation results on CelebA~\cite{liu2015IEEEInt.Conf.Comput.Vis.ICCV} and bFFHQ~\cite{kim2021IEEEInt.Conf.Comput.Vis.ICCVa} datasets. On the gender classification task, DebiAN achieves better debiasing results on CelebA \wrt multiple bias attributes.
We further show an interesting unknown bias discovered by DebiAN in gender classification---\texttt{visible hair area}. Lastly, we show that DebiAN applies to other image domains for broader tasks, such as action recognition and scene classification. Our method not only achieves better debiasing results, but also identifies interesting unknown biases in scene classifiers.

Our contributions are summarized as follows: (1) We propose a novel objective function, \textit{Equal Opportunity Violation} (EOV) loss, for identifying unknown biases of a classifier based on Equal Opportunity. (2) We propose a \textit{Reweighted Cross-Entropy} (RCE) loss to mitigate the discovered unknown biases by leveraging the soft bias group assignments. (3) We create Multi-Color MNIST dataset to benchmark debiasing algorithms in a multi-bias setting. (4) Our \textsc{Debiasing Alternate Networks} (DebiAN) outperforms previous unsupervised debiasing methods on both synthetic and real-world datasets.

\section{Related Work}
\label{sec.related_work}

\noindent \textbf{Bias Identification} \enspace Most previous works identify \textit{known} biases based on bias labels. In~\cite{buolamwini2018ACMConf.FairnessAccount.Transpar.}, face images are labeled with gender and skin tone to identify the performance gaps across intersectional groups. \citet{balakrishnan2020Eur.Conf.Comput.Vis.ECCV} further synthesize intersectional groups of images and analyze the biases with additional labels. Beyond face images, recent works~\cite{manjunatha2019IEEEConf.Comput.Vis.PatternRecognit.CVPR,wang2020Eur.Conf.Comput.Vis.ECCVd} compute the statistics of labels based on the rule mining algorithm~\cite{agrawal1994Int.Conf.VeryLargeDataBases} or external tools. \cite{krishnakumar2021Br.Mach.Vis.Conf.BMVC} uses clustering on image embeddings to discover unknown biases. \cite{lang2021IEEEInt.Conf.Comput.Vis.ICCV,li2021IEEEInt.Conf.Comput.Vis.ICCV} discovers \textit{unknown} biases without labels. However, these works rely on GAN~\cite{goodfellow2014Adv.NeuralInf.Process.Syst.,karras2019IEEEConf.Comput.Vis.PatternRecognit.CVPR} to synthesize images, which suffers from image quality issues. In contrast, DebiAN directly classifies real images into different bias attribute groups to discover the unknown biases.

\noindent \textbf{Supervised Debiasing} \enspace Supervised debiasing methods use bias labels for debiasing. \cite{kamiran2012KnowlInfSyst} proposes a supervised reweighing method. \citet{wang2020IEEEConf.Comput.Vis.PatternRecognit.CVPR} benchmark recent supervised debiasing methods~\cite{alvi2018Eur.Conf.Comput.Vis.WorkshopECCVW,tzeng2015IEEEInt.Conf.Comput.Vis.ICCV,zhang2018AAAIACMConf.AIEthicsSoc.,zhao2017Empir.MethodsNat.Lang.Process.}. \cite{creager2019Int.Conf.Mach.Learn.} lets the model be flexibly fair to different attributes during testing. \cite{sarhan2020Eur.Conf.Comput.Vis.ECCV} uses disentanglement for debiasing. \citet{singh2020IEEEConf.Comput.Vis.PatternRecognit.CVPR} propose a feature splitting approach to mitigate contextual bias. \cite{dhar2021IEEEInt.Conf.Comput.Vis.ICCV,gong2020Eur.Conf.Comput.Vis.ECCV} use adversarial training to mitigate biases in face recognition.

\noindent \textbf{Known Bias Mitigation with Prior knowledge} \enspace Without using labels, some works use prior knowledge to mitigate certain known biases. ReBias~\cite{bahng2020Int.Conf.Mach.Learn.} uses model capacity as the inductive bias to mitigate texture bias and static bias in image and video classification. HEX~\cite{wang2019Int.Conf.Learn.Represent.b} introduces a texture extractor to mitigate the texture bias. Beyond image classification, RUBi~\cite{cadene2019Adv.NeuralInf.Process.Syst.} and LearnedMixin~\cite{clark2019Empir.MethodsNat.Lang.Process.} mitigate unimodal bias for visual question answering~\cite{antol2015IEEEInt.Conf.Comput.Vis.ICCV} with prior knowledge.

\noindent \textbf{Unsupervised Debiasing} \enspace In the field of mitigating unknown biases, \citet{sohoni2020Adv.NeuralInf.Process.Syst.} apply clustering on samples in each class and use the clustering assignment as the predicted bias labels, which could be inaccurate due to its unsupervised nature. Li \etal~\cite{li2018Eur.Conf.Comput.Vis.ECCV,li2019IEEEConf.Comput.Vis.PatternRecognit.CVPRd} fix the parameters of feature extractors and focus on mitigating the representation bias. PI~\cite{bao2021Int.Conf.Mach.Learn.Predict} uses labels of one known bias attribute to interpolate environments, improving robustness against multiple unknown biases. TOFU~\cite{bao2022Proc.39thInt.Conf.Mach.Learn.Learning} infers unstable features from a source task and transfers them to the target task. Concurrent to our work, \citet{bao2022ArXiv} propose learning to split for detecting unknown biases, which can then be combined with GroupDRO~\cite{sagawa*2020Int.Conf.Learn.Represent.} for debiasing. LfF~\cite{nam2020Adv.NeuralInf.Process.Syst.} identifies biases by finding easier samples in the training data through training a bias-amplified network supervised by GCE loss~\cite{zhang2018Adv.NeuralInf.Process.Syst.}, which up-weights the samples with smaller loss values and down-weights the samples with larger loss values. In other words, GCE loss does not consider the information of the classifier, \eg, the classifier's output. Therefore, LfF's bias-amplified network blindly finds the biases in the data samples instead of the classifier. Unlike LfF, the EOV loss in DebiAN actively identifies biases in the classifier based on the classifier's predictions, leading to better debiasing performance. Following LfF, BiaSwap~\cite{kim2021IEEEInt.Conf.Comput.Vis.ICCVa} uses LfF to discover biases and generate more underrepesented images via style-transfer for training. Other works \cite{lahoti2020Adv.NeuralInf.Process.Syst.,wang2021Adv.NeuralInf.Process.Syst.SelfSupervised,ahmed2021Int.Conf.Learn.Represent.,creager2021Int.Conf.Mach.Learn.} inversely use the debiasing objective function to maximize an unbounded loss (\eg, gradient norm penalty in IRMv1~\cite{arjovsky2020ArXiv190702893CsStat}) for bias identification, which may encounter numerical or convergence problems. As a comparison, our EOV loss (\cref{eq.equal_opportunity_violation_loss}) minimizes negative log-likelihood, which is numerically stable and easier to converge.

\section{Method}

\begin{figure}
  \begin{center}
    \includegraphics[width=0.9\linewidth]{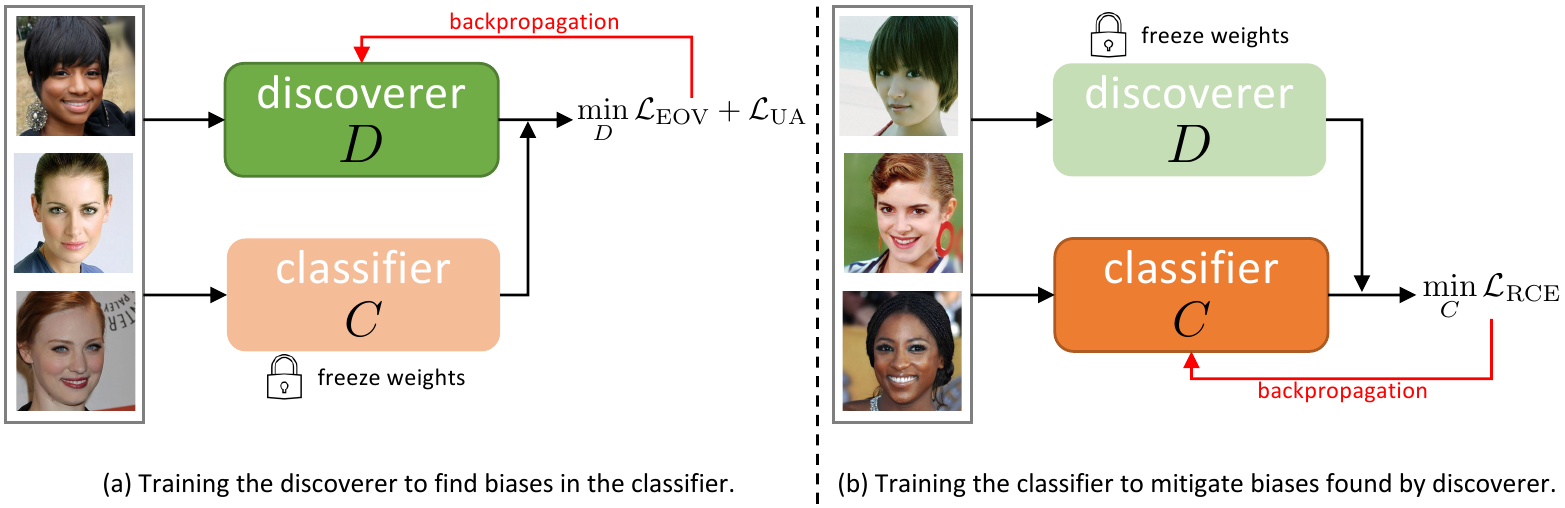}
  \end{center}
  \caption{Overview of \textsc{Debiasing Alternate Networks} (DebiAN). DebiAN consists of two networks---a \textit{discoverer} $D$ and a \textit{classifier} $C$. $D$ is trained with $\mathcal{L}_\text{EOV}$ and $\mathcal{L}_\text{UA}$ (Sec.~\ref{subsec.method_unknown_bias_discovery}) to find the unknown biases in $C$. $C$ is optimized with $\mathcal{L}_{RCE}$ (Sec.~\ref{subsec.method_unknown_bias_mitigation}) to mitigate the biases identified by $D$}
  \label{fig.method_overview}
\end{figure}

\noindent \textbf{Overview} \enspace The overview of our proposed \textsc{Debiasing Alternate Networks} (DebiAN) is shown in Fig.~\ref{fig.method_overview}. It contains two networks---a \textit{discoverer} $D$ and a \textit{classifier} $C$. As shown in Fig.~\ref{fig.method_overview} (a), the \textit{discoverer} $D$ tries to discover the unknown biases in the \textit{classifier} $C$ by optimizing our proposed EOV loss ($\mathcal{L}_\text{EOV}$) and UA penalty ($\mathcal{L}_\text{UA}$) (Sec.~\ref{subsec.method_unknown_bias_discovery}). As shown in Fig.~\ref{fig.method_overview} (b), the \textit{classifier} $C$'s goal is to mitigate the biases identified by $D$ via a novel \textit{Reweighted Cross-Entropy} loss ($\mathcal{L}_\text{RCE}$) (Sec.~\ref{subsec.method_unknown_bias_mitigation}). Lastly, we train the two networks in an alternate manner as the full model for discovering and mitigating the unknown biases (Sec.~\ref{subsec.method_full_model}).

\begin{figure}[t]
  \begin{center}
    \includegraphics[width=0.9\linewidth]{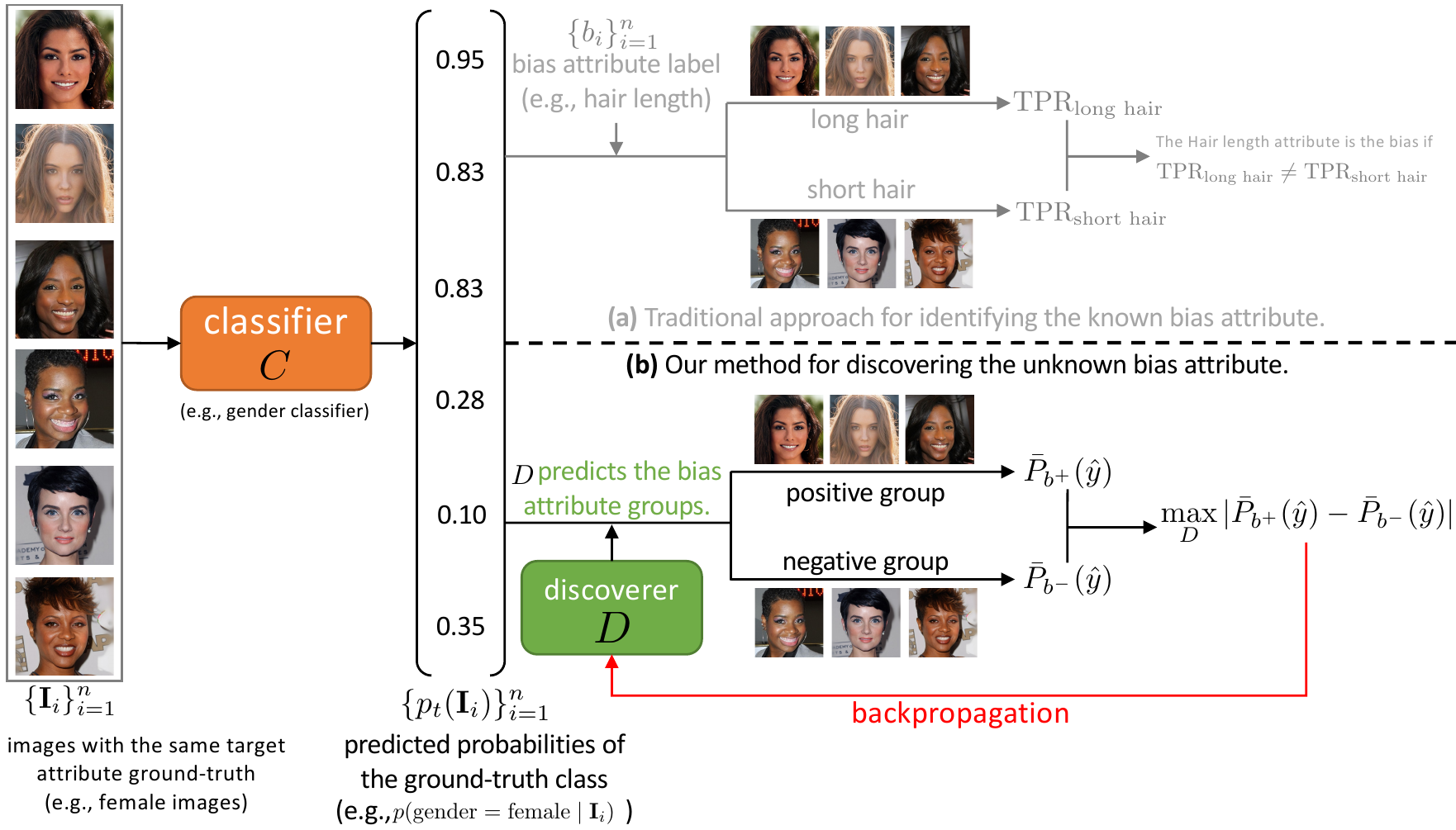}
  \end{center}
    \caption{\textbf{(a)}: The traditional approach for identifying the \textit{known} bias attribute (\eg, hair length) by comparing true positive rates (TPR) of the target attribute (\eg, gender) in two groups of bias attributes (\eg, long hair and short hair), where the group assignment of bias attribute is based on the labels of the bias attribute. \textbf{(b)}: Our method trains a \textit{discoverer} ($D$) to predict the groups of the unknown bias attribute such that the difference of averaged predicted probabilities on the target attribute (\eg, gender) in two groups are maximized (see Eq.~\ref{eq.equal_opportunity_violation_loss})}
  \label{fig.compare_known_vs_unknown_bias_id}
\end{figure}

\noindent \textbf{Background} \enspace To better explain our motivation for discovering the \textit{unknown} biases (without manual annotations of biases), let us first revisit the traditional approach for identifying \textit{known} biases when labels of biases (\eg, protected attributes) are available, which is illustrated in Fig.~\ref{fig.compare_known_vs_unknown_bias_id} (a). The following are given for identifying \textit{known} biases---a well-trained \textit{classifier} $C$ for predicting a target attribute, $n$ testing images $\{\mathbf{I}_i\}_{i=1}^n$, target attribute labels of each image $\{y_i\}_{i=1}^n$, and bias attribute labels $\{b_i\}_{i=1}^n$.
We denote the $i$-th image target attribute as $y_i \in \{1, 2, ... K\}$ and $K$ is the number of classes.
We consider the bias attribute that is binary or continuously valued (\ie, $b_i \in \{0, 1\}$ or $b_i \in [0, 1]$), such as biological gender (\eg, female and male) and skin tones (\eg, from dark skin tones to light skin tones in Fitzpatrick skin type scale). We leave bias attributes with multi-class values for future works. Then, the given \textit{classifier} $C$ is tested for predicting the target attribute $\hat{y}_i$ for each testing image $\mathbf{I}_i$. Finally, we check whether the predictions meet a fairness criterion, such as Equal Opportunity~\cite{hardt2016Adv.NeuralInf.Process.Syst.}:
\begin{equation}
  \text{Pr}\{ \hat{y} = k \mid b = 0, y = k \} = \text{Pr}\{ \hat{y} = k \mid b = 1, y = k \},
  \label{eq.equal_opportunity}
\end{equation}
where the LHS and RHS are true positive rates (TPR) in negative ($b=0$) and positive ($b=1$) bias attribute groups, respectively. $k \in \{1 ... K\}$ is a target attribute class. Equal Opportunity requires the same TPR across two different bias attribute groups. That is, if the TPR is significantly different in two groups of the bias attribute, we conclude that \textit{classifier} $C$ contains the bias of attribute $b$ because $C$ violates the Equal Opportunity fairness criterion. For example, as shown in Fig.~\ref{fig.compare_known_vs_unknown_bias_id} (a), although all images are female, a gender classifier may have a larger TPR for the group of long-hair female images than the group of short-hair female images. Thus the gender classifier is biased against different hair lengths.

\subsection{Unknown Bias Discovery}
\label{subsec.method_unknown_bias_discovery}
As for identifying unknown biases, we do not have the labels to assign images into two groups for comparing TPR since 1) we do not assume images come with bias attribute labels, and 2) the type of bias is even unknown.
However, we can compare the difference in TPR for any group assignments based on speculated biases---a significant difference in TPR hints that the Equal Opportunity fairness criterion is most violated (our method mainly focuses on the Equal Opportunity fairness criterion, and we leave other fairness criteria for future work).
Motivated by this finding, instead of using labels of bias attribute $\{b_i\}_{i=1}^n$ for group assignment, we train a \textit{discoverer} $D$ to predict the group assignment for each image, \ie, $p(\hat{b} \mid \mathbf{I}_i) \coloneqq D(\mathbf{I}_i)$. By optimizing loss functions below, we find the most salient bias of the \textit{classifier} $C$ that violates the Equal Opportunity fairness criterion, which is illustrated in Fig.~\ref{fig.compare_known_vs_unknown_bias_id} (b).

\noindent \textbf{Equal Opportunity Violation (EOV) Loss} \enspace To shepherd the \textit{discoverer} $D$ to find the group assignment where \textit{classifier} $C$ violates the Equal Opportunity fairness criterion, we propose the \textit{Equal Opportunity Violation} (EOV) loss, denoted by $\mathcal{L}_\text{EOV}$, as the objective function to train $D$. For computing $\mathcal{L}_\text{EOV}$, we sample a set of $n$ images $\{\mathbf{I}_i\}_{i=1}^n$ with the \textit{same} target attribute labels (\ie, $\forall_i y_i = k$), \eg, all images in Fig.~\ref{fig.compare_known_vs_unknown_bias_id} (b) are female. The \textit{classifier} $C$ has been trained for predicting the target attribute $y$ of the images (\ie, $p(\hat{y} \mid \mathbf{I}_i) \coloneqq C(\mathbf{I}_i)$). For simplicity, we denote $p_t$ as $C$'s prediction on images of the ground-truth class (\ie, $p_t(\mathbf{I}_i) = p(\hat{y} = y_i \mid \mathbf{I}_i)$). Meanwhile, the same set of images $\{ \mathbf{I}_i \}$ are fed to the \textit{discoverer} $D$ for predicting the binary bias attribute group assignment: $p(\hat{b} \mid \mathbf{I}_i) \coloneqq D(\mathbf{I}_i)$. Finally, we define the EOV loss as:
\begin{equation}
    \mathcal{L}_\text{EOV} = -\log \left( \left| \bar{P}_{b^+}(\hat{y}) - \bar{P}_{b^-}(\hat{y}) \right| \right),
    \label{eq.equal_opportunity_violation_loss}
\end{equation}
where $\bar{P}_{b^+}(\hat{y})$ and $\bar{P}_{b^-}(\hat{y})$ are defined by:
\begin{align}
\begin{split}
      &\bar{P}_{b^+}(\hat{y}) = \frac{\sum_{i=1}^n p(\hat{b} = 1 \mid \mathbf{I}_i) p_t(\mathbf{I}_i)}{ \sum_{i=1}^n p(\hat{b} = 1 \mid \mathbf{I}_i)},\\
    &\bar{P}_{b^-}(\hat{y}) = \frac{\sum_{i=1}^n p(\hat{b} = 0 \mid \mathbf{I}_i) p_t(\mathbf{I}_i)}{ \sum_{i=1}^n p(\hat{b} = 0 \mid \mathbf{I}_i)} .
\end{split}
\label{eq.weighted_predicted_prob}
\end{align}
Intuitively, $\bar{P}_{b^+}(\hat{y})$ and $\bar{P}_{b^-}(\hat{y})$
are the weighted average predicted probabilities of the target attribute in two bias attribute groups, which can be regarded as a relaxation to Equal Opportunity's true positive rate (Eq.~\ref{eq.equal_opportunity}) where the predicted probabilities are binarized into predictions with a threshold (\eg, 0.5). Minimizing $\mathcal{L}_\text{EOV}$ leads $D$ to maximize the discrepancy of averaged predicted probabilities of target attributes in two bias attribute groups (\ie, see $\max_{D} |\bar{P}_{b^+}(\hat{y}) - \bar{P}_{b^-}(\hat{y})| $ in Fig.~\ref{fig.compare_known_vs_unknown_bias_id}), thus finding the bias attribute group assignments where $C$ violates the Equal Opportunity fairness criterion. For example, in Fig.~\ref{fig.compare_known_vs_unknown_bias_id} (b), if the gender classifier $C$ is biased against different hair lengths, then by optimizing $\mathcal{L}_\text{EOV}$, $D$ can assign the female images into two bias attribute groups (\ie, short hair and long hair) with the predicted bias attribute group assignment probability $p(\hat{b} \mid \mathbf{I}_i)$, such that the difference of averaged predicted probabilities on gender in these two groups is maximized.

\noindent \textbf{Unbalanced Assignment (UA) penalty} \enspace However, we find that optimizing $\mathcal{L}_\text{EOV}$ alone may let the \textit{discoverer} $D$ find a trivial solution---assigning all images into one bias attribute group. For example, suppose $D$ assigns all images to the positive bias attribute group (\ie, $\forall_i, \: p(\hat{b} = 1 \mid \mathbf{I}_i) = 1$). In that case, $\bar{P}_{b^-}(\hat{y})$ becomes zero since the negative group contains no images. $\bar{P}_{b^+}(\hat{y})$ becomes a large positive number by simply averaging $p_t(\mathbf{I}_i)$ for all of the $n$ images, which can trivially increase $|\bar{P}_{b^+}(\hat{y}) - \bar{P}_{b^-}(\hat{y})|$, leading to a small $\mathcal{L}_\text{EOV}$. To prevent this trivial solution, we propose the \textit{Unbalanced Assignment} (UA) loss denoted by:
\begin{equation}
    \mathcal{L}_\text{UA} = -\log \left( 1 - \frac{1}{n} \left|  \sum_{i=1}^n p(\hat{b} = 1 \mid \mathbf{I}_i) - p(\hat{b} = 0 \mid \mathbf{I}_i) \right| \right).
    \label{eq.unbalanced_assignment_penalty}
\end{equation}
Intuitively, minimizing $\mathcal{L}_\text{UA}$ penalizes the unbalanced assignment that leads to large difference between $\sum_{i=1}^n p(\hat{b} = 1 \mid \mathbf{I}_i)$ and $\sum_{i=1}^n p(\hat{b} = 0 \mid \mathbf{I}_i)$, which can be regarded as the numbers of images assigned into positive and negative bias attribute groups, respectively. Therefore, $\mathcal{L}_\text{EOV}$ is jointly optimized with $\mathcal{L}_\text{UA}$ to prevent the trivial solution. We acknowledge a limitation of the UA penalty. Although it resolves the trivial solution, it introduces a trade-off since the bias attribute groups are usually spuriously correlated with the target attribute (\eg, more long-hair females than the short-hair females in the dataset). Hence encouraging balanced assignments may make the \textit{discoverer} harder to find the correct assignment. However, our ablation study shows that the benefits of using $\mathcal{L}_\text{UA}$ outweigh its limitations. The results are shown in \cref{subsec.exp_synthetic} and \cref{tab.multi_color_mnist_results}.

\subsection{Unknown Bias Mitigation by Reweighing}
\label{subsec.method_unknown_bias_mitigation}
We further mitigate $C$'s unknown biases identified by $D$. To this end, we propose a novel \textit{Reweighted Cross-Entropy} loss that adjusts the weight of each image's classification loss. Based on the bias attribute group assignment $p(\hat{b} \mid \mathbf{I}_i)$ predicted by $D$, we define the weight $\mathcal{W}(\mathbf{I}_i)$ of classification loss for each image $\mathbf{I}_i$ as:
\begin{align}
\begin{split}
        \mathcal{W}(\mathbf{I}_i) = &\mathbbm{1}\left[ \bar{P}_{b^+}(\hat{y}) \geq \bar{P}_{b^-}(\hat{y}) \right] p(\hat{b} = 0 \mid \mathbf{I}_i) \\
                                & + \mathbbm{1}\left[ \bar{P}_{b^+}(\hat{y}) < \bar{P}_{b^-}(\hat{y}) \right] p(\hat{b} = 1 \mid \mathbf{I}_i),
\end{split}
\end{align}
where $\mathbbm{1}$ is an indicator function. Then, the \textit{Reweighted Cross-Entropy} loss ($\mathcal{L}_\text{RCE}$) is defined by:
\begin{equation}
    \mathcal{L}_\text{RCE} = - \frac{1}{n} \sum_{i=1}^n \left(1 + \mathcal{W}\left(\mathbf{I}_i\right) \right) \log p_t\left( \mathbf{I}_i \right).
    \label{eq.rce}
\end{equation}

For example, when $C$ performs better on images from the positive bias attribute group (\ie, $\bar{P}_{b^+}(\hat{y}) \geq \bar{P}_{b^-}(\hat{y})$), we use $p(\hat{b} = 0 \mid \mathbf{I}_i)$ as the weight, which up-weights the images from the negative bias attribute group, where classifier $C$ is worse-performed. At the same time, it down-weights the images from the positive bias attribute group where $C$ is already better-performed. Adding one to the weight in \cref{eq.rce} lets the loss function degenerate to standard cross-entropy loss when $\mathcal{W}(\mathbf{I}_i) = 0$. By minimizing the \textit{Reweighted Cross-Entropy} loss, $C$ is guided to meet Equal Opportunity.

\subsection{Full Model}
\label{subsec.method_full_model}
We summarize the proposed losses in Sec.~\ref{subsec.method_unknown_bias_discovery} and Sec.~\ref{subsec.method_unknown_bias_mitigation} for the full model of \textsc{Debiasing Alternate Networks} (DebiAN), which is shown in Fig.~\ref{fig.method_overview}. When the task is to only discover (\ie, not mitigate) the unknown biases of a given classifier, the classifier's parameters are fixed and we only train the \textit{discoverer} $D$ by minimizing $\mathcal{L}_\text{EOV}$ (Eq.~\ref{eq.equal_opportunity_violation_loss}) and $\mathcal{L}_\text{UA}$ (Eq.~\ref{eq.unbalanced_assignment_penalty}) on the classifier's training data. When the task is to mitigate the unknown biases, we jointly train two networks in an alternate fashion:
\begin{align}
    &\min_{D} \mathcal{L}_\text{EOV} + \mathcal{L}_\text{UA} \label{eq.bias_discovery_obj}, \\
    &\min_{C} \mathcal{L}_\text{RCE} \label{eq.cls_obj}.
\end{align}
In Eq.~\ref{eq.bias_discovery_obj}, $C$'s parameters are fixed, and $D$ is optimized to identify $C$'s unknown biases where $C$ violates the Equal Opportunity. Through Eq.~\ref{eq.cls_obj}, $C$ is optimized for mitigating the unknown biases discovered by $D$ to satisfy the Equal Opportunity while $D$'s parameters are frozen. After the alternate training, $C$'s unknown biases identified by $D$ are mitigated, leading to a fairer and more robust \textit{classifier}. The pseudocode of the complete algorithm is in \cref{sub.supp.pseudo_code}.

\section{Experiment}
\label{sec.experiment}

We conduct extensive experiments to verify the efficacy of DebiAN. First, we evaluate the results on our newly created Multi-Color MNIST dataset (\cref{subsec.exp_synthetic}) in a multi-bias setting. We further conduct experiments on real-world datasets in multiple image domains--face (\cref{subsec.exp_face}) and other image domains (\eg, scene, action recognition) (\cref{subsec.exp_scene}). More details (\eg, evaluation metrics) are introduced in each subsection. The code and our newly created Multi-Color MNIST dataset are released at \url{https://github.com/zhihengli-UR/DebiAN}.

\noindent \textbf{Comparison Methods} \enspace We mainly compare with three unsupervised debiasing methods: 1) LfF~\cite{nam2020Adv.NeuralInf.Process.Syst.} uses Generalized Cross-entropy (GCE) loss~\cite{zhang2018Adv.NeuralInf.Process.Syst.} to train a ``biased model'' for reweighing the classifier; 2) EIIL~\cite{creager2021Int.Conf.Mach.Learn.} identifies the bias groups by optimizing bias group assignment to maximize the IRMv1~\cite{arjovsky2020ArXiv190702893CsStat} objective function. The identified bias groups will serve as pseudo bias labels for other supervised debiasing methods to mitigate the biases. Following \cite{creager2021Int.Conf.Mach.Learn.},  IRM~\cite{arjovsky2020ArXiv190702893CsStat} is used as the debiasing algorithm for EIIL. 3) PGI~\cite{ahmed2021Int.Conf.Learn.Represent.} follows EIIL to identify the biases by training a small multi-layer perceptron for bias label predictions. Concerning debiasing, PGI minimizes the KL-divergence of the classifier's predictions across different bias groups. We use the officially released code of LfF, EIIL, and PGI in our experiment. Besides, we also compare with vanilla models, which do not have any debiasing techniques (\ie, only using standard cross-entropy loss for training). On bFFHQ~\cite{kim2021IEEEInt.Conf.Comput.Vis.ICCVa} and BAR~\cite{nam2020Adv.NeuralInf.Process.Syst.} datasets, we also compare with BiaSwap~\cite{kim2021IEEEInt.Conf.Comput.Vis.ICCVa}, which follows LfF to identify unknown biases, and then uses style-transfer to generate more underrepresented images for training. Since its code has not been released, we cannot compare DebiAN with BiaSwap on other datasets. All results shown below are the mean results over three random seeds of runs, and we also report the standard deviation as the error bar.

\begin{figure}[t]
  \centering
  \begin{minipage}{.35\textwidth}
    \centering
    \includegraphics[width=0.9\linewidth]{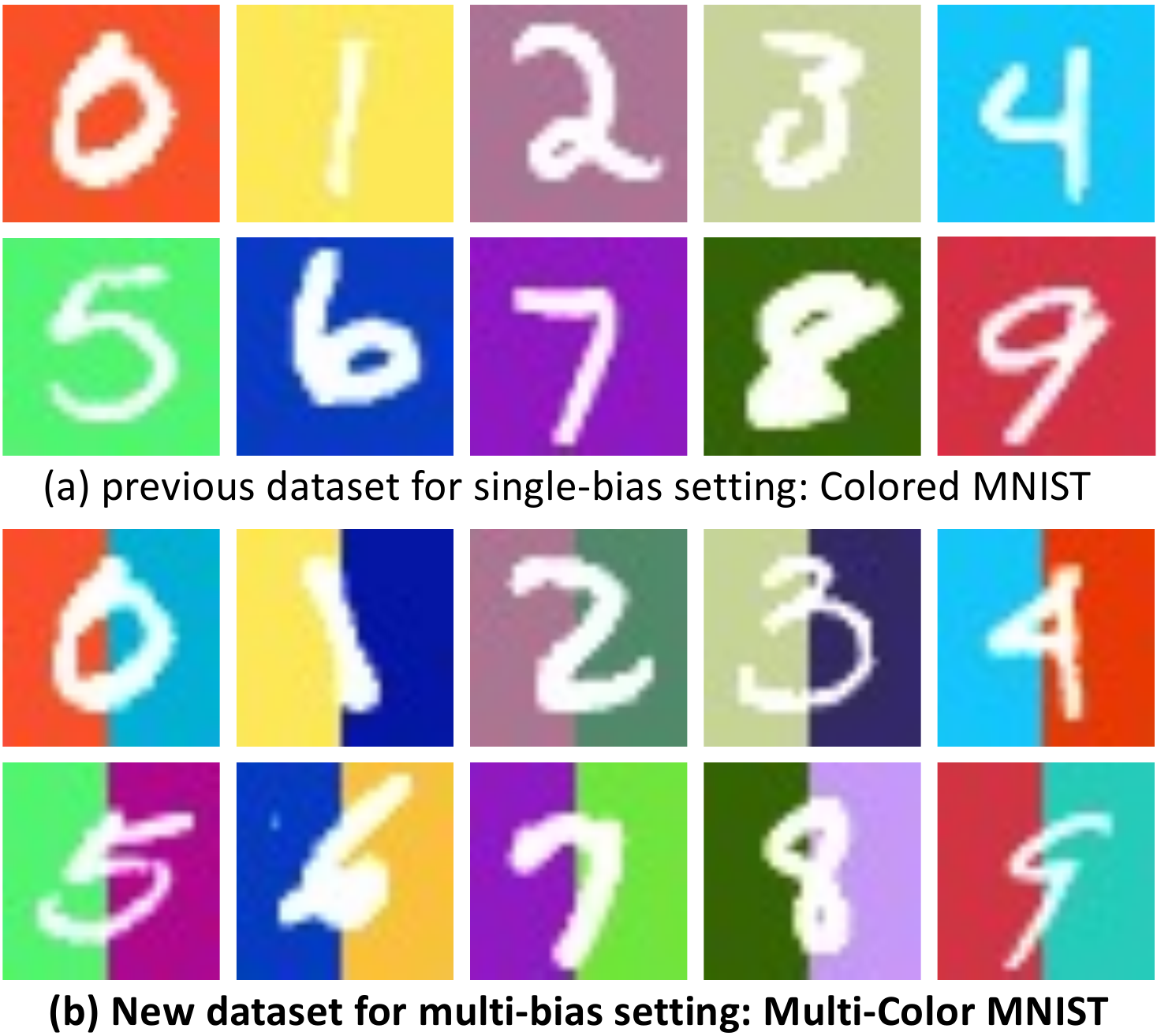}
    \caption{Comparison between (a) previous Colored MNIST~\cite{kim2019IEEEConf.Comput.Vis.PatternRecognit.CVPR,li2019IEEEConf.Comput.Vis.PatternRecognit.CVPRa,bahng2020Int.Conf.Mach.Learn.} with a single color bias and \textbf{(b)} our new Multi-Color MNIST dataset that contains \textit{two} bias attributes---\texttt{left color} and \texttt{right color}}
    \label{fig.multi_color_mnist}
  \end{minipage}\qquad
  \begin{minipage}{.55\textwidth}
      \centering
      \includegraphics[width=0.7\linewidth]{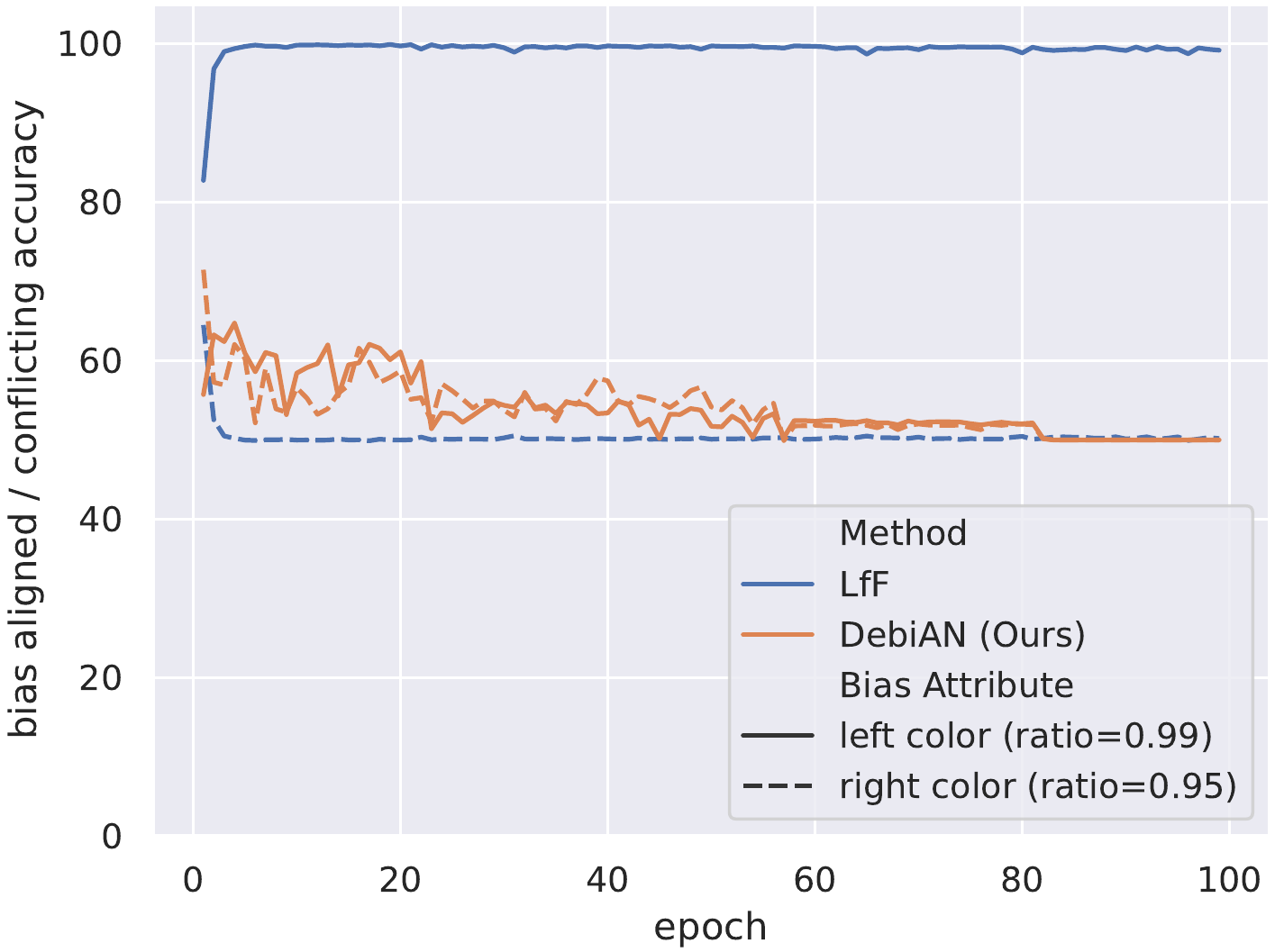}
      \caption{Evaluating bias discovery \wrt \texttt{left color}, \texttt{right color} biases throughout the training epochs on Multi-Color MNIST. LfF only finds the more salient \texttt{left color} bias (ratio=0.99), whereas DebiAN's \textit{discoverer} finds both biases at the early training stage. Then accuracies gradually converge to 50\% as debiasing is performed in the \textit{classifier}, making the \textit{discoverer} harder to find biases}
      \label{fig.bias_discovery_acc_trend}
  \end{minipage}
\end{figure}

\subsection{Experiment on Multi-Color MNIST}
\label{subsec.exp_synthetic}

Many previous works use synthetic datasets to benchmark bias mitigation performance. For example, Colored MNIST~\cite{kim2019IEEEConf.Comput.Vis.PatternRecognit.CVPR,li2019IEEEConf.Comput.Vis.PatternRecognit.CVPRa,bahng2020Int.Conf.Mach.Learn.} adds color bias to the original MNIST~\cite{lecun1998Proc.IEEE} dataset, where each digit class is spuriously correlated with color (see \cref{fig.multi_color_mnist} (a)). We compare DebiAN with other methods on the Colored MNIST dataset in \cref{sec.single_color_mnist_exp}. However, we believe that the single-bias setting is an oversimplification of the real-world scenario where multiple biases may exist. For instance, \citet{lang2021IEEEInt.Conf.Comput.Vis.ICCV} find that gender classifiers are biased with multiple independent bias attributes, including wearing lipsticks, eyebrow thickness, nose width, \etc. The benchmarking results on such a single-bias synthetic dataset may not help us to design better debiasing algorithms for real-world usage.

To this end, we propose \textbf{Multi-Color MNIST} dataset to benchmark debiasing methods under the multi-bias setting. In the training set, each digit class is spuriously correlated with \textit{two} bias attributes---\texttt{left color} and \texttt{right color} (\cref{fig.multi_color_mnist} (b)). Following the terms used in LfF~\cite{nam2020Adv.NeuralInf.Process.Syst.}, we call samples that can be correctly predicted with the bias attribute as bias-aligned samples. Samples that cannot be correctly predicted with the bias attribute are called bias-conflicting samples. For example, if most digit ``0'' images are in red \texttt{left color} in the training set, we call them bias-aligned samples \wrt \texttt{left color} attribute, and we regard digit ``0'' images in a different \texttt{left color} (\eg, yellow) as bias-conflicting samples. Since the dataset contains two bias attributes, there exist images that are bias-aligned \wrt to \texttt{left color} and bias-conflicting \wrt \texttt{right color} simultaneously, or vice versa. Following \cite{nam2020Adv.NeuralInf.Process.Syst.}, we use the \textit{ratio} of the bias-aligned samples for each bias attribute to indicate how strong the spurious correlation is in the training set. The two ratios for two bias attributes can be different, which is more common in the real-world scenario. The images in the testing set also contain two background colors, but the testing set has a balanced distribution of bias-aligned and bias-conflicting samples \wrt each bias attribute.

\noindent \textbf{Evaluation Metrics and Settings} \enspace Following \cite{nam2020Adv.NeuralInf.Process.Syst.}, we report the accuracy results in bias-aligned and bias-conflicting samples on the testing set. Since Multi-Color MNIST contains two bias attributes, we report the four accuracy results in the combination of (bias-aligned, bias-conflicting) $\times$ (\texttt{left color}, \texttt{right color}), \eg, middle four rows in \cref{tab.multi_color_mnist_results} for each method. We also report the unbiased accuracy, which averages the four results above. Here, we choose 0.99 as the ratio of bias-aligned samples \wrt \texttt{left color} and 0.95 as the ratio of bias-aligned samples \wrt \texttt{right color}. In this way, the \texttt{left color} is a \textit{more salient} bias than the \texttt{right color}. We report the results of other ratio combinations in \cref{subsec.ablate_ratios_multi_color_mnist}. We strictly use the same set of hyperparameters used in (single) Colored MNIST in LfF.
More details are in \cref{sec.supp.implementation_details}.

\begin{table}[t]
  \centering
  \caption{Debiasing results on Multi-Color MNIST dataset. The accuracy results in the four combinations of two bias attributes, (\ie, \texttt{left color} and \texttt{right color}) and (bias-aligned and bias-conflicting) are reported. Unbiased accuracy averages the results over all four combinations. We bold top-2 results and underline lowest results}
  \label{tab.multi_color_mnist_results}
  \begin{adjustbox}{width=0.9\linewidth}
  \begin{tabular}{@{}cccccccc@{}}
  \toprule
  \texttt{left color}       & \texttt{right color}      & \multirow{2}{*}{vanilla} & \multirow{2}{*}{LfF} & \multirow{2}{*}{EIIL} & \multirow{2}{*}{PGI} & \multirow{2}{*}{\textbf{w/o $\mathcal{L}_\text{UA}$ (Ours)}} & \multirow{2}{*}{\textbf{DebiAN (Ours)}} \\
  ratio = \textbf{0.99}      & ratio = \textbf{0.95}      &                          &                      &                       &                      &                          &                                         \\ \midrule
  bias-aligned     & bias-aligned     & \textbf{100.0}\tiny$\pm$0.0      & 99.6\tiny$\pm$0.5            & \textbf{100.0}\tiny$\pm$0.0   & \underline{98.6\tiny$\pm$2.3}            & \textbf{100.0}\tiny$\pm$0.0      & \textbf{100.0}\tiny$\pm$0.0                     \\
  bias-aligned     & bias-conflicting & 97.1\tiny$\pm$0.5                & \underline{4.7\tiny$\pm$0.5}             & \textbf{97.2}\tiny$\pm$1.5             & 82.6\tiny$\pm$19.6           & \textbf{97.2}\tiny$\pm$0.5       & 95.6\tiny$\pm$0.8                               \\
  bias-conflicting & bias-aligned     & 27.5\tiny$\pm$3.6                & \textbf{98.6}\tiny$\pm$0.4   & 70.8\tiny$\pm$4.9             & \underline{26.6\tiny$\pm$5.5}            & 71.6\tiny$\pm$0.7                & \textbf{76.5}\tiny$\pm$0.7                      \\
  bias-conflicting & bias-conflicting & 5.2\tiny$\pm$0.4                 & \underline{5.1\tiny$\pm$0.4}             & 10.9\tiny$\pm$0.8    & 9.5\tiny$\pm$3.2             & \textbf{13.8}\tiny$\pm$1.1                & \textbf{16.0}\tiny$\pm$1.8                      \\ \midrule
  \multicolumn{2}{c}{unbiased accuracy}        & 57.4\tiny$\pm$0.7                & \underline{52.0\tiny$\pm$0.1}            & 69.7\tiny$\pm$1.0    & 54.3\tiny$\pm$4.0            & \textbf{70.6}\tiny$\pm$0.3                & \textbf{72.0}\tiny$\pm$0.8                      \\ \bottomrule
  \end{tabular}
  \end{adjustbox}
\end{table}

\noindent \textbf{Debiasing Results on Multi-Color MNIST} \enspace The debiasing results are shown in \cref{tab.multi_color_mnist_results}. Except for LfF, all other methods achieve higher accuracy results on \texttt{left color} bias-aligned samples (1st and 2nd rows) than \texttt{right color} bias-aligned samples (1st and 3rd rows), indicating that most methods are more biased \wrt the more salient bias, \ie, \texttt{left color} (ratio=0.99) in the multi-bias setting. Unlike all other methods, LfF gives abnormal results---high accuracy results (\eg, 99.6, 98.6) for the \texttt{right color} bias-aligned samples and low accuracy results (\eg, 4.7, 5.1) for the \texttt{right color} bias-conflicting samples. Consequently, LfF achieves the worst unbiased accuracy (52.0). The results indicate that LfF only mitigates the more salient \texttt{left color} bias, rendering the classifier to learn the less salient \texttt{right color} bias (ratio=0.95). Compared with all other methods, DebiAN achieves better unbiased accuracy results (72.0). More importantly, DebiAN achieves much better debiasing result (16.0) in bias-conflicting samples \wrt both \texttt{left color} and \texttt{right color} attributes, where neither color can provide the shortcut for the correct digit class prediction, demonstrating better debiasing results of DebiAN for mitigating multiple biases simultaneously in the multi-bias setting, which is closer to the real-world scenarios.

\noindent \textbf{Bias Discovery: LfF vs. DebiAN} \enspace We further evaluate the bias discovery results throughout the entire training epochs, which helps us better understand LfF's abnormal results and DebiAN's advantages. We use LfF's ``biased model'' and DebiAN's \textit{discoverer} to predict if a given image is bias-aligned or bias-conflicting \wrt a bias attribute (\ie, binary classification, more details in \cref{subsec.bias_discovery_multi_color_mnist_impl_details}). We show the accuracy results of bias discovery \wrt each bias attribute at the end of each epoch in \cref{fig.bias_discovery_acc_trend}, which shows that LfF only discovers the more salient \texttt{left color} bias attribute (100\% accuracy), but completely ignores the less salient \texttt{right color} bias (50\% accuracy) throughout the entire training stage. It reveals the problem of LfF's definition of the unknown bias---an attribute in the dataset that is easier, which only holds in the single-bias setting but does not generalize to the multi-bias setting. In contrast, DebiAN uses the principled definition to define the bias---classifier's predictions that violate equal opportunity, enabling \textit{discoverer} to find both biases accurately at the beginning (it achieves about 60\% to 70\% accuracy because debiasing is simultaneously performed before the end of the first epoch). At the same time, DebiAN's alternate training scheme lets the classifier mitigate both biases, making the \textit{discoverer} harder to predict the biases, \eg, accuracies of both bias attributes gradually converge to 50\%. More discussions are in \cref{subsec.supp.discussion_bias_discovery}.

\noindent \textbf{Ablation Study on UA penalty} \enspace
We conduct an ablation study to show the effectiveness of Unbalanced Assignment (UA) penalty (\cref{subsec.method_unknown_bias_discovery}). \cref{tab.multi_color_mnist_results} shows that $\mathcal{L}_\text{UA}$ improves the debiasing results (see w/o $\mathcal{L}_\text{UA}$).
Besides, we also conduct ablation studies on different batch sizes, which are included in \cref{subsec.supp.ablate_batch_size}.

\begin{table}[t]
  \parbox{.55\linewidth}{
  \centering
  \caption{Results of mitigating the gender bias
  of \texttt{Blond Hair} classifier on CelebA~\cite{liu2015IEEEInt.Conf.Comput.Vis.ICCV}}
  \label{tab.celeba_blond}
  \begin{adjustbox}{width=\linewidth}
  \begin{tabular}{@{}cccccc@{}}
  \toprule
                   & vanilla   & LfF       & EIIL      & PGI                 & \textbf{DebiAN (Ours)}      \\ \midrule
  Avg Group Acc.   & 79.8\tiny$\pm$0.3 & 80.9\tiny$\pm$1.4 & 82.0\tiny$\pm$1.1 & 81.6\tiny$\pm$0.3          & \textbf{84.0}\tiny$\pm$1.4 \\
  Worst Group Acc. & 37.9\tiny$\pm$1.1 & 43.3\tiny$\pm$3.0 & 46.1\tiny$\pm$4.9 & 40.9\tiny$\pm$6.4          & \textbf{52.9}\tiny$\pm$4.7 \\ \bottomrule
  \end{tabular}
  \end{adjustbox}
  }
  \hfill
  \parbox{.4\linewidth}{
  \centering
  \caption{Accuracy results on bias-conflicting samples on bFFHQ~\cite{kim2021IEEEInt.Conf.Comput.Vis.ICCVa}
  }
  \label{tab.bffhq}
  \begin{adjustbox}{width=\linewidth}
  \begin{tabular}{@{}cccccc@{}}
  \toprule
  vanilla & LfF   & PGI       & EIIL      & BiaSwap & \textbf{DebiAN}             \\ \midrule
  51.03   & 55.61 & 55.2\tiny{$\pm$}5.3 & 59.2\tiny{$\pm$}1.9 & 58.87    & \textbf{62.8}\tiny{$\pm$}0.6 \\ \bottomrule
  \end{tabular}
  \end{adjustbox}
  }
\end{table}

\subsection{Experiments on Face Image Dataset}
\label{subsec.exp_face}
\noindent \textbf{Gender Bias Mitigation} \enspace In the face image domain, we conduct experiments to evaluate gender bias mitigation results on CelebA~\cite{liu2015IEEEInt.Conf.Comput.Vis.ICCV} dataset, which contains 200K celebrity faces annotated with 40 binary attributes. The dataset has spurious correlations between gender and \texttt{Blond Hair}, leading to gender biases when performing hair color classification. We follow most of the settings used in LfF, such as using ResNet-18~\cite{he2016IEEEConf.Comput.Vis.PatternRecognit.CVPR} as the backbone, using Adam~\cite{kingma2015Int.Conf.Learn.Represent.} optimizer, \etc. The only difference is that LfF reports the results on the validation set of CelebA, whereas we use the validation set to select the epoch with the best validation set accuracy (bias labels in the validation set are not used) to report the results on the testing set. All methods (including LfF) are benchmarked under the same setting. We report results in two evaluation metrics:
1) Average Group Accuracy (Avg. Group Acc.), which calculates the unweighted average of accuracies in four groups between target attribute and bias attribute, \ie, (male, female) $\times$ (blond, not blond);
2) Worst Group Accuracy (Worst Group Acc.)~\cite{sagawa*2020Int.Conf.Learn.Represent.}, which takes the lowest accuracy in the four groups. As shown in Tab.~\ref{tab.celeba_blond}, DebiAN achieves better Average and Worst Group accuracy results, which shows that DebiAN can better mitigate gender bias without labels. We also conduct experiments on bFFHQ~\cite{kim2021IEEEInt.Conf.Comput.Vis.ICCVa} where the training data contains the spurious correlation between age and gender. We compare DebiAN with other methods of gender bias mitigation. We strictly follow the setting in \cite{kim2021IEEEInt.Conf.Comput.Vis.ICCVa}. We report the age accuracy results on the bias-conflicting samples in the testing set in \cref{tab.bffhq}. The results of vanilla, LfF, and BiaSwap are from \cite{kim2021IEEEInt.Conf.Comput.Vis.ICCVa} and \cite{kim2021IEEEInt.Conf.Comput.Vis.ICCVa} does not provide the error bars. DebiAN achieves the best unsupervised results for mitigating gender bias.

\noindent \textbf{Mitigating Multiple Biases in Gender Classifier} \enspace The results on Multi-Color MNIST dataset suggest that DebiAN better mitigates multiple biases in the classifier. In the face image domain, a recent study~\cite{lang2021IEEEInt.Conf.Comput.Vis.ICCV} shows that gender classifier is biased by multiple attributes, such as \texttt{Heavy Makeup} and \texttt{Wearing Lipstick}. Hence, we train gender classifiers on CelebA dataset and evaluate Average Group Accuracy and Worst Group Accuracy \wrt these two bias attributes. As shown in \cref{tab.celeba_gender_multi_bias}, DebiAN achieves better debiasing results \wrt both bias attributes, proving that the \textit{discoverer} can find multiple biases in the classifier $C$ during the alternate training, enabling \textit{classifier} to mitigate multiple biases simultaneously.

\begin{table}[t]
	\caption{Results of mitigating multiple biases (\ie, \texttt{Wearing Lipstick} and \texttt{Heavy Makeup}) in gender classifier on CelebA dataset}
  \label{tab.celeba_gender_multi_bias}
\centering
\begin{adjustbox}{width=0.85\linewidth}
\begin{tabular}{@{}cc|ccccc@{}}
\toprule
bias attribute                    & metric           & vanilla            & LfF       & PGI       & EIIL               & \textbf{DebiAN (Ours)}            \\ \midrule
\multirow{2}{*}{\texttt{Wearing Lipstick}} & Avg. Group Acc.  & 86.6\tiny$\pm$0.4          & 87.0\tiny\tiny$\pm$0.9 & 86.9\tiny$\pm$3.1  & 86.3\tiny$\pm$1.0          & \textbf{88.5}\tiny$\pm$1.1 \\
                                  & Worst Group Acc. & 53.9\tiny$\pm$1.2          & 55.3\tiny$\pm$3.6 & 56.0\tiny$\pm$11.7 & 52.4\tiny$\pm$3.2          & \textbf{61.7}\tiny$\pm$4.2 \\ \midrule
\multirow{2}{*}{\texttt{Heavy Makeup}}     & Avg. Group Acc.  & 85.1\tiny$\pm$0.0          & 85.5\tiny$\pm$0.6 & 85.4\tiny$\pm$3.4  & 84.0\tiny$\pm$1.2          & \textbf{87.8}\tiny$\pm$1.3 \\
                                  & Worst Group Acc. & 45.4\tiny$\pm$0.0          & 46.9\tiny$\pm$2.6 & 46.9\tiny$\pm$13.1 & 40.9\tiny$\pm$4.5          & \textbf{56.0}\tiny$\pm$5.2 \\ \bottomrule
\end{tabular}
\end{adjustbox}
\end{table}

\begin{figure}[t]
\centering
\begin{minipage}{.45\textwidth}
  \begin{center}
    \includegraphics[width=\linewidth]{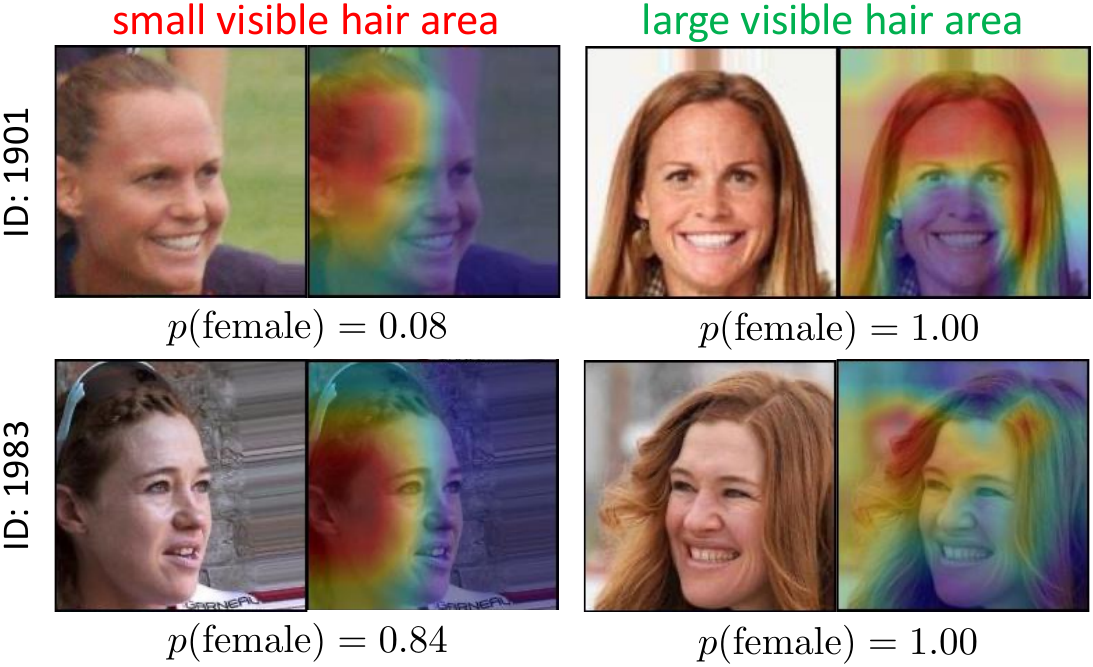}
  \end{center}
  \caption{Discovered bias of gender classifier: \texttt{visible hair area} based on \textit{discoverer}'s saliency map. $p(\text{female})$ is vanilla classifier's predicted probability of the face is female. In the two groups predicted by $D$, the visible hair areas are different, where the classifier has different confidences on gender for the same identity
  }
  \label{fig.hair_area_bias}
\end{minipage} \qquad
\begin{minipage}{.45\textwidth}
  \begin{center}
    \includegraphics[width=\linewidth]{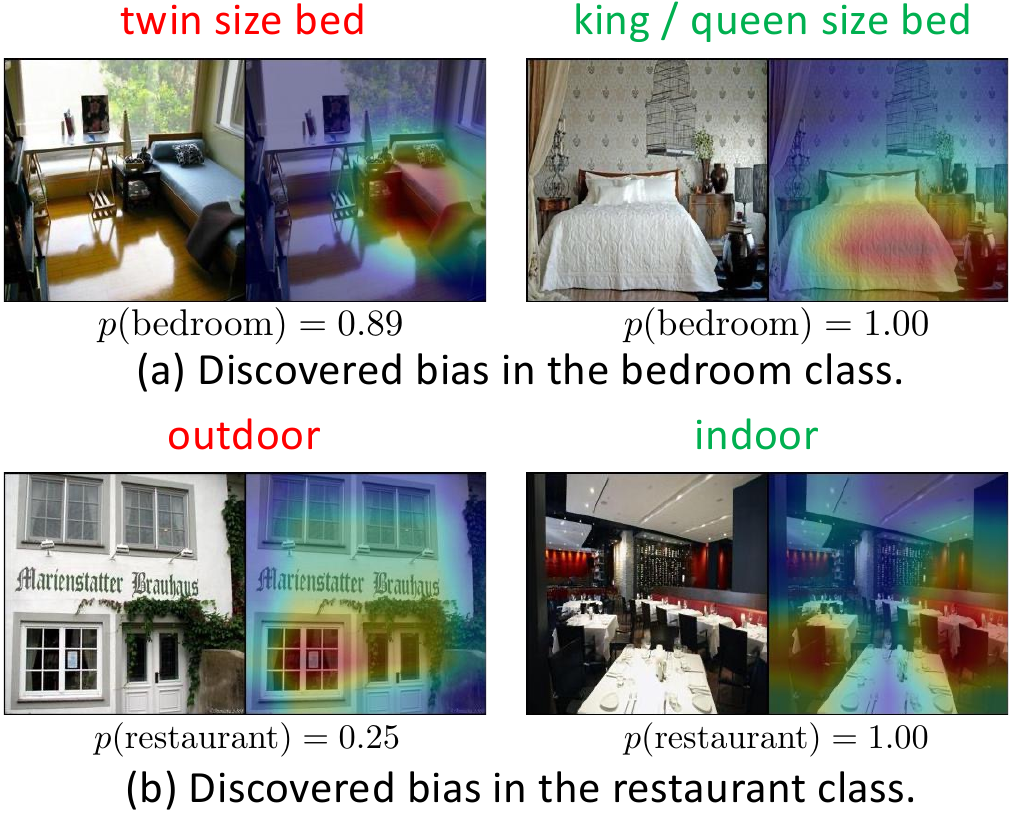}
  \end{center}
  \caption{Discovered biases in Places~\cite{zhou2018IEEETrans.PatternAnal.Mach.Intell.} dataset. We apply CAM on \textit{discoverer} to generate saliency map. The value $p(\text{bedroom})$ ($p(\text{restaurant})$) is vanilla classifier's predicted probability of the scene image is bedroom (restaurant)}
  \label{fig.scene_bias}
\end{minipage}
\end{figure}

\noindent \textbf{Identified Unknown Bias in Gender Classifier} \enspace Gender classifier can have more biases beyond \texttt{Wearing Lipstick} and \texttt{Heavy Makeup}. For example, \citet{balakrishnan2020Eur.Conf.Comput.Vis.ECCV} leverages StyleGAN2~\cite{karras2020IEEEConf.Comput.Vis.PatternRecognit.CVPR} to generate high-quality synthesized images and identify the \texttt{hair length} bias of the gender classifier, \eg, longer hair length makes the classifier predict the face as female. Related to their finding, the \textit{discoverer} $D$ in DebiAN identifies an interesting unknown bias: \texttt{visible hair area}. We use $D$ to predict the bias attribute group assignment on images in CelebA. To better interpret the bias attribute, we further use the identity labels in CelebA to cluster images with the same identity. Fig.~\ref{fig.hair_area_bias} shows that $D$ assigns images of the same identity into two distinct groups based on the visible hair area, which is verified by $D$'s CAM~\cite{zhou2016IEEEConf.Comput.Vis.PatternRecognit.CVPR} saliency maps. Strictly speaking, all females in Fig.~\ref{fig.hair_area_bias} have long hair. However, due to the hairstyle, pose, or occlusion, visible hair areas differ between the two groups. As a result, the gender classifier has lower predicted probabilities on the female images with smaller visible hair areas. More visualizations are shown in \cref{subsec.supp.more_face_bias_discovery_vis}.

\subsection{Experiments on Other Image Domains}
\label{subsec.exp_scene}

Our method is not limited to synthetic and face image domains. Here we conduct experiments on action recognition and scene classification tasks.

\noindent \textbf{Mitigating Place Bias in Action Recognition} \enspace We conduct experiments on Biased Action Recognition (BAR) dataset~\cite{nam2020Adv.NeuralInf.Process.Syst.}, an image dataset with the spurious correlation between action and place in the training set. The testing set only contains bias-conflicting samples. Hence, higher accuracy results on the testing set indicate better debiasing results. The accuracy results in \cref{tab.BAR} show that DebiAN achieves better debiasing results than other methods.

\begin{table}[t]
  \parbox{.45\linewidth}{
  \centering
  \caption{Results on Biased Action Recognition (BAR)~\cite{nam2020Adv.NeuralInf.Process.Syst.} dataset}
  \label{tab.BAR}
  \begin{adjustbox}{width=\linewidth}
  \begin{tabular}{@{}cccccc@{}}
  \toprule
  vanilla          & LfF         & PGI         & EIIL        & BiaSwap & \textbf{DebiAN} \\ \midrule
  51.85\tiny$\pm$5.92 &  62.98\tiny$\pm$2.76 & 65.19\tiny$\pm$1.32 & 65.44\tiny$\pm$1.17 & 52.44 & \textbf{69.88}\tiny$\pm$2.92   \\ \bottomrule
  \end{tabular}
  \end{adjustbox}
  }
  \hfill
  \parbox{.48\linewidth}{
  \centering
    \caption{Scene classification accuracy results on the \textit{unseen} LSUN~\cite{yu2016ArXiv150603365Cs} dataset
    }
    \label{tab.scene}
    \begin{adjustbox}{width=\linewidth}
      \begin{tabular}{@{}ccccc@{}}
        \toprule
        vanilla      & LfF        & PGI  & EIIL      & \textbf{DebiAN (Ours)}   \\ \midrule
        79.3\tiny$\pm$0.3 & 71.1\tiny$\pm$1.0 & 74.1\tiny{$\pm$1.9} & 79.4\tiny$\pm$0.2 & \textbf{80.0}\tiny$\pm$0.4 \\ \bottomrule
  \end{tabular}
  \end{adjustbox}
  }
  \end{table}

	\noindent \textbf{Improving Cross-dataset Generalization on Scene Classification} \enspace We conduct experiments on the more challenging scene classification task, where datasets are more complex and may contain multiple unknown biases. The biases in this task are underexplored by previous works partly due to the lack of attribute labels. Due to the absence of attribute labels, we use cross-dataset generalization~\cite{torralba2011IEEEConf.Comput.Vis.PatternRecognit.CVPR} to evaluate the debiasing results. Concretely, models are trained on Places~\cite{zhou2018IEEETrans.PatternAnal.Mach.Intell.} with ten classes overlapped with LSUN~\cite{yu2016ArXiv150603365Cs} (\eg, bedroom, classroom, \etc), and evaluated on the \textit{unseen} LSUN dataset. The results are shown in \cref{tab.scene}. DebiAN achieves the best result on the unseen LSUN dataset, showing that DebiAN unlearns the dataset biases~\cite{torralba2011IEEEConf.Comput.Vis.PatternRecognit.CVPR} in Places to improve the robustness against distributional shifts between different datasets.

\noindent \textbf{Identified Unknown Biases in Scene Classifier} \enspace DebiAN discovers Places dataset's unknown biases that humans may not preconceive. In Fig.~\ref{fig.scene_bias}, the \textit{discoverer} separates bedroom and restaurant images based on \texttt{size} of beds and \texttt{indoor/outdoor}. The vanilla classifier performs worse on bedroom images with twin-size beds and outdoor restaurant images (see more in \cref{subsec.supp.more_scene_bias_discovery_vis}).

\section{Conclusion}
\label{sec.conclusion}

We propose \textsc{Debiasing Alternate Networks} to discover and mitigate the unknown biases.
DebiAN identifies unknown biases that humans may not preconceive and achieves better unsupervised debiasing results. Our Multi-Color MNIST dataset surfaces previous methods' problems and demonstrates DebiAN's advantages in the multi-bias setting.
Admittedly, our work has some limitations, \eg, DebiAN focuses on binary or continuously valued bias attributes, not multi-class ones.
We hope our work can facilitate research on bias discovery and mitigation.

\subsubsection{Acknowledgment}
This work has been partially supported by the National Science Foundation (NSF) under Grant 1764415, 1909912, and 1934962 and by the Center of Excellence in Data Science, an Empire State Development-designated Center of Excellence. The article solely reflects the opinions and conclusions of its authors but not the funding agents.

\bibliographystyle{splncs04nat}
\bibliography{ref}

\begin{thebibliography}{79}
\providecommand{\natexlab}[1]{#1}
\providecommand{\url}[1]{\texttt{#1}}
\providecommand{\urlprefix}{URL }
\expandafter\ifx\csname urlstyle\endcsname\relax
  \providecommand{\doi}[1]{doi:\discretionary{}{}{}#1}\else
  \providecommand{\doi}{doi:\discretionary{}{}{}\begingroup
  \urlstyle{rm}\Url}\fi

\bibitem[{Agrawal and Srikant(1994)}]{agrawal1994Int.Conf.VeryLargeDataBases}
Agrawal, R., Srikant, R.: Fast {{Algorithms}} for {{Mining Association Rules}}
  in {{Large Databases}}. In: International {{Conference}} on {{Very Large Data
  Bases}} (1994)

\bibitem[{Ahmed \textit{et~al.}(2021)Ahmed, Bengio, van Seijen, and
  Courville}]{ahmed2021Int.Conf.Learn.Represent.}
Ahmed, F., Bengio, Y., van Seijen, H., Courville, A.: Systematic generalisation
  with group invariant predictions. In: International {{Conference}} on
  {{Learning Representations}} (2021)

\bibitem[{Albiero \textit{et~al.}(2020)Albiero, K.~S., Vangara, Zhang, King,
  and Bowyer}]{albiero2020IEEEWinterConf.Appl.Comput.Vis.WorkshopWACVW}
Albiero, V., K.~S., K., Vangara, K., Zhang, K., King, M.C., Bowyer, K.W.:
  Analysis of {{Gender Inequality In Face Recognition Accuracy}}. In: The
  {{IEEE Winter Conference}} on {{Applications}} of {{Computer Vision
  Workshops}} ({{WACVW}}) (2020)

\bibitem[{Alvi \textit{et~al.}(2018)Alvi, Zisserman, and
  Nellaaker}]{alvi2018Eur.Conf.Comput.Vis.WorkshopECCVW}
Alvi, M., Zisserman, A., Nellaaker, C.: Turning a {{Blind Eye}}: {{Explicit
  Removal}} of {{Biases}} and {{Variation}} from {{Deep Neural Network
  Embeddings}}. In: The {{European Conference}} on {{Computer Vision Workshop}}
  ({{ECCVW}}) (2018)

\bibitem[{Antol \textit{et~al.}(2015)Antol, Agrawal, Lu, Mitchell, Batra,
  Zitnick, and Parikh}]{antol2015IEEEInt.Conf.Comput.Vis.ICCV}
Antol, S., Agrawal, A., Lu, J., Mitchell, M., Batra, D., Zitnick, C.L., Parikh,
  D.: {{VQA}}: {{Visual Question Answering}}. In: The {{IEEE International
  Conference}} on {{Computer Vision}} ({{ICCV}}) (2015)

\bibitem[{Arjovsky \textit{et~al.}(2020)Arjovsky, Bottou, Gulrajani, and
  {Lopez-Paz}}]{arjovsky2020ArXiv190702893CsStat}
Arjovsky, M., Bottou, L., Gulrajani, I., {Lopez-Paz}, D.: Invariant {{Risk
  Minimization}}. arXiv:1907.02893 [cs, stat]  (2020)

\bibitem[{Bahng \textit{et~al.}(2020)Bahng, Chun, Yun, Choo, and
  Oh}]{bahng2020Int.Conf.Mach.Learn.}
Bahng, H., Chun, S., Yun, S., Choo, J., Oh, S.J.: Learning {{De-biased
  Representations}} with {{Biased Representations}}. In: International
  {{Conference}} on {{Machine Learning}} (2020)

\bibitem[{Balakrishnan \textit{et~al.}(2020)Balakrishnan, Xiong, Xia, and
  Perona}]{balakrishnan2020Eur.Conf.Comput.Vis.ECCV}
Balakrishnan, G., Xiong, Y., Xia, W., Perona, P.: Towards causal benchmarking
  of bias in face analysis algorithms. In: The {{European Conference}} on
  {{Computer Vision}} ({{ECCV}}) (2020)

\bibitem[{Bao and Barzilay(2022)}]{bao2022ArXiv}
Bao, Y., Barzilay, R.: Learning to {{Split}} for {{Automatic Bias Detection}}.
  ArXiv  (2022)

\bibitem[{Bao \textit{et~al.}(2022)Bao, Chang, and
  Barzilay}]{bao2022Proc.39thInt.Conf.Mach.Learn.Learning}
Bao, Y., Chang, S., Barzilay, D.R.: Learning {{Stable Classifiers}} by
  {{Transferring Unstable Features}}. In: {{International Conference}} on
  {{Machine Learning}} (2022)

\bibitem[{Bao \textit{et~al.}(2021)Bao, Chang, and
  Barzilay}]{bao2021Int.Conf.Mach.Learn.Predict}
Bao, Y., Chang, S., Barzilay, R.: Predict then {{Interpolate}}: {{A Simple
  Algorithm}} to {{Learn Stable Classifiers}}. In: International {{Conference}}
  on {{Machine Learning}} (2021)

\bibitem[{Buolamwini and
  Gebru(2018)}]{buolamwini2018ACMConf.FairnessAccount.Transpar.}
Buolamwini, J., Gebru, T.: Gender {{Shades}}: {{Intersectional Accuracy
  Disparities}} in {{Commercial Gender Classification}}. In: {{ACM Conference}}
  on {{Fairness}}, {{Accountability}}, and {{Transparency}} (2018)

\bibitem[{Cadene \textit{et~al.}(2019)Cadene, Dancette, {Ben younes}, Cord, and
  Parikh}]{cadene2019Adv.NeuralInf.Process.Syst.}
Cadene, R., Dancette, C., {Ben younes}, H., Cord, M., Parikh, D.: {{RUBi}}:
  {{Reducing Unimodal Biases}} for {{Visual Question Answering}}. In: Advances
  in {{Neural Information Processing Systems}} (2019)

\bibitem[{Choi \textit{et~al.}(2019)Choi, Gao, Messou, and
  Huang}]{choi2019Adv.NeuralInf.Process.Syst.}
Choi, J., Gao, C., Messou, J.C.E., Huang, J.B.: Why {{Can}}'t {{I Dance}} in
  the {{Mall}}? {{Learning}} to {{Mitigate Scene Bias}} in {{Action
  Recognition}}. In: Advances in {{Neural Information Processing Systems}}
  (2019)

\bibitem[{Clark \textit{et~al.}(2019)Clark, Yatskar, and
  Zettlemoyer}]{clark2019Empir.MethodsNat.Lang.Process.}
Clark, C., Yatskar, M., Zettlemoyer, L.: Don't {{Take}} the {{Easy Way Out}}:
  {{Ensemble Based Methods}} for {{Avoiding Known Dataset Biases}}. In:
  Empirical {{Methods}} in {{Natural Language Processing}} (2019)

\bibitem[{{Corbett-Davies} \textit{et~al.}(2017){Corbett-Davies}, Pierson,
  Feller, Goel, and
  Huq}]{corbett-davies2017Proc.23rdACMSIGKDDInt.Conf.Knowl.Discov.DataMin.}
{Corbett-Davies}, S., Pierson, E., Feller, A., Goel, S., Huq, A.: Algorithmic
  {{Decision Making}} and the {{Cost}} of {{Fairness}}. In: Proceedings of the
  23rd {{ACM SIGKDD International Conference}} on {{Knowledge Discovery}} and
  {{Data Mining}} (2017)

\bibitem[{Creager \textit{et~al.}(2021)Creager, Jacobsen, and
  Zemel}]{creager2021Int.Conf.Mach.Learn.}
Creager, E., Jacobsen, J.H., Zemel, R.: Environment {{Inference}} for
  {{Invariant Learning}}. In: International {{Conference}} on {{Machine
  Learning}} (2021)

\bibitem[{Creager \textit{et~al.}(2019)Creager, Madras, Jacobsen, Weis,
  Swersky, Pitassi, and Zemel}]{creager2019Int.Conf.Mach.Learn.}
Creager, E., Madras, D., Jacobsen, J.H., Weis, M., Swersky, K., Pitassi, T.,
  Zemel, R.: Flexibly {{Fair Representation Learning}} by {{Disentanglement}}.
  In: International {{Conference}} on {{Machine Learning}} (2019)

\bibitem[{Dhar \textit{et~al.}(2021)Dhar, Gleason, Roy, Castillo, and
  Chellappa}]{dhar2021IEEEInt.Conf.Comput.Vis.ICCV}
Dhar, P., Gleason, J., Roy, A., Castillo, C.D., Chellappa, R.: {{PASS}}:
  {{Protected Attribute Suppression System}} for {{Mitigating Bias}} in {{Face
  Recognition}}. In: The {{IEEE International Conference}} on {{Computer
  Vision}} ({{ICCV}}) (2021)

\bibitem[{Dwork \textit{et~al.}(2012)Dwork, Hardt, Pitassi, Reingold, and
  Zemel}]{dwork2012Proc.3rdInnov.Theor.Comput.Sci.Conf.}
Dwork, C., Hardt, M., Pitassi, T., Reingold, O., Zemel, R.: Fairness through
  awareness. In: Proceedings of the 3rd {{Innovations}} in {{Theoretical
  Computer Science Conference}} (2012)

\bibitem[{Geirhos \textit{et~al.}(2020)Geirhos, Jacobsen, Michaelis, Zemel,
  Brendel, Bethge, and Wichmann}]{geirhos2020Nat.Mach.Intell.}
Geirhos, R., Jacobsen, J.H., Michaelis, C., Zemel, R., Brendel, W., Bethge, M.,
  Wichmann, F.A.: Shortcut learning in deep neural networks. Nature Machine
  Intelligence  (2020)

\bibitem[{Gong \textit{et~al.}(2020)Gong, Liu, and
  Jain}]{gong2020Eur.Conf.Comput.Vis.ECCV}
Gong, S., Liu, X., Jain, A.K.: Jointly {{De-biasing Face Recognition}} and
  {{Demographic Attribute Estimation}}. In: The {{European Conference}} on
  {{Computer Vision}} ({{ECCV}}) (2020)

\bibitem[{Goodfellow \textit{et~al.}(2014)Goodfellow, {Pouget-Abadie}, Mirza,
  Xu, {Warde-Farley}, Ozair, Courville, and
  Bengio}]{goodfellow2014Adv.NeuralInf.Process.Syst.}
Goodfellow, I., {Pouget-Abadie}, J., Mirza, M., Xu, B., {Warde-Farley}, D.,
  Ozair, S., Courville, A., Bengio, Y.: Generative adversarial nets. In:
  Advances in {{Neural Information Processing Systems}} (2014)

\bibitem[{{Grgic-Hlaca} \textit{et~al.}(2016){Grgic-Hlaca}, Zafar, Gummadi, and
  Weller}]{grgic-hlaca2016NIPSSymp.Mach.Learn.Law}
{Grgic-Hlaca}, N., Zafar, M.B., Gummadi, K.P., Weller, A.: The case for process
  fairness in learning: {{Feature}} selection for fair decision making. In:
  {{NIPS}} Symposium on Machine Learning and the Law (2016)

\bibitem[{Hanna \textit{et~al.}(2020)Hanna, Denton, Smart, and
  {Smith-Loud}}]{hanna2020Conf.FairnessAccount.Transpar.critical}
Hanna, A., Denton, E., Smart, A., {Smith-Loud}, J.: Towards a critical race
  methodology in algorithmic fairness. In: Conference on {{Fairness}},
  {{Accountability}}, and {{Transparency}} (2020)

\bibitem[{Hardt \textit{et~al.}(2016)Hardt, Price, and
  Srebro}]{hardt2016Adv.NeuralInf.Process.Syst.}
Hardt, M., Price, E., Srebro, N.: Equality of {{Opportunity}} in {{Supervised
  Learning}}. In: Advances in {{Neural Information Processing Systems}} (2016)

\bibitem[{Hazirbas \textit{et~al.}(2021)Hazirbas, Bitton, Dolhansky, Pan,
  Gordo, and Ferrer}]{hazirbas2021ArXiv210402821Cs}
Hazirbas, C., Bitton, J., Dolhansky, B., Pan, J., Gordo, A., Ferrer, C.C.:
  Towards {{Measuring Fairness}} in {{AI}}: The {{Casual Conversations
  Dataset}}. arXiv:2104.02821 [cs]  (2021)

\bibitem[{He \textit{et~al.}(2016)He, Zhang, Ren, and
  Sun}]{he2016IEEEConf.Comput.Vis.PatternRecognit.CVPR}
He, K., Zhang, X., Ren, S., Sun, J.: Deep residual learning for image
  recognition. In: The {{IEEE Conference}} on {{Computer Vision}} and {{Pattern
  Recognition}} ({{CVPR}}) (2016)

\bibitem[{Hendricks \textit{et~al.}(2018)Hendricks, Burns, Saenko, Darrell, and
  Rohrbach}]{hendricks2018Eur.Conf.Comput.Vis.ECCV}
Hendricks, L.A., Burns, K., Saenko, K., Darrell, T., Rohrbach, A.: Women also
  {{Snowboard}}: {{Overcoming Bias}} in {{Captioning Models}}. In: The
  {{European Conference}} on {{Computer Vision}} ({{ECCV}}) (2018)

\bibitem[{Jia \textit{et~al.}(2020)Jia, Meng, Zhao, and
  Chang}]{jia2020Annu.Meet.Assoc.Comput.Linguist.}
Jia, S., Meng, T., Zhao, J., Chang, K.W.: Mitigating {{Gender Bias
  Amplification}} in {{Distribution}} by {{Posterior Regularization}}. In:
  Annual {{Meeting}} of the {{Association}} for {{Computational Linguistics}}
  (2020)

\bibitem[{Joo and
  K{\"a}rkk{\"a}inen(2020)}]{joo2020Int.WorkshopFairnessAccount.Transpar.EthicsMultimed.}
Joo, J., K{\"a}rkk{\"a}inen, K.: Gender {{Slopes}}: {{Counterfactual Fairness}}
  for {{Computer Vision Models}} by {{Attribute Manipulation}}. In:
  International {{Workshop}} on {{Fairness}}, {{Accountability}},
  {{Transparency}} and {{Ethics}} in {{Multimedia}} (2020)

\bibitem[{Kamiran and Calders(2012)}]{kamiran2012KnowlInfSyst}
Kamiran, F., Calders, T.: Data preprocessing techniques for classification
  without discrimination. Knowledge and Information Systems  (2012)

\bibitem[{Karras \textit{et~al.}(2019)Karras, Laine, and
  Aila}]{karras2019IEEEConf.Comput.Vis.PatternRecognit.CVPR}
Karras, T., Laine, S., Aila, T.: A {{Style-Based Generator Architecture}} for
  {{Generative Adversarial Networks}}. In: The {{IEEE Conference}} on
  {{Computer Vision}} and {{Pattern Recognition}} ({{CVPR}}) (2019)

\bibitem[{Karras \textit{et~al.}(2020)Karras, Laine, Aittala, Hellsten,
  Lehtinen, and Aila}]{karras2020IEEEConf.Comput.Vis.PatternRecognit.CVPR}
Karras, T., Laine, S., Aittala, M., Hellsten, J., Lehtinen, J., Aila, T.:
  Analyzing and {{Improving}} the {{Image Quality}} of {{StyleGAN}}. In: The
  {{IEEE Conference}} on {{Computer Vision}} and {{Pattern Recognition}}
  ({{CVPR}}) (2020)

\bibitem[{Kim \textit{et~al.}(2019)Kim, Kim, Kim, Kim, and
  Kim}]{kim2019IEEEConf.Comput.Vis.PatternRecognit.CVPR}
Kim, B., Kim, H., Kim, K., Kim, S., Kim, J.: Learning {{Not}} to {{Learn}}:
  {{Training Deep Neural Networks With Biased Data}}. In: The {{IEEE
  Conference}} on {{Computer Vision}} and {{Pattern Recognition}} ({{CVPR}})
  (2019)

\bibitem[{Kim \textit{et~al.}(2021)Kim, Lee, and
  Choo}]{kim2021IEEEInt.Conf.Comput.Vis.ICCVa}
Kim, E., Lee, J., Choo, J.: {{BiaSwap}}: {{Removing}} dataset bias with
  bias-tailored swapping augmentation. In: The {{IEEE International
  Conference}} on {{Computer Vision}} ({{ICCV}}) (2021)

\bibitem[{Kingma and Ba(2015)}]{kingma2015Int.Conf.Learn.Represent.}
Kingma, D.P., Ba, J.: Adam: {{A Method}} for {{Stochastic Optimization}}. In:
  International {{Conference}} on {{Learning Representations}} (2015)

\bibitem[{Krishnakumar \textit{et~al.}(2021)Krishnakumar, Prabhu, Sudhakar, and
  Hoffman}]{krishnakumar2021Br.Mach.Vis.Conf.BMVC}
Krishnakumar, A., Prabhu, V., Sudhakar, S., Hoffman, J.: {{UDIS}}:
  {{Unsupervised Discovery}} of {{Bias}} in {{Deep Visual Recognition Models}}.
  In: British {{Machine Vision Conference}}, {{BMVC}} (2021)

\bibitem[{Kusner \textit{et~al.}(2017)Kusner, Loftus, Russell, and
  Silva}]{kusner2017Adv.NeuralInf.Process.Syst.}
Kusner, M.J., Loftus, J., Russell, C., Silva, R.: Counterfactual {{Fairness}}.
  In: Advances in {{Neural Information Processing Systems}} (2017)

\bibitem[{Lahoti \textit{et~al.}(2020)Lahoti, Beutel, Chen, Lee, Prost, Thain,
  Wang, and Chi}]{lahoti2020Adv.NeuralInf.Process.Syst.}
Lahoti, P., Beutel, A., Chen, J., Lee, K., Prost, F., Thain, N., Wang, X., Chi,
  E.: Fairness without {{Demographics}} through {{Adversarially Reweighted
  Learning}}. In: Advances in {{Neural Information Processing Systems}} (2020)

\bibitem[{Lang \textit{et~al.}(2021)Lang, Gandelsman, Yarom, Wald, Elidan,
  Hassidim, Freeman, Isola, Globerson, Irani, and
  Mosseri}]{lang2021IEEEInt.Conf.Comput.Vis.ICCV}
Lang, O., Gandelsman, Y., Yarom, M., Wald, Y., Elidan, G., Hassidim, A.,
  Freeman, W.T., Isola, P., Globerson, A., Irani, M., Mosseri, I.: Explaining
  in {{Style}}: {{Training}} a {{GAN}} to explain a classifier in
  {{StyleSpace}}. In: The {{IEEE International Conference}} on {{Computer
  Vision}} ({{ICCV}}) (2021)

\bibitem[{Lecun \textit{et~al.}(1998)Lecun, Bottou, Bengio, and
  Haffner}]{lecun1998Proc.IEEE}
Lecun, Y., Bottou, L., Bengio, Y., Haffner, P.: Gradient-based learning applied
  to document recognition. Proceedings of the IEEE  (1998)

\bibitem[{Lee \textit{et~al.}(2021)Lee, Kim, Lee, Lee, and
  Choo}]{lee2021Adv.NeuralInf.Process.Syst.}
Lee, J., Kim, E., Lee, J., Lee, J., Choo, J.: Learning {{Debiased
  Representation}} via {{Disentangled Feature Augmentation}}. In: Advances in
  {{Neural Information Processing Systems}} (2021)

\bibitem[{Li \textit{et~al.}(2019)Li, Zhang, Zhang, Huang, He, Lyu, and
  Gao}]{li2019IEEEConf.Comput.Vis.PatternRecognit.CVPRd}
Li, W., Zhang, P., Zhang, L., Huang, Q., He, X., Lyu, S., Gao, J.:
  Object-{{Driven Text-To-Image Synthesis}} via {{Adversarial Training}}. In:
  The {{IEEE Conference}} on {{Computer Vision}} and {{Pattern Recognition}}
  ({{CVPR}}) (2019)

\bibitem[{Li \textit{et~al.}(2018)Li, Li, and
  Vasconcelos}]{li2018Eur.Conf.Comput.Vis.ECCV}
Li, Y., Li, Y., Vasconcelos, N.: {{RESOUND}}: {{Towards Action Recognition}}
  without {{Representation Bias}}. In: The {{European Conference}} on
  {{Computer Vision}} ({{ECCV}}) (2018)

\bibitem[{Li and
  Vasconcelos(2019)}]{li2019IEEEConf.Comput.Vis.PatternRecognit.CVPRa}
Li, Y., Vasconcelos, N.: {{REPAIR}}: {{Removing Representation Bias}} by
  {{Dataset Resampling}}. In: The {{IEEE Conference}} on {{Computer Vision}}
  and {{Pattern Recognition}} ({{CVPR}}) (2019)

\bibitem[{Li and Xu(2021)}]{li2021IEEEInt.Conf.Comput.Vis.ICCV}
Li, Z., Xu, C.: Discover the {{Unknown Biased Attribute}} of an {{Image
  Classifier}}. In: The {{IEEE International Conference}} on {{Computer
  Vision}} ({{ICCV}}) (2021)

\bibitem[{Lin \textit{et~al.}(2017)Lin, Goyal, Girshick, He, and
  Dollar}]{lin2017IEEEInt.Conf.Comput.Vis.ICCV}
Lin, T.Y., Goyal, P., Girshick, R., He, K., Dollar, P.: Focal {{Loss}} for
  {{Dense Object Detection}}. In: The {{IEEE International Conference}} on
  {{Computer Vision}} ({{ICCV}}) (2017)

\bibitem[{Lin \textit{et~al.}(2014)Lin, Maire, Belongie, Hays, Perona, Ramanan,
  Doll{\'a}r, and Zitnick}]{lin2014Eur.Conf.Comput.Vis.ECCV}
Lin, T.Y., Maire, M., Belongie, S., Hays, J., Perona, P., Ramanan, D.,
  Doll{\'a}r, P., Zitnick, C.L.: Microsoft {{COCO}}: {{Common Objects}} in
  {{Context}}. In: The {{European Conference}} on {{Computer Vision}}
  ({{ECCV}}) (2014)

\bibitem[{Liu \textit{et~al.}(2015)Liu, Luo, Wang, and
  Tang}]{liu2015IEEEInt.Conf.Comput.Vis.ICCV}
Liu, Z., Luo, P., Wang, X., Tang, X.: Deep {{Learning Face Attributes}} in the
  {{Wild}}. In: The {{IEEE International Conference}} on {{Computer Vision}}
  ({{ICCV}}) (2015)

\bibitem[{Manjunatha \textit{et~al.}(2019)Manjunatha, Saini, and
  Davis}]{manjunatha2019IEEEConf.Comput.Vis.PatternRecognit.CVPR}
Manjunatha, V., Saini, N., Davis, L.S.: Explicit {{Bias Discovery}} in {{Visual
  Question Answering Models}}. In: The {{IEEE Conference}} on {{Computer
  Vision}} and {{Pattern Recognition}} ({{CVPR}}) (2019)

\bibitem[{Mitchell \textit{et~al.}(2019)Mitchell, Wu, Zaldivar, Barnes,
  Vasserman, Hutchinson, Spitzer, Raji, and
  Gebru}]{mitchell2019ACMConf.FairnessAccount.Transpar.}
Mitchell, M., Wu, S., Zaldivar, A., Barnes, P., Vasserman, L., Hutchinson, B.,
  Spitzer, E., Raji, I.D., Gebru, T.: Model {{Cards}} for {{Model Reporting}}.
  In: {{ACM Conference}} on {{Fairness}}, {{Accountability}}, and
  {{Transparency}} (2019)

\bibitem[{Nam \textit{et~al.}(2020)Nam, Cha, Ahn, Lee, and
  Shin}]{nam2020Adv.NeuralInf.Process.Syst.}
Nam, J., Cha, H., Ahn, S., Lee, J., Shin, J.: Learning from {{Failure}}:
  {{Training Debiased Classifier}} from {{Biased Classifier}}. In: Advances in
  {{Neural Information Processing Systems}} (2020)

\bibitem[{Paszke \textit{et~al.}(2019)Paszke, Gross, Massa, Lerer, Bradbury,
  Chanan, Killeen, Lin, Gimelshein, Antiga, Desmaison, Kopf, Yang, DeVito,
  Raison, Tejani, Chilamkurthy, Steiner, Fang, Bai, and
  Chintala}]{paszke2019Adv.NeuralInf.Process.Syst.}
Paszke, A., Gross, S., Massa, F., Lerer, A., Bradbury, J., Chanan, G., Killeen,
  T., Lin, Z., Gimelshein, N., Antiga, L., Desmaison, A., Kopf, A., Yang, E.,
  DeVito, Z., Raison, M., Tejani, A., Chilamkurthy, S., Steiner, B., Fang, L.,
  Bai, J., Chintala, S.: {{PyTorch}}: {{An Imperative Style}},
  {{High-Performance Deep Learning Library}}. In: Advances in {{Neural
  Information Processing Systems}} (2019)

\bibitem[{Pleiss \textit{et~al.}(2017)Pleiss, Raghavan, Wu, Kleinberg, and
  Weinberger}]{pleiss2017Adv.NeuralInf.Process.Syst.}
Pleiss, G., Raghavan, M., Wu, F., Kleinberg, J., Weinberger, K.Q.: On
  {{Fairness}} and {{Calibration}}. In: Advances in {{Neural Information
  Processing Systems}} (2017)

\bibitem[{Reddy \textit{et~al.}(2021)Reddy, Sharma, Mehri, {Romero-Soriano},
  Shabanian, and
  Honari}]{reddy2021Thirty-FifthConf.NeuralInf.Process.Syst.DatasetsBenchmarksTrackRound1}
Reddy, C., Sharma, D., Mehri, S., {Romero-Soriano}, A., Shabanian, S., Honari,
  S.: Benchmarking {{Bias Mitigation Algorithms}} in {{Representation
  Learning}} through {{Fairness Metrics}}. In: Thirty-Fifth {{Conference}} on
  {{Neural Information Processing Systems Datasets}} and {{Benchmarks Track}}
  ({{Round}} 1) (2021)

\bibitem[{Sagawa* \textit{et~al.}(2020)Sagawa*, Koh*, Hashimoto, and
  Liang}]{sagawa*2020Int.Conf.Learn.Represent.}
Sagawa*, S., Koh*, P.W., Hashimoto, T.B., Liang, P.: Distributionally {{Robust
  Neural Networks}} for {{Group Shifts}}: {{On}} the {{Importance}} of
  {{Regularization}} for {{Worst-Case Generalization}}. In: International
  {{Conference}} on {{Learning Representations}} (2020)

\bibitem[{Sarhan \textit{et~al.}(2020)Sarhan, Navab, and
  Albarqouni}]{sarhan2020Eur.Conf.Comput.Vis.ECCV}
Sarhan, M.H., Navab, N., Albarqouni, S.: Fairness by {{Learning Orthogonal
  Disentangled Representations}}. In: The {{European Conference}} on {{Computer
  Vision}} ({{ECCV}}) (2020)

\bibitem[{Selvaraju \textit{et~al.}(2017)Selvaraju, Cogswell, Das, Vedantam,
  Parikh, and Batra}]{selvaraju2017IEEEInt.Conf.Comput.Vis.ICCV}
Selvaraju, R.R., Cogswell, M., Das, A., Vedantam, R., Parikh, D., Batra, D.:
  Grad-cam: {{Visual}} explanations from deep networks via gradient-based
  localization. In: The {{IEEE International Conference}} on {{Computer
  Vision}} ({{ICCV}}) (2017)

\bibitem[{Selvaraju \textit{et~al.}(2019)Selvaraju, Cogswell, Das, Vedantam,
  Parikh, and Batra}]{selvaraju2019Int.J.Comput.Vis.}
Selvaraju, R.R., Cogswell, M., Das, A., Vedantam, R., Parikh, D., Batra, D.:
  Grad-{{CAM}}: {{Visual Explanations}} from {{Deep Networks}} via
  {{Gradient-Based Localization}}. International Journal of Computer Vision
  (2019)

\bibitem[{Shrestha \textit{et~al.}(2022)Shrestha, Kafle, and
  Kanan}]{shrestha2022IEEEWinterConf.Appl.Comput.Vis.WACV}
Shrestha, R., Kafle, K., Kanan, C.: An {{Investigation}} of {{Critical Issues}}
  in {{Bias Mitigation Techniques}}. In: The {{IEEE Winter Conference}} on
  {{Applications}} of {{Computer Vision}} ({{WACV}}) (2022)

\bibitem[{Singh \textit{et~al.}(2020)Singh, Mahajan, Grauman, Lee, Feiszli, and
  Ghadiyaram}]{singh2020IEEEConf.Comput.Vis.PatternRecognit.CVPR}
Singh, K.K., Mahajan, D., Grauman, K., Lee, Y.J., Feiszli, M., Ghadiyaram, D.:
  Don't {{Judge}} an {{Object}} by {{Its Context}}: {{Learning}} to {{Overcome
  Contextual Bias}}. In: The {{IEEE Conference}} on {{Computer Vision}} and
  {{Pattern Recognition}} ({{CVPR}}) (2020)

\bibitem[{Sohoni \textit{et~al.}(2020)Sohoni, Dunnmon, Angus, Gu, and
  R{\'e}}]{sohoni2020Adv.NeuralInf.Process.Syst.}
Sohoni, N.S., Dunnmon, J.A., Angus, G., Gu, A., R{\'e}, C.: No {{Subclass Left
  Behind}}: {{Fine-Grained Robustness}} in {{Coarse-Grained Classification
  Problems}}. In: Advances in {{Neural Information Processing Systems}} (2020)

\bibitem[{Torralba and
  Efros(2011)}]{torralba2011IEEEConf.Comput.Vis.PatternRecognit.CVPR}
Torralba, A., Efros, A.A.: Unbiased look at dataset bias. In: The {{IEEE
  Conference}} on {{Computer Vision}} and {{Pattern Recognition}} ({{CVPR}})
  (2011)

\bibitem[{Tzeng \textit{et~al.}(2015)Tzeng, Hoffman, Darrell, and
  Saenko}]{tzeng2015IEEEInt.Conf.Comput.Vis.ICCV}
Tzeng, E., Hoffman, J., Darrell, T., Saenko, K.: Simultaneous {{Deep Transfer
  Across Domains}} and {{Tasks}}. In: The {{IEEE International Conference}} on
  {{Computer Vision}} ({{ICCV}}) (2015)

\bibitem[{Verma and
  Rubin(2018)}]{verma20182018IEEEACMInt.WorkshopSoftw.FairnessFairWare}
Verma, S., Rubin, J.: Fairness {{Definitions Explained}}. In: 2018
  {{IEEE}}/{{ACM International Workshop}} on {{Software Fairness}}
  ({{FairWare}}) (2018)

\bibitem[{Wang \textit{et~al.}(2020{\natexlab{a}})Wang, Narayanan, and
  Russakovsky}]{wang2020Eur.Conf.Comput.Vis.ECCVd}
Wang, A., Narayanan, A., Russakovsky, O.: {{REVISE}}: {{A Tool}} for
  {{Measuring}} and {{Mitigating Bias}} in {{Image Datasets}}. In: The
  {{European Conference}} on {{Computer Vision}} ({{ECCV}})
  (2020{\natexlab{a}})

\bibitem[{Wang \textit{et~al.}(2019{\natexlab{a}})Wang, He, Lipton, and
  Xing}]{wang2019Int.Conf.Learn.Represent.b}
Wang, H., He, Z., Lipton, Z.C., Xing, E.P.: Learning {{Robust Representations}}
  by {{Projecting Superficial Statistics Out}}. In: International
  {{Conference}} on {{Learning Representations}} (2019{\natexlab{a}})

\bibitem[{Wang \textit{et~al.}(2021{\natexlab{a}})Wang, Liu, and
  Wang}]{wang2021Empir.MethodsNat.Lang.Process.}
Wang, J., Liu, Y., Wang, X.E.: Are {{Gender-Neutral Queries Really
  Gender-Neutral}}? {{Mitigating Gender Bias}} in {{Image Search}}. In:
  Empirical {{Methods}} in {{Natural Language Processing}} (2021{\natexlab{a}})

\bibitem[{Wang \textit{et~al.}(2021{\natexlab{b}})Wang, Yue, Huang, Sun, and
  Zhang}]{wang2021Adv.NeuralInf.Process.Syst.SelfSupervised}
Wang, T., Yue, Z., Huang, J., Sun, Q., Zhang, H.: Self-{{Supervised Learning
  Disentangled Group Representation}} as {{Feature}}. In: Advances in {{Neural
  Information Processing Systems}} (2021{\natexlab{b}})

\bibitem[{Wang \textit{et~al.}(2020{\natexlab{b}})Wang, Ang, and
  Lee}]{wang2020IEEEConf.Comput.Vis.PatternRecognit.CVPRe}
Wang, X., Ang, M.H., Lee, G.H.: Cascaded {{Refinement Network}} for {{Point
  Cloud Completion}}. In: The {{IEEE Conference}} on {{Computer Vision}} and
  {{Pattern Recognition}} ({{CVPR}}) (2020{\natexlab{b}})

\bibitem[{Wang \textit{et~al.}(2019{\natexlab{b}})Wang, Liu, Li, Sheng, Yan,
  Wang, and Shao}]{wang2019IEEEInt.Conf.Comput.Vis.ICCVa}
Wang, Z., Liu, X., Li, H., Sheng, L., Yan, J., Wang, X., Shao, J.: {{CAMP}}:
  {{Cross-Modal Adaptive Message Passing}} for {{Text-Image Retrieval}}. In:
  The {{IEEE International Conference}} on {{Computer Vision}} ({{ICCV}})
  (2019{\natexlab{b}})

\bibitem[{Wang \textit{et~al.}(2020{\natexlab{c}})Wang, Qinami, Karakozis,
  Genova, Nair, Hata, and
  Russakovsky}]{wang2020IEEEConf.Comput.Vis.PatternRecognit.CVPR}
Wang, Z., Qinami, K., Karakozis, I.C., Genova, K., Nair, P., Hata, K.,
  Russakovsky, O.: Towards {{Fairness}} in {{Visual Recognition}}: {{Effective
  Strategies}} for {{Bias Mitigation}}. In: The {{IEEE Conference}} on
  {{Computer Vision}} and {{Pattern Recognition}} ({{CVPR}})
  (2020{\natexlab{c}})

\bibitem[{Yu \textit{et~al.}(2016)Yu, Seff, Zhang, Song, Funkhouser, and
  Xiao}]{yu2016ArXiv150603365Cs}
Yu, F., Seff, A., Zhang, Y., Song, S., Funkhouser, T., Xiao, J.: {{LSUN}}:
  {{Construction}} of a {{Large-scale Image Dataset}} using {{Deep Learning}}
  with {{Humans}} in the {{Loop}}. arXiv:1506.03365 [cs]  (2016)

\bibitem[{Zhang \textit{et~al.}(2018)Zhang, Lemoine, and
  Mitchell}]{zhang2018AAAIACMConf.AIEthicsSoc.}
Zhang, B.H., Lemoine, B., Mitchell, M.: Mitigating {{Unwanted Biases}} with
  {{Adversarial Learning}}. In: {{AAAI}}/{{ACM Conference}} on {{AI}},
  {{Ethics}}, and {{Society}} (2018)

\bibitem[{Zhang and Sabuncu(2018)}]{zhang2018Adv.NeuralInf.Process.Syst.}
Zhang, Z., Sabuncu, M.: Generalized {{Cross Entropy Loss}} for {{Training Deep
  Neural Networks}} with {{Noisy Labels}}. In: Advances in {{Neural Information
  Processing Systems}} (2018)

\bibitem[{Zhao \textit{et~al.}(2017)Zhao, Wang, Yatskar, Ordonez, and
  Chang}]{zhao2017Empir.MethodsNat.Lang.Process.}
Zhao, J., Wang, T., Yatskar, M., Ordonez, V., Chang, K.W.: Men {{Also Like
  Shopping}}: {{Reducing Gender Bias Amplification}} using {{Corpus-level
  Constraints}}. In: Empirical {{Methods}} in {{Natural Language Processing}}
  (2017)

\bibitem[{Zhou \textit{et~al.}(2016)Zhou, Khosla, Lapedriza, Oliva, and
  Torralba}]{zhou2016IEEEConf.Comput.Vis.PatternRecognit.CVPR}
Zhou, B., Khosla, A., Lapedriza, A., Oliva, A., Torralba, A.: Learning {{Deep
  Features}} for {{Discriminative Localization}}. In: The {{IEEE Conference}}
  on {{Computer Vision}} and {{Pattern Recognition}} ({{CVPR}}) (2016)

\bibitem[{Zhou \textit{et~al.}(2018)Zhou, Lapedriza, Khosla, Oliva, and
  Torralba}]{zhou2018IEEETrans.PatternAnal.Mach.Intell.}
Zhou, B., Lapedriza, A., Khosla, A., Oliva, A., Torralba, A.: Places: {{A}} 10
  million image database for scene recognition. IEEE Transactions on Pattern
  Analysis and Machine Intelligence  (2018)

\end{thebibliography}

\clearpage
\section*{Appendix}

\appendix

\section{Pseudocode of DebiAN}
\label{sub.supp.pseudo_code}

We present the pseudocode of DebiAN for two tasks -- 1) discover the unknown biases (Alg.~\ref{alg.discover}); 2) mitigate the unknown biases (Alg.~\ref{alg.mitigate}). To ensure that the sampled images have the same target attribute labels, we select images with the same target attribute label in a batch to compute the loss (line 3-7 in Alg.~\ref{alg.discover},~\ref{alg.mitigate}, and line 10-14 in Alg.~\ref{alg.mitigate}).

\begin{algorithm}
  \DontPrintSemicolon
    \KwInput{$C$: trained \textit{classifier}, $T$: number of iterations, $K$: number of target attribute classes}
    \KwOutput{$D$: \textit{discoverer}}
    \KwData{$\mathcal{D}$: training set}
    \For{$t : 1 \dots T$}
    {
      $\mathcal{B} \coloneqq \{ (\mathbf{I}_i, y_i) \}_{i=1}^{N} \sim \mathcal{D}$    \tcp*{Sample a batch $\mathcal{B}$ with $N$ pairs of images $\mathbf{I}_i$ and target attribute labels $y_i$}
      \tcc{for each target attribute class $t$}
      \For{$k : 1 \dots K$}
      {
        $\mathcal{B}_k \coloneqq \{ (\mathbf{I}_j, y_j) \mid y_j = k , (\mathbf{I}_j, y_j) \in \mathcal{B} \}_{j=1}^{M}$  \tcp*{Select $M$ pairs from $\mathcal{B}$ whose labels are $k$}
        $p(\hat{y} \mid \mathbf{I}_j) \coloneqq C(\mathbf{I}_j), \mathbf{I}_j \in \mathcal{B}_k$ \tcp*{$C$ predicts target attribute}
        $p(\hat{b} \mid \mathbf{I}_j) \coloneqq D(\mathbf{I}_j), \mathbf{I}_j \in \mathcal{B}_k$ \tcp*{$D$ predicts bias attribute groups}
        $\mathcal{L}_k = \mathcal{L}_\text{EOV} + \mathcal{L}_\text{UA}$ \tcp*{Compute loss on $\mathcal{B}_t$.}
      }
      update $D$ with loss $\nicefrac{1}{K} \sum_{k=1}^K \mathcal{L}_k$
    }
  \caption{Discover unknown biases.}
  \label{alg.discover}
\end{algorithm}

\begin{algorithm}[h]
  \DontPrintSemicolon
    \KwInput{$T$: number of iterations, $K$: number of target attribute classes}
    \KwOutput{$C$: \textit{classifier}, $D$: \textit{discoverer}}
    \KwData{$\mathcal{D}$: training set}
    \For{$t : 1 \dots T$}
    {
      \tcc{======== Start: optimize $C$, freeze $D$ ========}
      $\mathcal{B} \coloneqq \{ (\mathbf{I}_i, y_i) \}_{i=1}^{N} \sim \mathcal{D}$    \tcp*{Sample a batch $\mathcal{B}$ with $N$ pairs of images $\mathbf{I}_i$ and target attribute labels $y_i$}
      \tcc{for each target attribute class $t$}
      \For{$k : 1 \dots K$}
      {
        $\mathcal{B}_k \coloneqq \{ (\mathbf{I}_j, y_j) \mid y_j = k , (\mathbf{I}_j, y_j) \in \mathcal{B} \}_{j=1}^{M}$  \tcp*{Select $M$ pairs from $\mathcal{B}$ whose labels are $k$}
        $p(\hat{y} \mid \mathbf{I}_j) \coloneqq C(\mathbf{I}_j), \mathbf{I}_j \in \mathcal{B}_k$ \tcp*{$C$ predicts target attribute}
        $p(\hat{b} \mid \mathbf{I}_j) \coloneqq D(\mathbf{I}_j), \mathbf{I}_j \in \mathcal{B}_k$ \tcp*{$D$ predicts bias attribute groups}
        $\mathcal{L}_k^C = \mathcal{L}_\text{RCE}$ \tcp*{Compute loss on $\mathcal{B}_k$}
      }
      update $C$ with loss $\nicefrac{1}{K} \sum_{k=1}^K  \mathcal{L}_k^C$\;
      \tcc{======== End: optimize $C$, freeze $D$ ========}

      \tcc{======== Start: optimize $D$, freeze $C$ ========}
      $\mathcal{B} \coloneqq \{ (\mathbf{I}_i, y_i) \}_{i=1}^{N} \sim \mathcal{D}$    \tcp*{Sample a batch $\mathcal{B}$ with $N$ pairs of images $\mathbf{I}_i$ and target attribute labels $y_i$}
      \tcc{for each target attribute class $k$}
      \For{$k : 1 \dots K$}
      {
        $\mathcal{B}_k \coloneqq \{ (\mathbf{I}_j, y_j) \mid y_j = k , (\mathbf{I}_j, y_j) \in \mathcal{B} \}_{j=1}^{M}$  \tcp*{Select $M$ pairs from $\mathcal{B}$ whose labels are $k$}
        $p(\hat{y} \mid \mathbf{I}_j) \coloneqq C(\mathbf{I}_j), \mathbf{I}_j \in \mathcal{B}_k$ \tcp*{$C$ predicts target attribute}
        $p(\hat{b} \mid \mathbf{I}_j) \coloneqq D(\mathbf{I}_j), \mathbf{I}_j \in \mathcal{B}_k$ \tcp*{$D$ predicts bias attribute groups}
        $\mathcal{L}_k^D = \mathcal{L}_\text{EOV} + \mathcal{L}_\text{UA}$ \tcp*{Compute loss on $\mathcal{B}_k$}
      }
      update $D$ with loss $\nicefrac{1}{K} \sum_{k=1}^K \mathcal{L}_k^D$\;
      \tcc{======== End: optimize $D$, freeze $C$ ========}
    }
  \caption{Mitigate unknown biases.}
  \label{alg.mitigate}
\end{algorithm}

\section{Implementation Details}
\label{sec.supp.implementation_details}
In DebiAN, the \textit{discoverer} and \textit{classifier} use the same architecture but do not share the parameters.

On Multi-Color MNIST dataset (Sec.~4.1), we follow the same setting used in LfF~\cite{nam2020Adv.NeuralInf.Process.Syst.}'s experiment on Colored MNIST. We use Adam~\cite{kingma2015Int.Conf.Learn.Represent.} optimizer with $10^{-3}$ learning rate and 256 batch size. We use an MLP with three hidden layers (obtained from the LfF's official code\footnote{\url{https://github.com/alinlab/LfF}}). All models are trained for 100 epochs.

In the experiments for gender bias mitigation on CelebA~\cite{liu2015IEEEInt.Conf.Comput.Vis.ICCV} dataset (Sec.~4.2), we follow most of the settings used in LfF. We use ResNet-18~\cite{he2016IEEEConf.Comput.Vis.PatternRecognit.CVPR} as the network architecture. We use horizontal flip for data augmentation during training. We use Adam optimizer with $10^{-4}$ learning rate and 256 batch size. All models are trained for 50 epochs. The only difference is that we use CelebA's validation set to choose the epoch where models achieve the best validation set accuracy and report the results on the testing set. Note that validation set accuracy does not use any bias attribute labels because unsupervised debiasing should not rely on any labels of bias attributes. LfF directly reports the results at the 50 epoch on the validation set, which is hard to be replicated as reported by other users in their official code GitHub repository\footnote{\url{https://github.com/alinlab/LfF/issues/2}}.

In the experiments for gender bias mitigation on bFFHQ dataset~\cite{kim2021IEEEInt.Conf.Comput.Vis.ICCVa}, we use the same setting in \cite{kim2021IEEEInt.Conf.Comput.Vis.ICCVa}. We use Adam as the optimizer with 256 batch size. All models are trained for 200 epochs. We use ResNet-18 as the backbone. We notice that \citet{lee2021Adv.NeuralInf.Process.Syst.} use a different setting on bFFHQ dataset with StepLR for the learning rate scheduling, which is more complicated than the one in the original paper~\cite{kim2021IEEEInt.Conf.Comput.Vis.ICCVa}. Thus, we choose the former one as the setting on bFFHQ dataset.

In the experiment of mitigating multiple biases in gender classifier on CelebA dataset (Sec.~4.2), we choose 64 as the batch size and ResNet-50 as the backbone of classifiers. All models are trained with 50 epochs. We use CelebA's validation set to choose the epoch that has the best validation set accuracy for each method. We report the results on the testing set. We use Adam as the optimizer with $10^{-4}$ learning rate.

On Biased Action Recognition (BAR)~\cite{nam2020Adv.NeuralInf.Process.Syst.} dataset, we use the setting in \cite{nam2020Adv.NeuralInf.Process.Syst.}. We use Adam as the optimizer with $10^{-4}$ learning rate. The batch size is 256. We use 224 $\times$ 224 random cropping for data augmentation. All models are trained with 90 epochs.

In the scene classification task, we choose ResNet-18 as the backbone of classifiers and 128 as the batch size. We use Adam as the optimizer with $10^{-4}$ learning rate. All models are trained only on the Places~\cite{zhou2018IEEETrans.PatternAnal.Mach.Intell.} dataset for 50 epochs. We choose the epoch where the model achieves the best accuracy on Places's validation set and report the results on LSUN's~\cite{yu2016ArXiv150603365Cs} validation set.

The code is based on PyTorch~\cite{paszke2019Adv.NeuralInf.Process.Syst.}. We modify LfF's code\footnote{\url{https://github.com/alinlab/LfF/blob/master/make_dataset.py}} that generates Colored MNIST to create Multi-Color MNIST dataset.

In implementation, we add $\epsilon = 10^{-6}$ to the denominators of $\bar{P}_{b^+}(\hat{y})$ and $\bar{P}_{b^-}(\hat{y})$ (\cref{eq.weighted_predicted_prob}) to avoid zero division.

For the \textit{discoverer} $D$, we choose two different implementations for different numbers of classes of the target attribute. When the target attribute is binary (\eg, experiments in Sec.~4.2), \ie number of classes is two, $D$ predicts one value for each image, which is the predicted bias attribute group. We denote this implementation as ``global'' since two classes globally share the predicted bias attribute groups. When the target attribute has $c > 2$ classes, \eg, ten classes in the digit classification, action recognition, and scene classification in Sec.~4.1 and Sec.~4.3, $D$ predicts $c$ values, where each value is the predicted bias attribute group of the corresponding target attribute class. We denote this implementation as ``per class'' since $D$ predicts bias attribute groups for each target attribute class. We provide an ablation study on this in \cref{subsec.supp.global_vs_per_cls}.

\section{Ablation Study}

\begin{table}[t]
  \caption{Ablation study on Unbalanced Assignment (UA) penalty (\ie, $\mathcal{L}_\text{UA}$) in mitigating gender bias of \texttt{Blond Hair} classifier on CelebA~\cite{liu2015IEEEInt.Conf.Comput.Vis.ICCV} dataset}
  \label{tab.ablate_ua_blond_cls}
  \centering
  \begin{tabular}{@{}ccc@{}}
  \toprule
                   & w/o $\mathcal{L}_\text{UA}$   & DebiAN             \\ \midrule
  Avg group Acc.   & 79.6\tiny{$\pm$}1.7 & \textbf{84.0}\tiny{$\pm$}1.4 \\
  Worst group Acc. & 38.5\tiny{$\pm$}4.7 & \textbf{52.9}\tiny{$\pm$}4.7 \\ \bottomrule
  \end{tabular}
\end{table}

\begin{table}[t]
  \caption{Ablation study on Unbalanced Assignment (UA) penalty (\ie, $\mathcal{L}_\text{UA}$) in mitigating multiple biases of gender classifier on CelebA~\cite{liu2015IEEEInt.Conf.Comput.Vis.ICCV} dataset}
  \label{tab.ablate_ua_gender}
  \centering
  \begin{tabular}{@{}cccc@{}}
    \toprule
    bias attribute                    & metric           & w/o $\mathcal{L}_\text{UA}$   & DebiAN             \\ \midrule
    \multirow{2}{*}{\texttt{Wearing Lipstick}} & Avg. Group Acc.  & 87.7\tiny{$\pm$}0.4 & \textbf{88.5}\tiny{$\pm$}1.1 \\
                                      & Worst Group Acc. & 58.1\tiny{$\pm$}1.2 & \textbf{61.7}\tiny{$\pm$}4.2 \\ \midrule
    \multirow{2}{*}{\texttt{Heavy Makeup}}     & Avg. Group Acc.  & 85.6\tiny{$\pm$}1.2 & \textbf{87.8}\tiny{$\pm$}1.3 \\
                                      & Worst Group Acc. & 46.9\tiny{$\pm$}5.2 & \textbf{56.0}\tiny{$\pm$}5.2 \\ \bottomrule
  \end{tabular}
\end{table}

\subsection{Unbalanced Assignment (UA) penalty}

Here we show more ablation study results on the Unbalanced Assignment (UA) penalty on CelebA dataset. The results are shown in \cref{tab.ablate_ua_blond_cls,tab.ablate_ua_gender}, which further proves that $\mathcal{L}_\text{UA}$ can improve fairness results by avoiding the trivial solution---assigning all images into a single bias group (see Unbalanced Assignment (UA) penalty in Sec.~3.1).

\begin{table}[t]
  \caption{Ablation study on different batch sizes on Multi-Color MNIST dataset. We report the accuracy results for images that are both bias-conflicting \wrt \texttt{left color} and \texttt{right color} bias attributes. We also report the unbiased accuracy results. DebiAN achieves better debiasing results under all batch sizes}
  \label{tab.ablate_batch_size}
  \centering
  \begin{tabular}{@{}cccc@{}}
    \toprule
    batch size           &                       & vanilla    & \textbf{DebiAN (Ours)} \\ \midrule
    \multirow{2}{*}{32}  & both bias-conflicting & 8.0\tiny{$\pm$}0.5  & \textbf{18.9}\tiny{$\pm$}1.2     \\
                         & unbiased accuracy     & 61.7\tiny{$\pm$}1.0 & \textbf{75.0}\tiny{$\pm$}0.8     \\ \midrule
    \multirow{2}{*}{64}  & both bias-conflicting & 8.2\tiny{$\pm$}1.9  & \textbf{18.1}\tiny{$\pm$}0.8     \\
                         & unbiased accuracy     & 61.7\tiny{$\pm$}1.4 & \textbf{74.2}\tiny{$\pm$}0.4     \\ \midrule
    \multirow{2}{*}{128} & both bias-conflicting & 5.6\tiny{$\pm$}1.8  & \textbf{17.2}\tiny{$\pm$}1.2     \\
                         & unbiased accuracy     & 58.7\tiny{$\pm$}2.5 & \textbf{72.1}\tiny{$\pm$}0.7     \\ \midrule
    \multirow{2}{*}{256} & both bias-conflicting & 5.2\tiny{$\pm$}0.4  & \textbf{16.0}\tiny{$\pm$}1.8     \\
                         & unbiased accuracy     & 57.4\tiny{$\pm$}0.7 & \textbf{72.0}\tiny{$\pm$}0.8     \\ \midrule
    \multirow{2}{*}{512} & both bias-conflicting & 4.8\tiny{$\pm$}1.0  & \textbf{12.5}\tiny{$\pm$}1.9     \\
                         & unbiased accuracy     & 56.1\tiny{$\pm$}1.3 & \textbf{70.1}\tiny{$\pm$}1.1     \\ \bottomrule
    \end{tabular}
\end{table}

\subsection{Batch Size}
\label{subsec.supp.ablate_batch_size}

In practice, $\{\mathbf{I}_i\}_{i=1}^n$ (defined in Sec.~3) is a mini-batch of images sampled from the dataset for optimizing the networks. One may have the concern that the sampled batch may not have enough images from different bias groups for the \textit{discoverer} to assign. Therefore, we conduct an ablation study on different batch sizes on Multi-Color MNIST dataset with the same setting introduced in Sec~4.1, where the ratio of the \texttt{left color} is 0.99 and \texttt{right color} is 0.95. We report the accuracy results for images that are both bias-conflicting \wrt \texttt{left color} and \texttt{right color} bias attributes (see ``both bias-conflicting'' in \cref{tab.ablate_batch_size}). We also report the unbiased accuracy results.
The results in \cref{tab.ablate_batch_size} show that DebiAN can achieve better debiasing results under different batch sizes compared with the vanilla model.

\begin{table}[t]
  \caption{Ablation study on the ratios on Multi-Color MNIST dataset. Top-1 accuracy results are bolded, and the lowest accuracy results are underlined}
  \label{tab.ablate_ratio_multi_color}
  \centering
  \begin{adjustbox}{width=\linewidth}
  \begin{tabular}{@{}ccccccc@{}}
  \toprule
  left color       & right color      & \multirow{2}{*}{vanilla} & \multirow{2}{*}{LfF} & \multirow{2}{*}{EIIL} & \multirow{2}{*}{PGI} & \multirow{2}{*}{\textbf{DebiAN (Ours)}} \\
  ratio = 0.995     & ratio = 0.95      &                          &                      &                       &                      &                                         \\ \midrule
  bias-aligned     & bias-aligned     & \textbf{100.0}\tiny{$\pm$}0.0      & {\ul 96.3\tiny{$\pm$}0.5}      & \textbf{100.0}\tiny{$\pm$}0.0   & \textbf{100.0}\tiny{$\pm$}0.0  & \textbf{100.0}\tiny{$\pm$}0.0                     \\
  bias-aligned     & bias-conflicting & \textbf{98.7}\tiny{$\pm$}0.6       & {\ul 7.6\tiny{$\pm$}1.0}       & 98.4\tiny{$\pm$}0.2             & 92.2\tiny{$\pm$}11.1           & 98.1\tiny{$\pm$}0.4                               \\
  bias-conflicting & bias-aligned     & {\ul 6.5\tiny{$\pm$}1.0}           & \textbf{96.5}\tiny{$\pm$}1.6   & 46.8\tiny{$\pm$}0.5             & 27.7\tiny{$\pm$}14.2           & 55.4\tiny{$\pm$}2.1                               \\
  bias-conflicting & bias-conflicting & {\ul 2.0\tiny{$\pm$}0.4}           & 5.9\tiny{$\pm$}1.3             & 7.4\tiny{$\pm$}0.1              & 6.3\tiny{$\pm$}2.4             & \textbf{9.2}\tiny{$\pm$}0.8                       \\ \midrule
  \multicolumn{2}{c}{unbiased}        & 51.8\tiny{$\pm$}0.2                & {\ul 51.6\tiny{$\pm$}0.6}      & 63.2\tiny{$\pm$}0.1             & 56.5\tiny{$\pm$}4.3            & \textbf{65.7}\tiny{$\pm$}0.7                      \\ \midrule
  left color       & right color      & \multirow{2}{*}{vanilla} & \multirow{2}{*}{LfF} & \multirow{2}{*}{EIIL} & \multirow{2}{*}{PGI} & \multirow{2}{*}{\textbf{DebiAN (Ours)}} \\
  ratio = 0.99      & ratio = 0.95      &                          &                      &                       &                      &                                         \\ \midrule
  bias-aligned     & bias-aligned     & \textbf{100.0}\tiny{$\pm$}0.0      & 99.6\tiny{$\pm$}0.5            & \textbf{100.0}\tiny{$\pm$}0.0   & 98.6\tiny{$\pm$}2.3            & \textbf{100.0}\tiny{$\pm$}0.0                     \\
  bias-aligned     & bias-conflicting & 97.1\tiny{$\pm$}0.5                & {\ul 4.7\tiny{$\pm$}0.5}       & \textbf{97.2}\tiny{$\pm$}1.5    & 82.6\tiny{$\pm$}19.6           & 95.6\tiny{$\pm$}0.8                               \\
  bias-conflicting & bias-aligned     & 27.5\tiny{$\pm$}3.6                & \textbf{98.6}\tiny{$\pm$}0.4   & 70.8\tiny{$\pm$}4.9             & {\ul 26.6\tiny{$\pm$}5.5}      & 76.5\tiny{$\pm$}0.7                               \\
  bias-conflicting & bias-conflicting & 5.2\tiny{$\pm$}0.4                 & {\ul 5.1\tiny{$\pm$}0.4}       & 10.9\tiny{$\pm$}0.8             & 9.5\tiny{$\pm$}3.2             & \textbf{16.0}\tiny{$\pm$}1.8                      \\ \midrule
  \multicolumn{2}{c}{unbiased}        & 57.4\tiny{$\pm$}0.7                & {\ul 52.0\tiny{$\pm$}0.1}      & 69.7\tiny{$\pm$}1.0             & 54.3\tiny{$\pm$}4.0            & \textbf{72.0}\tiny{$\pm$}0.8                      \\ \midrule
  left color       & right color      & \multirow{2}{*}{vanilla} & \multirow{2}{*}{LfF} & \multirow{2}{*}{EIIL} & \multirow{2}{*}{PGI} & \multirow{2}{*}{\textbf{DebiAN (Ours)}} \\
  ratio = 0.98      & ratio = 0.95      &                          &                      &                       &                      &                                         \\ \midrule
  bias-aligned     & bias-aligned     & \textbf{100.0}\tiny{$\pm$}0.0      & 99.0\tiny{$\pm$}1.7            & \textbf{100.0}\tiny{$\pm$}0.0   & {\ul 89.0\tiny{$\pm$}19.0}     & \textbf{100.0}\tiny{$\pm$}0.0                     \\
  bias-aligned     & bias-conflicting & 96.6\tiny{$\pm$}1.2                & {\ul 9.7\tiny{$\pm$}0.7}       & 96.0\tiny{$\pm$}0.3             & 78.6\tiny{$\pm$}32.4           & \textbf{97.1}\tiny{$\pm$}0.8                      \\
  bias-conflicting & bias-aligned     & {\ul 64.4\tiny{$\pm$}2.3}          & \textbf{98.3}\tiny{$\pm$}0.9   & 84.0\tiny{$\pm$}0.6             & 69.5\tiny{$\pm$}27.7           & 85.1\tiny{$\pm$}3.4                               \\
  bias-conflicting & bias-conflicting & 12.4\tiny{$\pm$}1.1                & {\ul 11.5\tiny{$\pm$}1.1}      & 16.0\tiny{$\pm$}1.7             & 16.4\tiny{$\pm$}1.1            & \textbf{19.4}\tiny{$\pm$}1.3                      \\ \midrule
  \multicolumn{2}{c}{unbiased}        & 68.3\tiny{$\pm$}1.4                & {\ul 54.6\tiny{$\pm$}0.5}      & 74.0\tiny{$\pm$}0.5             & 63.42\tiny{$\pm$}19.3          & \textbf{75.4}\tiny{$\pm$}0.9                      \\ \midrule
  left color       & right color      & \multirow{2}{*}{vanilla} & \multirow{2}{*}{LfF} & \multirow{2}{*}{EIIL} & \multirow{2}{*}{PGI} & \multirow{2}{*}{\textbf{DebiAN (Ours)}} \\
  ratio = 0.95      & ratio = 0.95      &                          &                      &                       &                      &                                         \\ \midrule
  bias-aligned     & bias-aligned     & \textbf{100.0}\tiny{$\pm$}0.0      & {\ul 93.4\tiny{$\pm$}5.8}      & \textbf{100.0}\tiny{$\pm$}0.0   & \textbf{100.0}\tiny{$\pm$}0.0  & \textbf{100.0}\tiny{$\pm$}0.0                     \\
  bias-aligned     & bias-conflicting & 91.1\tiny{$\pm$}2.3                & {\ul 71.1\tiny{$\pm$}33.7}     & 92.7\tiny{$\pm$}0.5             & 76.5\tiny{$\pm$}17.8           & \textbf{94.7}\tiny{$\pm$}0.9                      \\
  bias-conflicting & bias-aligned     & 87.0\tiny{$\pm$}3.7                & {\ul 69.8\tiny{$\pm$}25.9}     & 90.0\tiny{$\pm$}1.1             & 74.4\tiny{$\pm$}17.7           & \textbf{92.7}\tiny{$\pm$}1.3                      \\
  bias-conflicting & bias-conflicting & 26.0\tiny{$\pm$}1.3                & \textbf{39.6}\tiny{$\pm$}6.9   & 34.7\tiny{$\pm$}3.3             & {\ul 15.8\tiny{$\pm$}4.7}      & \textbf{39.6}\tiny{$\pm$}0.2                      \\ \midrule
  \multicolumn{2}{c}{unbiased}        & 76.0\tiny{$\pm$}1.6                & 68.5\tiny{$\pm$}3.2            & 79.3\tiny{$\pm$}0.7             & {\ul 66.7\tiny{$\pm$}10.0}     & \textbf{81.8}\tiny{$\pm$}0.6                      \\ \bottomrule
  \end{tabular}
\end{adjustbox}
\end{table}

\subsection{Ablation Study on Different Ratios}
\label{subsec.ablate_ratios_multi_color_mnist}

We conduct an ablation study on different ratios of bias-aligned samples in Multi-Color MNIST's training set. We keep the ratio for the \texttt{right color} bias attribute to 0.95 and use different ratios for \texttt{left color} bias attribute, ranging from 0.995 to 0.95. The results are shown in \cref{tab.ablate_ratio_multi_color}. DebiAN achieves better unbiased accuracy results and accuracy results on samples that are bias-conflicting \wrt both bias attributes. The only exception is the accuracy results of the samples that are bias-conflicting \wrt both bias attributes when both ratios are 0.95 (last section in \cref{tab.ablate_ratio_multi_color}). Both LfF and DebiAN achieve 39.6 accuracy results. However, our method achieves a lower standard deviation (0.2) than LfF (6.9) and achieves much better final unbiased results (81.8 vs. 68.5). We also notice that LfF's debiasing results have a large standard deviation when the ratios of both bias attributes are 0.95. We provide an explanation in \cref{subsec.supp.equal_salient}.

\begin{table}[t]
  \caption{Ablation study on alternate training on Multi-Color MNIST dataset}
  \label{tab.supp.ablate_alternate_training}
  \centering
  \begin{tabular}{@{}cccc@{}}
  \toprule
  left   color      & right color       & \multirow{2}{*}{w/o alternate training} & \multirow{2}{*}{DebiAN} \\
  ratio = 0.99       & ratio = 0.95       &                                         &                         \\ \midrule
  bias-aligned      & bias-aligned      & \textbf{100.0}\tiny{$\pm$}0.0                     & \textbf{100.0}\tiny{$\pm$}0.0     \\
  bias-aligned      & bias-conflicting  & \textbf{97.3}\tiny{$\pm$}0.1                      & 95.6\tiny{$\pm$}0.8               \\
  bias-conflicting  & bias-aligned      & 74.0\tiny{$\pm$}0.3                               & \textbf{76.5}\tiny{$\pm$}0.7      \\
  bias-conflicting  & bias-conflicting  & 11.5\tiny{$\pm$}0.7                               & \textbf{16.0}\tiny{$\pm$}1.8      \\ \midrule
  \multicolumn{2}{c}{unbiased accuracy} & 70.7\tiny{$\pm$}0.1                               & \textbf{72.0}\tiny{$\pm$}0.8      \\ \bottomrule
  \end{tabular}
\end{table}

\subsection{Alternate Training}
\label{subsec.supp.alternate_training}

We conduct an ablation study on alternate training on the Multi-Color MNIST dataset with the same setting used in Sec.~4.1. To remove the alternate training from DebiAN, we follow EIIL~\cite{creager2021Int.Conf.Mach.Learn.} and PGI~\cite{ahmed2021Int.Conf.Learn.Represent.} to train the \textit{discoverer} to identify the unknown biases in a \textit{classifier} trained with one epoch. After training \textit{discoverer}, we fix the parameters of \textit{discoverer} and only train the \textit{classifier} to perform debiasing. The results in \cref{tab.supp.ablate_alternate_training} show that alternate training can improve the debiasing results, \eg, higher unbiased accuracy and higher accuracy for samples that are bias-conflicting \wrt both bias attributes (4th row), which demonstrates the necessity of alternate training.

\begin{table}[h]
  \caption{Results of ablation study on \textit{discoverer}'s outputs. Bolded methods are used to report results in the main paper. For binary classification (\ie, age classification on bFFHQ dataset) task, there is no significant difference whether or not the \textit{discoverer} predicts bias attribute groups per class or globally. When it comes to the multi-class digit classification on Multi-Color MNIST dataset, predicting bias attribute groups per class has better accuracy results on the images that are bias-conflicting \wrt both \texttt{left color} and \texttt{right color} bias attributes}
  \label{tab.supp.ablate_per_cls_global}
  \centering
  \begin{tabular}{cccc}
  \toprule
  dataset   & \#classes & per class    & \textbf{DebiAN (global)}                   \\ \midrule
  bFFHQ & 2 & 62.80\tiny{$\pm$}0.60 & 62.87\tiny{$\pm$}0.61     \\ \midrule \midrule
  dataset   & \#classes & global       & \textbf{DebiAN (per class)}                  \\ \midrule
  Multi-Color MNIST & 10   & 13.5\tiny{$\pm$}0.1 & \textbf{16.0}\tiny{$\pm$}1.8 \\ \bottomrule
  \end{tabular}
\end{table}

\subsection{Bias Attribute Groups: Global vs. Per Class}
\label{subsec.supp.global_vs_per_cls}

As mentioned in \cref{sec.supp.implementation_details}, the predicted bias attribute groups from the \textit{discoverer} ($D$) are shared by both classes in the binary classification task. In the multi-class classification setting (\eg, digit classification, scene classification task, \etc), $D$ predicts binary bias group assignments for each class. We justify our implementation choices with the results in \cref{tab.supp.ablate_per_cls_global}.

For the binary age classification on bFFHQ dataset, there is no significant difference between the two implementation choices (\ie, differences are within error bars). Therefore, we choose ``global'' \textit{discoverer} for the binary classification task due to its simplicity.

However, in the multi-class digit classification on Multi-Color MNIST dataset, we do observe the better result produced by the \textit{discoverer} that predicts bias group assignment for each class (\ie, improvement is greater than the error bar). We suspect that predicting bias attribute groups per class in the binary classification task is redundant because the binary target attribute is spuriously correlated with the binary bias attribute. For example, if the target attribute \texttt{age} is spuriously correlated with the bias attribute \texttt{gender}, \ie, more young females and old males than old females and young males, then it is not necessary to predict bias attribute group for both genders since both genders share the same bias attribute groups. However, this may not be the case for multi-class settings. For example, in Multi-Color MNIST dataset, each digit class is spuriously correlated with a unique left color, \eg, for bias-aligned samples, digit class 0's left color is red but digit class 1's left color is yellow (Fig.~4). In other words, the bias attribute values may not be shared globally across different target attribute classes. Therefore, we choose different numbers of outputs for the \textit{discoverer} under different tasks.

\begin{table}[t]
  \caption{Results of an alternative design for debiasing in DebiAN -- $D$ and $C$ play the minmax game (see DebiAN (minmax)) on Multi-Color MNIST dataset. Our RCE loss significantly outperforms the alternative design}
  \label{tab.supp.ablate_minmax}
  \centering
  \begin{tabular}{@{}cccc@{}}
  \toprule
  left color       & right color      & \multirow{2}{*}{DebiAN (minmax)} & \multirow{2}{*}{\textbf{DebiAN (RCE)}} \\
  ratio = 0.99      & ratio = 0.95      &                          &                                \\ \midrule
  bias-aligned     & bias-aligned     & 97.3\tiny{$\pm$}4.6                & \textbf{100.0}\tiny{$\pm$}0.0            \\
  bias-aligned     & bias-conflicting & \textbf{95.7\tiny{$\pm$}0.8}       & 95.6\tiny{$\pm$}0.8                      \\
  bias-conflicting & bias-aligned     & 64.5\tiny{$\pm$}2.0                & \textbf{76.5}\tiny{$\pm$}0.7             \\
  bias-conflicting & bias-conflicting & 7.7\tiny{$\pm$}1.9                 & \textbf{16.0}\tiny{$\pm$}1.8             \\ \midrule
  \multicolumn{2}{c}{unbiased accuracy}        & 66.3\tiny{$\pm$}0.9                & \textbf{72.0}\tiny{$\pm$}0.8             \\ \bottomrule
  \end{tabular}
\end{table}

\subsection{Alternative Design for Debiasing: $\max_C \mathcal{L}_\text{EOV}$}

One may consider an alternative design for debiasing---train the \textit{classifier} $C$ to maximize the Equal Opportunity Violation (EOV) loss, or formally $\max_C \mathcal{L}_\text{EOV}$. This alternative design, to some degree similar to GAN~\cite{goodfellow2014Adv.NeuralInf.Process.Syst.}'s training strategy, may look more ``unified'' since it lets the \textit{discoverer} $D$ and \textit{classifier} $C$ play the minmax game:
\begin{equation}
  \min_C \max_D |\bar{P}_{b^+}(\hat{y}) - \bar{P}_{b^-}(\hat{y})|,
  \label{eq.supp.minmax}
\end{equation}
where $\bar{P}_{b^+}(\hat{y})$ and $\bar{P}_{b^-}(\hat{y})$ are defined in Eq.~3. This alternative design enables $C$ to \textit{directly} meet the Equal Opportunity~\cite{hardt2016Adv.NeuralInf.Process.Syst.,pleiss2017Adv.NeuralInf.Process.Syst.} fairness criterion. More concretely, we implement this alternative design of $C$'s objectives by the following loss function:
\begin{align}
  \min_C -\log \left(  1 - \left| \bar{P}_{b^+}(\hat{y}) - \bar{P}_{b^-}(\hat{y}) \right| \right) + \text{CE}(p_t(\mathbf{I}_i), y_i),
\end{align}
where the first $-\log$ term implements $C$'s objective in the minmax game (\cref{eq.supp.minmax}) and the second term $\text{CE}$ is the standard cross-entropy loss. We conduct an ablation study on the design for debiasing (\ie, playing minmax game vs. RCE loss (Eq.~6)) and the results on Multi-Color MNIST dataset are shown in Tab.~\ref{tab.supp.ablate_minmax}. The results demonstrate that our RCE loss performs much better than the alternative design. We suspect the reason is that $C$ in DebiAN has two goals -- 1) fooling the \textit{discoverer} to achieve fairer results; 2) achieving higher accuracy by optimizing the standard cross-entropy loss, which is different from GAN~\cite{goodfellow2014Adv.NeuralInf.Process.Syst.} where the generator only has one goal -- fooling the discriminator to achieve better image quality of the synthesized images. Therefore, it is hard to control the balance between the two goals of $C$ in this alternative design. In contrast, our RCE loss can better incorporate the two goals within a single objective function $\mathcal{L}_\text{RCE}$, leading to better debiasing results.

\begin{table}[t]
  \caption{Results on Colored MNIST (foreground color) under different ratios of bias-aligned samples in the training set. We bold top-1 results (except LfF's results reported in the original paper since they cannot be replicated by their officially released code) and underline the lowest results based on the mean value. Although LfF achieves better bias-conflicting accuracy, it achieves lower bias-aligned accuracy, resulting in low unbiased accuracy. Overall, DebiAN achieves comparable or slightly lower unbiased accuracy results compared with other methods}
  \label{tab.fg_colored_mnist}
  \centering
  \begin{adjustbox}{width=\linewidth}
  \begin{tabular}{@{}cccccccc@{}}
  \toprule
  \multirow{2}{*}{ratio} & \multirow{2}{*}{} & \multirow{2}{*}{vanilla} & LfF                                         & LfF                                               & \multirow{2}{*}{EIIL} & \multirow{2}{*}{PGI} & \multirow{2}{*}{\textbf{DebiAN (Ours)}} \\
  &                   &                          & \multicolumn{1}{l}{(reported in the paper)} & \multicolumn{1}{l}{(replicate via official code)} &                       &                      &                                         \\ \midrule
  \multirow{3}{*}{0.995} & bias-aligned      & 100.00\tiny{$\pm$}0.00         & -           & {\ul 54.13\tiny{$\pm$}6.33}                 & 99.90\tiny{$\pm$}0.01       & 99.86\tiny{$\pm$}0.11          & \textbf{100.00}\tiny{$\pm$}0.00  \\
                         & bias-conflicting  & 7.92\tiny{$\pm$}4.68           & 63.49\tiny{$\pm$}1.94 & \textbf{57.11}\tiny{$\pm$}6.22              & {\ul 24.63\tiny{$\pm$}0.37} & 27.01\tiny{$\pm$}5.49          & 24.83\tiny{$\pm$}1.83            \\
                         & unbiased accuracy & {\ul 53.96\tiny{$\pm$}2.34}    & 63.39\tiny{$\pm$}1.97 & 55.62\tiny{$\pm$}6.26                       & 62.27\tiny{$\pm$}1.85       & \textbf{64.44}\tiny{$\pm$}2.78 & 62.41\tiny{$\pm$}0.91            \\ \midrule
  \multirow{3}{*}{0.99}  & bias-aligned      & \textbf{99.97}\tiny{$\pm$}0.05 & -           & {\ul 61.96\tiny{$\pm$}3.29}                 & 99.81\tiny{$\pm$}0.19       & 99.86\tiny{$\pm$}0.07          & 99.86\tiny{$\pm$}0.05            \\
                         & bias-conflicting  & {\ul 18.73\tiny{$\pm$}2.78}    & 74.19\tiny{$\pm$0.94}  & \textbf{67.20}\tiny{$\pm$}4.58              & 40.71\tiny{$\pm$}1.93       & 41.88\tiny{$\pm$}0.99          & 43.33\tiny{$\pm$}0.86            \\
                         & unbiased accuracy & {\ul 59.35\tiny{$\pm$}1.40}    & 74.01\tiny{$\pm$}2.21 & 64.58\tiny{$\pm$}2.22                       & 70.26\tiny{$\pm$}1.06       & 70.87\tiny{$\pm$}0.53          & \textbf{71.60}\tiny{$\pm$}0.44   \\ \midrule
  \multirow{3}{*}{0.98}  & bias-aligned      & \textbf{99.80}\tiny{$\pm$}0.16 & -           & {\ul 73.11\tiny{$\pm$}5.41}                 & 99.77\tiny{$\pm$}0.14       & 99.73\tiny{$\pm$}0.11          & 99.76\tiny{$\pm$}0.05            \\
                         & bias-conflicting  & {\ul 39.23\tiny{$\pm$}1.63}    & 80.67\tiny{$\pm$}0.56 & \textbf{78.23}\tiny{$\pm$}1.56              & 54.44\tiny{$\pm$}1.26       & 57.46\tiny{$\pm$}1.13          & 55.46\tiny{$\pm$}0.71            \\
                         & unbiased accuracy & {\ul 69.52\tiny{$\pm$}0.90}    & 80.48\tiny{$\pm$}0.45 & 75.67\tiny{$\pm$}2.95                       & 77.10\tiny{$\pm$}0.55       & \textbf{78.60}\tiny{$\pm$}0.55 & 77.61\tiny{$\pm$}0.37            \\ \midrule
  \multirow{3}{*}{0.95}  & bias-aligned      & 99.60\tiny{$\pm$}0.17          & -           & {\ul 71.67\tiny{$\pm$}0.68}                 & 99.61\tiny{$\pm$}0.25       & 99.37\tiny{$\pm$}0.31          & \textbf{99.70}\tiny{$\pm$}0.17   \\
                         & bias-conflicting  & 70.99\tiny{$\pm$}2.45          & 85.77\tiny{$\pm$}0.66 & \textbf{82.37}\tiny{$\pm$}1.49              & 73.01\tiny{$\pm$}1.03       & {\ul 70.63\tiny{$\pm$}2.24}    & 73.04\tiny{$\pm$}2.20            \\
                         & unbiased accuracy & 85.30\tiny{$\pm$}1.30          & 85.39\tiny{$\pm$}0.94 & {\ul 77.02\tiny{$\pm$}4.11}                 & 86.31\tiny{$\pm$}0.46       & 85.00\tiny{$\pm$}1.05          & \textbf{86.37}\tiny{$\pm$}1.10   \\ \bottomrule
  \end{tabular}
  \end{adjustbox}
\end{table}

\begin{table}[t]
  \caption{Results on Colored MNIST (background color) under different ratios of bias-aligned samples in the training set. We bold top-1 results and underline the lowest results based on the mean value. Similar to the results in Colored MNIST (foreground color) (\cref{tab.fg_colored_mnist}), although LfF achieves better bias-conflicting accuracy, it achieves lower bias-aligned accuracy, resulting in low unbiased accuracy. Overall, DebiAN achieves comparable or slightly higher unbiased accuracy results compared with other methods}
  \label{tab.bg_colored_mnist}
  \centering
  \begin{adjustbox}{width=\linewidth}
  \begin{tabular}{@{}ccccccc@{}}
  \toprule
  ratio                  &                   & vanilla               & LfF                  & EIIL                  & PGI                   & \textbf{DebiAN (Ours)} \\ \midrule
  \multirow{3}{*}{0.995} & bias-aligned      & 99.97\tiny{$\pm$}0.05           & {\ul 42.98\tiny{$\pm$}1.67}    & \textbf{100.00}\tiny{$\pm$}0.00 & \textbf{100.00}\tiny{$\pm$}0.00 & \textbf{100.00}\tiny{$\pm$}0.00  \\
                         & bias-conflicting  & {\ul 1.98\tiny{$\pm$}0.35}      & \textbf{61.37}\tiny{$\pm$}1.69 & 14.16\tiny{$\pm$}4.89           & 9.95\tiny{$\pm$}3.28            & 20.94\tiny{$\pm$}3.92            \\
                         & unbiased accuracy & {\ul 50.98\tiny{$\pm$}0.18}     & 52.17\tiny{$\pm$}1.62          & 57.08\tiny{$\pm$}2.44           & 54.98\tiny{$\pm$}1.64           & \textbf{60.47}\tiny{$\pm$}1.96   \\ \midrule
  \multirow{3}{*}{0.99}  & bias-aligned      & \textbf{100.00}\tiny{$\pm$}0.00 & {\ul 52.24\tiny{$\pm$}1.84}    & 99.97\tiny{$\pm$}0.05           & 99.26\tiny{$\pm$}1.10           & \textbf{100.00}\tiny{$\pm$}0.00  \\
                         & bias-conflicting  & {\ul 5.43\tiny{$\pm$}0.32}      & \textbf{67.94}\tiny{$\pm$}0.88 & 35.71\tiny{$\pm$}1.78           & 12.89\tiny{$\pm$}1.62           & 42.57\tiny{$\pm$}0.36            \\
                         & unbiased accuracy & {\ul 52.75\tiny{$\pm$}0.16}     & 60.09\tiny{$\pm$}1.26          & 67.84\tiny{$\pm$}0.90           & 56.07\tiny{$\pm$}0.81           & \textbf{71.29}\tiny{$\pm$}0.18   \\ \midrule
  \multirow{3}{*}{0.98}  & bias-aligned      & \textbf{99.97}\tiny{$\pm$}0.05  & {\ul 71.90\tiny{$\pm$}4.86}    & 99.90\tiny{$\pm$}0.09           & 97.83\tiny{$\pm$}3.54           & 99.90\tiny{$\pm$}0.00            \\
                         & bias-conflicting  & 21.84\tiny{$\pm$}5.39           & \textbf{80.28}\tiny{$\pm$}0.98 & 51.25\tiny{$\pm$}2.86           & {\ul 18.06\tiny{$\pm$}5.47}     & 53.41\tiny{$\pm$}1.21            \\
                         & unbiased accuracy & 60.91\tiny{$\pm$}2.71           & 76.09\tiny{$\pm$}2.91          & 75.58\tiny{$\pm$}1.46           & {\ul 59.75\tiny{$\pm$}4.02}     & \textbf{76.66}\tiny{$\pm$}0.60   \\ \midrule
  \multirow{3}{*}{0.95}  & bias-aligned      & 99.87\tiny{$\pm$}0.15           & {\ul 69.71\tiny{$\pm$}7.07}    & 99.90\tiny{$\pm$}0.16           & 96.78\tiny{$\pm$}4.90           & \textbf{99.90}\tiny{$\pm$}0.01   \\
                         & bias-conflicting  & 55.14\tiny{$\pm$}2.06           & \textbf{81.67}\tiny{$\pm$}4.54 & 69.17\tiny{$\pm$}3.15           & {\ul 33.90\tiny{$\pm$}15.55}    & 70.70\tiny{$\pm$}0.76            \\
                         & unbiased accuracy & 77.51\tiny{$\pm$}1.09           & 75.69\tiny{$\pm$}5.50          & 84.53\tiny{$\pm$}1.54           & {\ul 65.34\tiny{$\pm$}9.80}     & \textbf{85.30}\tiny{$\pm$}0.37   \\ \bottomrule
  \end{tabular}
  \end{adjustbox}
\end{table}

\begin{figure}[t]
  \centering
  \includegraphics[width=\linewidth]{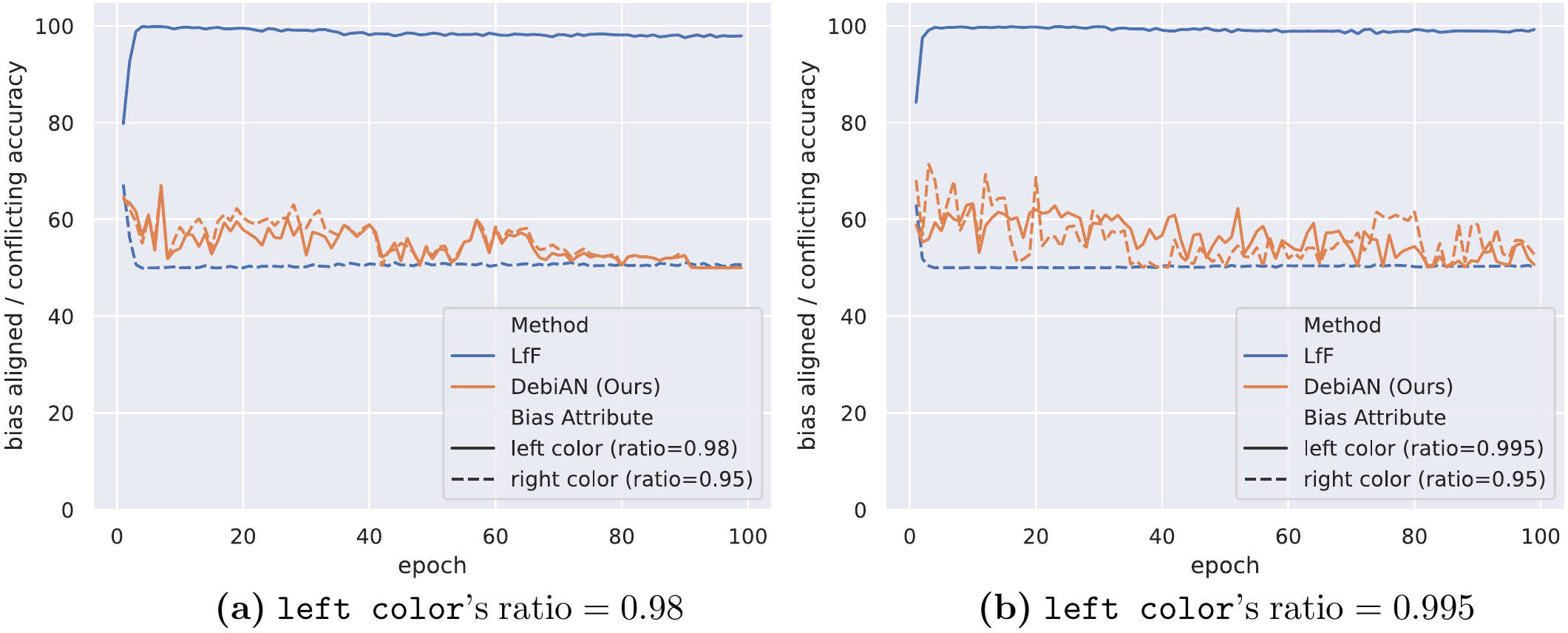}
  \caption{More bias discovery results \wrt \texttt{left color} and \texttt{right color} biases throughout the training epochs on Multi-Color MNIST dataset under different ratios of bias-aligned samples \wrt \texttt{left color}. The results are consistent with \cref{fig.bias_discovery_acc_trend}}
  \label{fig.supp.bias_discovery_trend_diff_ratios}
\end{figure}

\begin{figure}[t]
  \includegraphics[width=\linewidth]{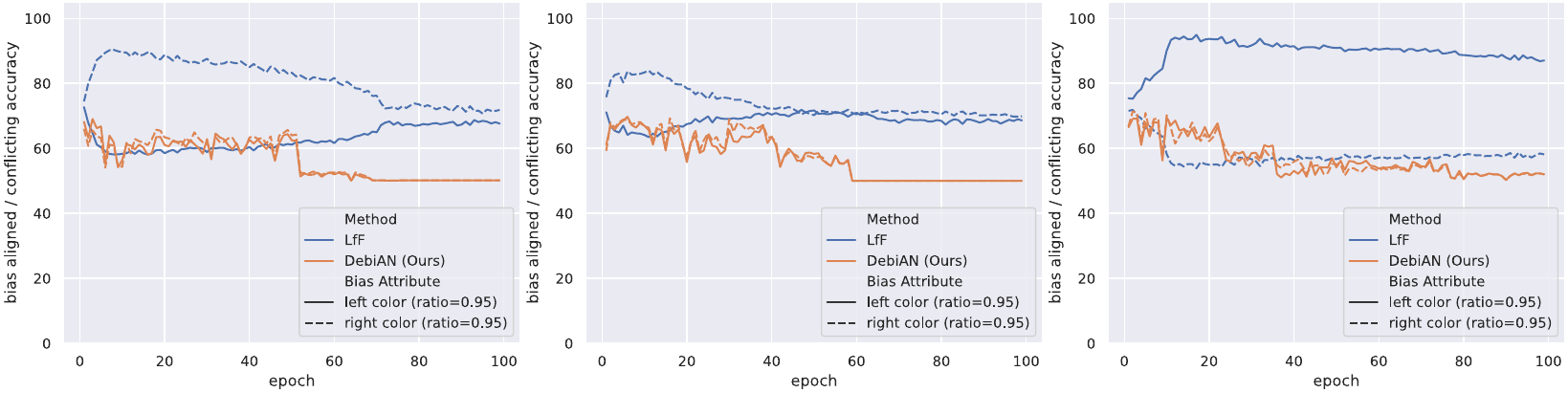}
  \caption{Bias discovery results when \texttt{left color} and \texttt{right color} are equally salient (both ratios are 0.95). The three plots are results under three different random seeds. In the first two plots, LfF mainly discovers \texttt{right color} at the early training stage and gradually discovers both biases. In the third seed, it mainly discovers \texttt{left color} bias. The results show that LfF are unstable in bias discovery when two biases are equally salient. In contrast, our DebiAN consistently discovers both biases at the early stage under different random seeds and gradually converges 50\% bias discovery accuracy as debiasing is performed in the \textit{classifier}}
  \label{fig.supp.bias_discovery_trend_equally_salient}
\end{figure}

\section{Bias Discovery on Multi-Color MNIST}

\subsection{Implementation Details}
\label{subsec.bias_discovery_multi_color_mnist_impl_details}

To evaluate the bias discovery results, we transform the outputs from LfF's ``biased model'' and DebiAN's \textit{discoverer} into bias-aligned / bias-conflicting prediction by the following approaches.

For LfF, since the biased model is trained to amplify the biases, the biased model's outputs predict ten colors aligned with digits. Thus, when its predicted color class is the same as the ground-truth digit class, \eg, $c$-th color for the $c$-th class, we regard its prediction as bias-aligned. Otherwise, its prediction is bias-conflicting.

For DebiAN, we use \textit{discoverer} and \textit{classifier} to predict each image $\mathbf{I}_i$'s predicted bias group assignment $p(\hat{b} \mid \mathbf{I}_i)$ and predicted probability of the ground-truth class of the target attribute $p_t(\mathbf{I}_i)$ on the entire testing set, respectively. Then, we compute the weighted average predicted probabilities $\bar{P}_{b^+} (\hat{y})$ and $\bar{P}_{b^-} (\hat{y})$ (see Eq.~3) in two bias groups. If $\bar{P}_{b^+} (\hat{y}) \geq \bar{P}_{b^-} (\hat{y})$, we use $p(\hat{b} = 1 \mid \mathbf{I}_i)$ as the bias-aligned prediction and $p(\hat{b} = 0 \mid \mathbf{I}_i)$ as the bias-conflicting prediction since the positive bias group has higher weighted average predicted probability, \ie, \textit{classifier} performs better on the positive bias group. If $\bar{P}_{b^+} (\hat{y}) < \bar{P}_{b^-} (\hat{y})$, we use $p(\hat{b} = 1 \mid \mathbf{I}_i)$ as the bias-conflicting prediction and $p(\hat{b} = 0 \mid \mathbf{I}_i)$ as the bias-aligned prediction.

After obtaining bias-aligned and bias-conflicting predictions, we compute the bias-aligned and bias-conflicting accuracy for \texttt{left color} and \texttt{right color} biases as follows. Each testing image has two labels—(1) bias-aligned/bias-conflicting w.r.t. left color; (2) bias-aligned/bias-conflicting w.r.t. right color. Thus, the left color (or right color) bias discovery accuracy is computed based on the bias-aligned/bias-conflicting predictions against the left color (or right color) bias-aligned/bias-conflicting labels.

\subsection{More Bias Discovery Results under Different Ratios}
\label{subsec.supp.ablate_ratios_multi_color}

In the main paper, we evaluate LfF and DebiAN's bias discovery results when the ratio of \texttt{left color} is 0.99 in \cref{fig.bias_discovery_acc_trend}. Here, we show results under more ratios in \cref{fig.supp.bias_discovery_trend_diff_ratios}, where the ratios of \texttt{left color} are 0.98 (\cref{fig.supp.bias_discovery_trend_diff_ratios} (a)) and 0.995 (\cref{fig.supp.bias_discovery_trend_diff_ratios} (b)). The results are consistent with \cref{fig.bias_discovery_acc_trend}---LfF can only discover the more salient \texttt{left color} and cannot identify the less salient \texttt{right color} bias, whereas DebiAN's \textit{discoverer} can discover both biases at the early training stage and the bias discovery accuracy gradually converges to 50\% when debiasing is performed in the \textit{classifier}.

\subsection{Bias Discovery Results under Equally Salient Biases}
\label{subsec.supp.equal_salient}

We further evaluate bias discovery results when \texttt{left color} and \texttt{right color} biases are equally salient, \ie, ratios of both biases are 0.95. We found that LfF shows more unstable results under different random seeds than in previous settings. Therefore, we show bias discovery results under three different random seeds in \cref{fig.supp.bias_discovery_trend_equally_salient}. Under the first two random seeds (left and middle plots in \cref{fig.supp.bias_discovery_trend_equally_salient}), LfF first discovers \texttt{right color} bias and gradually discovers both biases. However, under the third random seed (the right plot in \cref{fig.supp.bias_discovery_trend_equally_salient}), LfF mainly discovers the \texttt{left color} bias. Therefore, our Multi-Color MNIST dataset reveals another weakness of LfF---unstable bias discovery results when two biases are equally salient, which also explains LfF's unstable debiasing results under the equally salient biases (LfF has large error bars, \eg, $\pm$33.7 and $\pm$25.9, of the accuracy results in \cref{tab.ablate_ratio_multi_color}). In contrast, DebiAN's bias discovery results are stable---consistent results across different random seeds and under different ratios.

\subsection{Detailed Discussion on Bias Discovery}
\label{subsec.supp.discussion_bias_discovery}

\subsubsection{Why LfF's bias discovery accuracies do not converge to 50\%?} In \cref{fig.bias_discovery_acc_trend} and \cref{fig.supp.bias_discovery_trend_diff_ratios}, LfF's bias discovery accuracies maintain at about 100\% or 50\% throughout the training epochs. One may wonder why it does not converge to 50\% as DebiAN does. The reason is that DebiAN discovers biases from the \textit{classifier} ($\mathcal{L}_\text{EOV}$ is based on \textit{classifier}'s output), whereas LfF identifies biases from the dataset. Concretely, LfF uses the assumption that the bias attribute is easier than the target attribute to define the bias and uses Generalized Cross-entropy (GCE) loss~\cite{zhang2018Adv.NeuralInf.Process.Syst.} to train a biased model. GCE loss is defined by $- p_t^q \log p_t $, where $p_t$ is bias model's predicted probability of the ground-truth class and $q$ is a hyperparameter. Intuitively, it up-weights easy examples (\ie, high $p_t$) with high weight $p_t^q$ and down-weights hard examples (\ie, low $p_t$) with low weight $p_t^q$. Therefore, the ``biased model'' focuses more on the easy examples in the dataset. However, it does not know any biases in the \textit{classifier} (no classifier's outputs used in GCE). Therefore, whether the classifier is performing debiasing will not affect LfF's bias discovery, making the bias attribute accuracy stays the same throughout the entire training stage. In contrast, DebiAN's \textit{discoverer} actively identifies biases in the \textit{classifier}. Therefore, \textit{discoverer}'s bias discovery results will converge to 50\% as debiasing is performing in the \textit{classifier}, making the \textit{discoverer} harder to find the biases.

\subsubsection{Bias Discovery: EIIL and PGI} Here, we discuss the connection and difference between DebiAN and two previous methods---EIIL and PGI. In contrast to LfF that finds biases from the dataset based on the assumption, all EIIL, PGI, and DebiAN actively find biases from the classifier. However, DebiAN differs from EIIL and PGI in terms of the objective function, network architecture, and training scheme. We mainly introduce EIIL because PGI is a follow-up work for EIIL with a difference in the network architecture.

In terms of objective function, EIIL (and PGI) inversely uses the debiasing objective function---IRMv1~\cite{arjovsky2020ArXiv190702893CsStat}. In other words, while minimizing IRMv1 was designed for debiasing in previous works, EIIL maximizes the gradient norm penalty in IRMv1 to identify biases. However, this is suboptimal for two reasons. First, IRMv1 approximates IRM with gradient norm penalty to make it computationally tractable. However, the zero gradient norm may only indicate a local minimum instead of the global minimum. In contrast, DebiAN's EOV loss uses the principled definition to define the bias---violation of equal opportunity fairness criterion. Second, since the gradient norm does not have an upper bound, maximizing it leads to an optimization problem. As a comparison, DebiAN's EOV loss minimizes a bounded negative log-likelihood (Eq.~2), which is easier to be optimized.

Regarding network architecture, EIIL does not train any networks but directly optimizes a vector $\mathbf{q} \in \mathbb{R}^{N}$ for $N$ images in the training set to maximize IRMv1's gradient norm, which can be regarded as directly optimize the bias group assignments. Since the vector $\mathbf{q}$ is only fitted to the training set, we cannot evaluate EIIL's bias discovery results on the balanced testing set. PGI uses a slightly different approach by training a small MLP that takes the classifier's features as the input and predicts the bias group assignments. PGI's bias discovery objective function is identical with EIIL's---maximizing IRMv1. We use the trained MLP to infer the bias-aligned / bias-conflicting on the testing set on Multi-Color MNIST dataset under the setting used in Sec.~4.1 (ratio \wrt \texttt{left color} is 0.99 and ratio \wrt \texttt{right color} is 0.95) based on a classifier trained with one epoch (explain in the next paragraph). The bias discovery accuracy results \wrt \texttt{left color} and \texttt{right color} are 50.6$\pm$1.9 and 49.9$\pm$0.3. In contrast, DebiAN's \textit{discoverer} has the same network architecture as the classifier to predict bias group assignments from raw images, which enables \textit{discoverer} to learn its own feature of the bias attribute directly from the raw images. As a result, DebiAN achieves about 60\% to 75\% accuracy at the first epoch (see \cref{fig.bias_discovery_acc_trend}). Therefore, we empirically show that DebiAN can discover biases more accurately.

Finally, with respect to the training scheme, by hypothesizing that the classifier learns bias features in the early stage, EIIL and PGI use a fixed classifier trained with one epoch to discover biases. However, this might not be the case in the multi-bias setting where multiple biases may be learned at different training stages. In contrast, DebiAN trains \textit{discoverer} and \textit{classifier} in an alternate fashion, enabling \textit{discoverer} to find biases in the \textit{classifier} during the entire training stage. Our ablation study on the alternate training (\cref{subsec.supp.alternate_training}) further demonstrates its benefits.

\section{Results on Colored MNIST (single-bias setting)}
\label{sec.single_color_mnist_exp}

We also compare with other methods on Colored MNIST in a single-bias setting. There are two variants of Colored MNIST datasets in previous works---(1) adding colors to the foreground (\ie, digit)~\cite{kim2019IEEEConf.Comput.Vis.PatternRecognit.CVPR,li2019IEEEConf.Comput.Vis.PatternRecognit.CVPRa,ahmed2021Int.Conf.Learn.Represent.}; (2) adding colors to the background~\cite{bahng2020Int.Conf.Mach.Learn.,reddy2021Thirty-FifthConf.NeuralInf.Process.Syst.DatasetsBenchmarksTrackRound1}. Therefore, we conduct experiment on both variants of Colored MNIST dataset. We denote the Colored MNIST with foreground color as ``Colored MNIST (foreground)'' and the Colored MNIST with background color as ``Colored MNIST (background).'' Same with the experiment setting on Multi-Color MNIST, we follow the setting used in LfF~\cite{nam2020Adv.NeuralInf.Process.Syst.}, including using MLP as the network architecture, training with 100 epochs, \etc. Same with LfF, we report the results under four ratios of bias-aligned samples in the training set---0.995, 0.99, 0.98, and 0.95. In terms of evaluation metrics, we follow LfF to report the accuracy results on bias-conflicting samples and unbiased accuracy. We additionally report the accuracy results on bias-aligned samples.

We report LfF's reported results on Colored MNIST (foreground). Besides, we also replicate LfF's results via their officially released code. We cannot replicate their reported results. This issue is also reported in the official code's GitHub repository by other people\footnote{\url{https://github.com/alinlab/LfF/issues/1}}.

The results on Colored MNIST (foreground color) are in \cref{tab.fg_colored_mnist}. Although LfF achieves better bias-conflicting accuracy results, it also achieves much lower bias-aligned accuracy, revealing that LfF's reweighing method overly focuses on the bias-conflicting samples than the bias-aligned samples. As a result, LfF achieves low unbiased accuracy. Overall, DebiAN achieves comparable or slightly lower unbiased accuracy results.

The results on Colored MNIST (background color) are in \cref{tab.bg_colored_mnist}. Similar to the results on Colored MNIST (foreground color), LfF achieves better bias-conflicting accuracy but low bias-aligned accuracy and unbiased accuracy. Overall, DebiAN achieves comparable or slightly better unbiased accuracy results.

We also notice that PGI achieves inconsistent results on Colored MNIST (foreground color) and Colored MNIST (background color). While PGI achieves very good results on Colored MNIST (foreground color), \eg, top-1 unbiased accuracy when ratio = 0.995 and ratio = 0.98 in \cref{tab.fg_colored_mnist}, it achieves bad results on Colored MNIST (background color), \eg the lowest unbiased accuracy results in \cref{tab.bg_colored_mnist}. We suspect the reason is that PGI is overly sensitive to hyperparameters, \eg, the coefficient of the KL-divergence for debiasing. In contrast, our method achieves good results across two Colored MNIST variants without tuning or changing any hyperparameters.

Finally, we restate that Colored MNIST is a single-bias setting, which may not be the case in real-world scenarios where multiple biases exist. Therefore, we regard that we should focus more on our new Multi-Color dataset to evaluate the debiasing results \wrt multiple biases (\cref{tab.ablate_ratio_multi_color}), where DebiAN achieves better debiasing results.

\begin{table}[t]
  \caption{Average group accuracy results of gender classification on Transects~\cite{balakrishnan2020Eur.Conf.Comput.Vis.ECCV} dataset. All models are trained on CelebA dataset and evaluated on Transects \wrt two bias attributes---\texttt{Hair Length} and \texttt{Skin Color}. DebiAN achieves better results, which demonstrates that DebiAN better mitigate multiple biases simultaneously in the real-world multi-bias setting. Besides, it also reflects that DebiAN discovers \texttt{visible hair area} bias attribute to achieve better debiasing results \wrt \texttt{Hair Length} bias attribute}
  \label{tab.transects}
  \centering
  \begin{tabular}{@{}cccccc@{}}
  \toprule
  bias attribute & vanilla   & LfF       & EIIL       & PGI       & \textbf{DebiAN (Ours)} \\ \midrule
  \texttt{Hair Length}    & 55.1\tiny{$\pm$}5.8 & 54.7\tiny{$\pm$}2.9 & 54.0\tiny{$\pm$}0.4  & 56.2\tiny{$\pm$}1.3 & \textbf{60.5}\tiny{$\pm$}1.7     \\
  \texttt{Skin Color}     & 53.5\tiny{$\pm$}5.3 & 53.3\tiny{$\pm$}2.9 & 53.1\tiny{$\pm$}0.08 & 57.4\tiny{$\pm$}0.3 & \textbf{60.1}\tiny{$\pm$}1.2     \\ \bottomrule
  \end{tabular}
\end{table}

\section{More results on Mitigating Multiple Biases in Gender Classification}
\label{sec.supp.transects}

In Tab.~4, we show better DebiAN's better debiasing results \wrt \texttt{Wearing Lipstick} and \texttt{Heavy Makeup}. To demonstrate that DebiAN's better debiasing results \wrt more bias attributes, we evaluate on Transects~\cite{balakrishnan2020Eur.Conf.Comput.Vis.ECCV} dataset (mentioned on L567 in the main paper). Transects dataset contains high-quality face images synthesized by StyleGAN2~\cite{karras2020IEEEConf.Comput.Vis.PatternRecognit.CVPR}, which are also balanced \wrt multiple biases such as \texttt{Hair Length} and \texttt{Skin Color}. The dataset does not contain training split because it is designed as a testing set to identify biases in gender classifier. We use gender classifiers trained on CelebA dataset (same setting used in Sec.~4.2) to evaluate their debiasing performance \wrt \texttt{Hair Length} and \texttt{Skin Color} bias attributes on Transects dataset. The results are shown in \cref{tab.transects}, which demonstrates DebiAN's better capability in mitigating multiple biases simultaneously. Furthermore, the better results \wrt \texttt{Hair Length} can also reflect that DebiAN's \textit{discoverer} identifies \texttt{visible hair area} bias attribute (Fig.~6 and \cref{fig.supp.male_hair}).

\section{More Qualitative Results}

\subsection{More Examples of Discovered Biases on Face Images}
\label{subsec.supp.more_face_bias_discovery_vis}

\begin{figure}[t]
  \centering
  \includegraphics[width=0.5\linewidth]{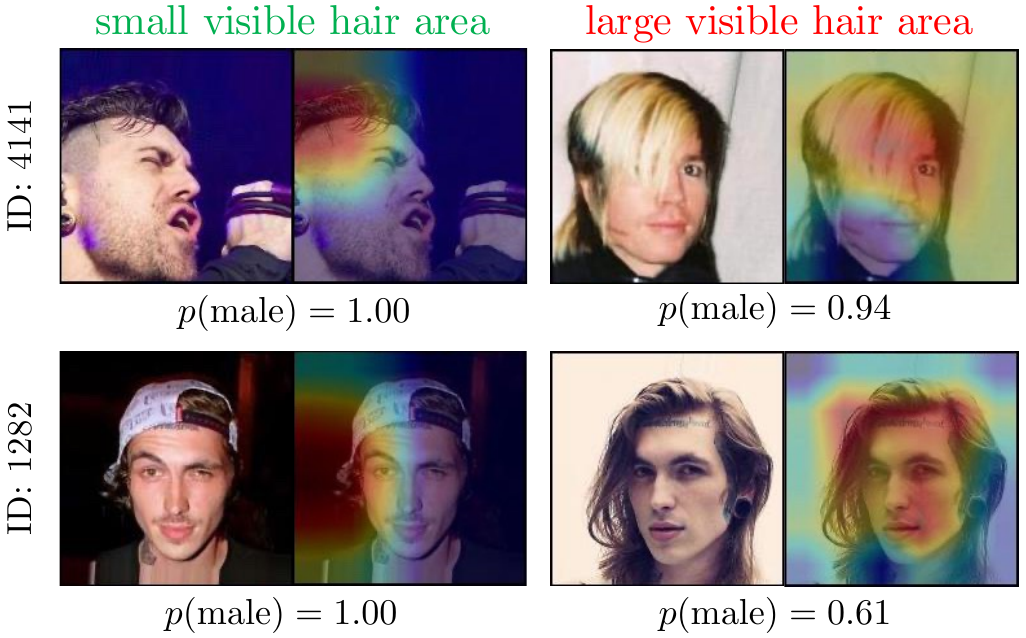}
  \caption{Discovered bias of gender classifier: \texttt{visible hair area}. $D$'s CAM saliency map is paired with each image. The vanilla gender classifier train on CelebA dataset performs worse for males with larger visible hair area}
  \label{fig.supp.male_hair}
\end{figure}

While \cref{fig.bias_discovery_acc_trend} shows the discovered \texttt{visible hair area} bias of gender classifier on female images, we further show the male image examples in \cref{fig.supp.male_hair}. $D$ focuses on the \texttt{visible hair area} to separate images into ``small visible hair area'' and ``large visible hair area'' groups, which is consistent with the female examples shown in \cref{fig.bias_discovery_acc_trend}.

\subsection{More Examples of Discovered Biases on Scene Images}
\label{subsec.supp.more_scene_bias_discovery_vis}

\begin{figure}[t]
  \centering
  \includegraphics[width=\linewidth]{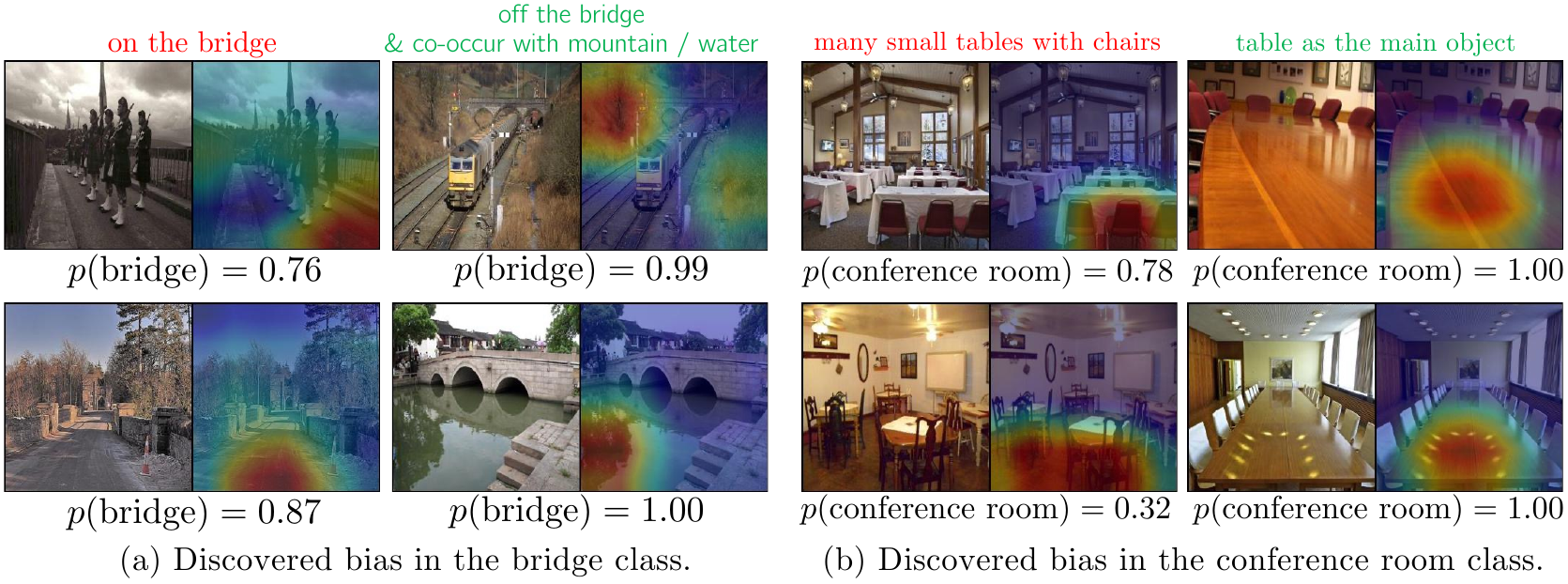}
  \caption{Discovered biases of vanilla scene classifiers. $D$'s CAM is paired with the image}
  \label{fig.supp.scene_bias}
\end{figure}

We show more examples of discovered biases on the scene images in \cref{fig.supp.scene_bias}. For bridge images, $D$ predicts two bias groups. When the photos are taken on the bridge, the vanilla classifier performs worse. In comparison, the vanilla classifier performs better when the photos are taken off the bridge with some correlated backgrounds, such as mountains or water. For conference room images, the vanilla classifier performs better when the table is the major object in the scene. However, it performs worse when the conference room images have many tables, and the tables do not occlude the chairs.

\section{Discussion}

\begin{table}[t]
  \caption{Comparing with focal loss~\cite{lin2017IEEEInt.Conf.Comput.Vis.ICCV}, a hard negative method, on Multi-Color MNIST dataset}
  \label{tab.supp.compare_focal}
  \centering
  \begin{tabular}{@{}ccccc@{}}
  \toprule
  left color       & right color      & \multirow{2}{*}{vanilla} & \multirow{2}{*}{focal} & \multirow{2}{*}{\textbf{DebiAN (Ours)}} \\
  skew = 0.995     & skew = 0.95      &                          &                        &                                         \\ \midrule
  bias-aligned     & bias-aligned     & \textbf{100.0}\tiny{$\pm$}0.0      & \textbf{100.0}\tiny{$\pm$}0.0    & \textbf{100.0}\tiny{$\pm$}0.0                     \\
  bias-aligned     & bias-conflicting & \textbf{98.7}\tiny{$\pm$}0.6       & {\ul 97.9\tiny{$\pm$}0.9}        & 98.1\tiny{$\pm$}0.4                               \\
  bias-conflicting & bias-aligned     & 6.5\tiny{$\pm$}1.0                 & {\ul 0.4\tiny{$\pm$}0.2}         & \textbf{55.4}\tiny{$\pm$}2.1                      \\
  bias-conflicting & bias-conflicting & 2.0\tiny{$\pm$}0.4                 & {\ul 1.2\tiny{$\pm$}0.3}         & \textbf{9.2}\tiny{$\pm$}0.8                       \\ \midrule
  \multicolumn{2}{c}{unbiased}        & 51.8\tiny{$\pm$}0.2                & {\ul 49.2\tiny{$\pm$}0.2}        & \textbf{65.7}\tiny{$\pm$}0.7                      \\ \midrule
  left color       & right color      & \multirow{2}{*}{vanilla} & \multirow{2}{*}{focal} & \multirow{2}{*}{\textbf{DebiAN (Ours)}} \\
  skew = 0.99      & skew = 0.95      &                          &                        &                                         \\  \midrule
  bias-aligned     & bias-aligned     & \textbf{100.0}\tiny{$\pm$}0.0      & \textbf{100.0}\tiny{$\pm$}0.0    & \textbf{100.0}\tiny{$\pm$}0.0                     \\
  bias-aligned     & bias-conflicting & \textbf{97.1}\tiny{$\pm$}0.5       & 95.7\tiny{$\pm$}0.6              & {\ul 95.6\tiny{$\pm$}0.8}                         \\
  bias-conflicting & bias-aligned     & 27.5\tiny{$\pm$}3.6                & {\ul 3.3\tiny{$\pm$}2.0}         & \textbf{76.5}\tiny{$\pm$}0.7                      \\
  bias-conflicting & bias-conflicting & 5.2\tiny{$\pm$}0.4                 & {\ul 2.4\tiny{$\pm$}0.3}         & \textbf{16.0}\tiny{$\pm$}1.8                      \\ \midrule
  \multicolumn{2}{c}{unbiased}        & 57.4\tiny{$\pm$}0.7                & {\ul 50.3\tiny{$\pm$}0.4}        & \textbf{72.0}\tiny{$\pm$}0.8                      \\ \midrule
  left color       & right color      & \multirow{2}{*}{vanilla} & \multirow{2}{*}{focal} & \multirow{2}{*}{\textbf{DebiAN (Ours)}} \\
  skew = 0.98      & skew = 0.95      &                          &                        &                                         \\ \midrule
  bias-aligned     & bias-aligned     & \textbf{100.0}\tiny{$\pm$}0.0      & \textbf{100.0}\tiny{$\pm$}0.0    & \textbf{100.0}\tiny{$\pm$}0.0                     \\
  bias-aligned     & bias-conflicting & 96.6\tiny{$\pm$}1.2                & {\ul 85.1\tiny{$\pm$}2.1}        & \textbf{97.1}\tiny{$\pm$}0.8                      \\
  bias-conflicting & bias-aligned     & 64.4\tiny{$\pm$}2.3                & {\ul 15.9\tiny{$\pm$}4.4}        & \textbf{85.1}\tiny{$\pm$}3.4                      \\
  bias-conflicting & bias-conflicting & 12.4\tiny{$\pm$}1.1                & {\ul 6.0\tiny{$\pm$}0.1}         & \textbf{19.4}\tiny{$\pm$}1.3                      \\ \midrule
  \multicolumn{2}{c}{unbiased}        & 68.3\tiny{$\pm$}1.4                & {\ul 51.7\tiny{$\pm$}0.8}        & \textbf{75.4}\tiny{$\pm$}0.9                      \\ \midrule
  left color       & right color      & \multirow{2}{*}{vanilla} & \multirow{2}{*}{focal} & \multirow{2}{*}{\textbf{DebiAN (Ours)}} \\
  skew = 0.95      & skew = 0.95      &                          &                        &                                         \\ \midrule
  bias-aligned     & bias-aligned     & \textbf{100.0}\tiny{$\pm$}0.0      & \textbf{100.0}\tiny{$\pm$}0.0    & \textbf{100.0}\tiny{$\pm$}0.0                     \\
  bias-aligned     & bias-conflicting & 91.1\tiny{$\pm$}2.3                & {\ul 63.7\tiny{$\pm$}3.8}        & \textbf{94.7}\tiny{$\pm$}0.9                      \\
  bias-conflicting & bias-aligned     & 87.0\tiny{$\pm$}3.7                & {\ul 54.4\tiny{$\pm$}4.1}        & \textbf{92.7}\tiny{$\pm$}1.3                      \\
  bias-conflicting & bias-conflicting & 26.0\tiny{$\pm$}1.3                & {\ul 11.3\tiny{$\pm$}0.1}        & \textbf{39.6}\tiny{$\pm$}0.2                      \\ \midrule
  \multicolumn{2}{c}{unbiased}        & 76.0\tiny{$\pm$}1.6                & {\ul 57.3\tiny{$\pm$}1.2}        & \textbf{81.8}\tiny{$\pm$}0.6                      \\ \bottomrule
  \end{tabular}
  \end{table}

\subsection{Is DebiAN a hard negative method?}
\label{subsec.supp.hard_neg}

No. Hard negative methods focus on addressing the imbalanced problem by letting the classifier focus on hard misclassified examples and pay less attention to easy examples. A seminal hard negative method is focal loss~\cite{lin2017IEEEInt.Conf.Comput.Vis.ICCV}, which reweighs the standard cross-entropy loss ($-\log (p_t)$, where $p_t$ is the predicted probability of the ground-truth class) to $- \alpha_t (1-p_t)^\gamma \log(p_t)$, where $\alpha_t$ and $\gamma$ are hyperparameters. Intuitively, it uses classifier's predicted probability to reweigh the cross-entropy loss, where hard examples (\ie, low $p_t$) are up-weighted with high $\alpha_t (1-p_t)^\gamma$ weights and easy examples (\ie, high $p_t$) are down-weighted with low $\alpha_t (1-p_t)^\gamma$ weights. Different from focal loss, our RCE loss (Eq.~6) uses \textit{discoverer}'s predicted bias group assignments to reweigh the cross-entropy loss, and the \textit{discoverer} is trained with $\mathcal{L}_\text{EOV}$ to differentiate \textit{classifier}'s $p_t$ on examples from the same target class where the Equal Opportunity is violated. Therefore, DebiAN is not a hard negative method because we do not use easy or hard samples (\ie, low or high $p_t$) to perform reweighing, but rather use samples' estimated bias group assignments to perform debiasing.

We also compare with focal loss on Multi-Color MNIST dataset. We use $\alpha=0.25$ and $\gamma=2.0$ as they perform the best in \cite{lin2017IEEEInt.Conf.Comput.Vis.ICCV}. The results are shown in \cref{tab.supp.compare_focal}, where focal loss's results are even worse than the vanilla model. The results prove that hard negative methods are not well-suited for debiasing. Since DebiAN is different from hard negative methods by using estimated bias group assignments to mitigate biases, our method achieves much better debiasing results.

\subsection{Is EOV loss simply doing clustering?}
No. EOV loss is used to train \textit{discoverer} to classify different bias groups values. Therefore, instead of simply clustering \textit{classifier}'s prediction, EOV loss guide the \textit{discoverer} to do a classification for the bias group assignment. The results in \cref{fig.bias_discovery_acc_trend}, \cref{fig.supp.bias_discovery_trend_diff_ratios}, and \cref{fig.supp.bias_discovery_trend_equally_salient} show that \textit{discoverer} can accurately classify if the samples on the testing set (unseen during training) are bias-aligned or bias-conflicting, demonstrating that EOV loss guides the \textit{discoverer} to do classification based on different bias groups values and it can generalize to testing set's images.

\begin{figure}[t]
  \centering
  \includegraphics[width=\linewidth]{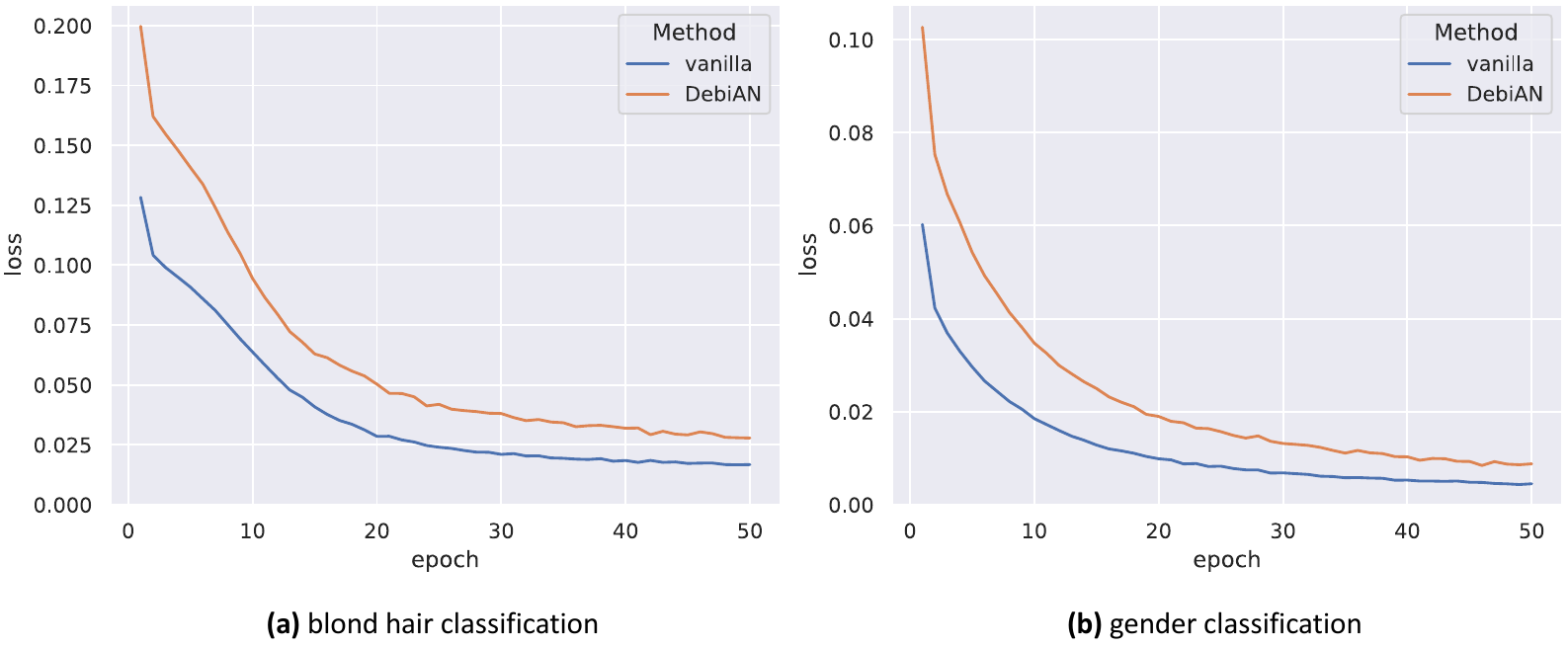}
  \caption{Vanilla's cross-entropy loss and DebiAN's RCE loss in (a) blond hair classification and (b) gender classification}
  \label{fig.supp.ce_loss}
\end{figure}

\subsection{Will \textit{classifier} achieve 100\% accuracy such that \textit{discoverer} cannot predict bias group assignments?}

No. First, note that \textit{discoverer}'s EOV loss is based on \textit{classifier}'s predicted probabilities instead of thresholded hard predictions. Therefore, 100\% accuracy does not indicate that \textit{discoverer} cannot predict the bias group assignments. Second, we show the vanilla model's cross-entropy loss and DebiAN's RCE loss on blond hair classification and gender classification tasks in \cref{fig.supp.ce_loss}. The results show that the losses do not completely converge to zero, which proves that there always exist samples in the training set that \textit{classifier} does not achieve 1.0 predicted probabilities. Therefore, it still leaves the room for \textit{discoverer} to predict bias group assignments based on \textit{classifier}'s different predictions on different samples.

\subsection{What if the mini-batch only contains the samples from a single bias group?}

It mainly happens under two conditions---(1) very strong spurious correlation; (2) small mini-batch size. For the first case, our ablation study on the ratios on Multi-Color MNIST dataset (\cref{subsec.supp.ablate_ratios_multi_color}) shows that DebiAN achieves better debiasing results even when the ratio of \texttt{left color} is 0.995 (\ie, very strong spurious correlation). For the second case, our ablation study on different batch sizes (\cref{subsec.supp.ablate_batch_size}) shows that our method still achieves strong debiasing results when the batch size is small.

\subsection{RCE loss compared with previous reweighing-based methods}

LfF and focal loss are two previous reweighing-based methods for unsupervised debiasing. At a high level, LfF, focal loss, and DebiAN's RCE loss all target at up-weighting worse performed samples and down-weighting better-performed samples. The difference is how to compute the weight. LfF uses the ratio of cross-entropy loss between the biased model and the classifier to compute weights. Focal loss, as a hard negative method (see \cref{subsec.supp.hard_neg}), directly uses classifier's predicted probabilities to compute the weights. Different from previous methods, RCE loss uses \textit{discoverer}'s predicted bias group assignment to compute the weights. Compared with LfF and focal, DebiAN achieves better debiasing results (Tab.~1-6 and \cref{tab.supp.compare_focal}).

\subsection{Evaluation on Discovered Unknown Biases}
In Fig.~6, Fig.~7, \cref{fig.supp.male_hair}, and \cref{fig.supp.scene_bias}, we show some interesting unknown biases that human may not preconceive via saliency map. One may wonder if there exist other approaches to evaluate the results. First, it is hard to directly quantify the findings due to lack of annotations of the discovered bias attributes, \eg, CelebA does not have attribute annotations or segmentation annotations of \texttt{visible hair area}. Using other datasets (\eg, COCO~\cite{lin2014Eur.Conf.Comput.Vis.ECCV}) with more attribute or segmentation ground-truth may not help since the discovered unknown biases may still be out of the annotations. Second, UDIS~\cite{krishnakumar2021Br.Mach.Vis.Conf.BMVC}, a recent bias discovery method, also uses saliency maps to interpret the bias. We believe that using saliency maps is an established evaluation protocol in this task. Third, although it is hard to evaluate bias discovery in real-world dataset, our evaluation of bias discovery on Multi-Color MNIST (\cref{fig.bias_discovery_acc_trend}, \cref{fig.supp.bias_discovery_trend_diff_ratios}, and \cref{fig.supp.bias_discovery_trend_equally_salient}) has shown that DebiAN achieves strong bias discovery results. Finally, we believe that our better debiasing results \wrt \texttt{Hair Length} bias attribute on the Transects dataset can also indirectly prove that \textit{discoverer} identifies \texttt{visible hair area} bias (see \cref{sec.supp.transects}).

\subsection{Why not add two colors to the foreground in Multi-Color MNIST?}
The reason is that foreground digits are not always well aligned to the center of the images. If we assign two colors to the foreground digit based on whether the foreground is on the left or right, we may encounter cases where the digit is mainly on the right and only has a tiny area on the left, \eg an italic digit ``1.'' Thus, we choose to add colors to the background.

\subsection{Difference between Multi-Color MNIST and Biased MNIST}

\citet{shrestha2022IEEEWinterConf.Appl.Comput.Vis.WACV} recently proposed the Biased MNIST dataset, which contains seven biases. However, all seven biases in Biased MNIST share the same bias-aligned ratio (\ie, 0.7). In contrast, our Multi-color MNIST contains two biases that are in different bias-aligned ratios, which we believe is more common in real-world scenarios and can better reveal the failure modes of existing debiasing methods. For example, while LfF performs the best in the Biased MNIST benchmark, our Multi-Color MNIST dataset reveals that LfF can only discover the more salient bias---the bias with a larger bias-align ratio.

\subsection{Why evaluate on scene classification task?}
First, we regard that scene classification as a core vision task on par with object classification. Second, while many previous debiasing works create datasets~\cite{nam2020Adv.NeuralInf.Process.Syst.,kim2021IEEEInt.Conf.Comput.Vis.ICCVa} that contain a single bias (\eg, artificially introducing the spurious correlation \wrt a single bias), we believe that the classical cross-dataset generalization evaluation approach~\cite{torralba2011IEEEConf.Comput.Vis.PatternRecognit.CVPR} does not have the single-bias assumption. The subgroup distribution \wrt multiple biases may vary across different datasets, which is closer to the real-world setting.

\subsection{Limitations and Future Directions} We list some limitations that DebiAN has not fully resolved. First, we only assume that the bias attribute is binary or continuously valued from 0 to 1 (\ie, two bias attribute groups). Future works can focus on extending DebiAN to discover and mitigate unknown biases with more than two groups. Second, DebiAN can only discover the biases caused by spurious correlation rather than lack of coverage. For example, suppose a face image dataset \text{only} contains long-hair female images and does not contain any short-hair female images. In that case, DebiAN cannot discover the \texttt{hair length} bias attribute because the \textit{discoverer} does not have samples to categorize female images into two groups in terms of the \texttt{hair length} bias attribute. Finally, in terms of interpreting the discovered biases, DebiAN's approach, using the saliency maps on real-world images, is not as easy as interpreting biases from synthesized counterfactual images~\cite{lang2021IEEEInt.Conf.Comput.Vis.ICCV,li2021IEEEInt.Conf.Comput.Vis.ICCV}. Future works can further explore better interpreting the discovered unknown biases on real-world images.

\subsection{Potential Negative Social Impact}
\label{subsec.supp.neg_social_impact}

One potential negative social impact is that DebiAN's discovered biases could be used as a way to choose real-world images as the adversarial images to attack visual models in some safety-critical domains, \eg, self-driving cars. Therefore, we encourage the defender to use DebiAN to mitigate the biases as the defense strategy.

Since our bias discovery approach relies on the fairness criterion based on equations, \eg, equal true positive rates among two groups, our method cannot identify the biases that a fairness criterion cannot capture, \eg, discrimination against the historically disadvantaged group. To mitigate this issue, we include a model card~\cite{mitchell2019ACMConf.FairnessAccount.Transpar.} in the released code to clarify that our method's intended use case is discovering and mitigating biases that violate the equal opportunity fairness criterion~\cite{hardt2016Adv.NeuralInf.Process.Syst.}, and the model's out-of-scope use case is identifying or mitigating other biases that cannot be captured by the equation of a fairness criterion, \eg, discrimination against the historically disadvantaged group~\cite{hanna2020Conf.FairnessAccount.Transpar.critical}.

\end{document}